\documentclass{birkjour_t2}
\usepackage[noadjust]{cite}
\usepackage{amsthm, amssymb, amsmath} 

\usepackage{graphicx}
\usepackage{caption}

\usepackage{tabularx,multirow,multicol,subcaption,rotating}
\usepackage{comment}
\usepackage{hyperref,hypcap}
\usepackage{cleveref}
\usepackage{xcolor} 
\usepackage{tikz}
\usepackage{pgfplots}

\hypersetup{allcolors=black}
\numberwithin{equation}{section}
\usepackage[toc,page]{appendix}
 \theoremstyle{definition}
 
 \theoremstyle{remark}

\newcommand{\Inv}{\mathop{\mathrm{Inv}}}

\newcommand{\Modification}[1]{ {\color{black} #1} }

\begin{document}
\bibliographystyle{spmpsci}
%
%
%
%
%
%
%
%

\captionsetup[figure]{labelfont={bf},labelformat={default},labelsep=quad,name={Fig.}}
\captionsetup[table]{labelfont=bf,name={Table},labelformat={default},labelsep=quad}

\title[Machine Learning Clifford invariants of ADE Coxeter elements]
 {Machine Learning Clifford invariants of ADE Coxeter elements}

\author[Chen]{Siqi Chen}
\address{
Physics and Astronomy Department,
Stony Brook University, Stony Brook, NY 11794, USA
}
\email{siqi.chen.1@stonybrook.edu}

\author[Dechant]{Pierre-Philippe Dechant}
\address{%
School of Mathematics, University of Leeds, Leeds  LS2 9JT, United Kingdom}
\email{ppd22@cantab.net}

\author[He]{Yang-Hui He}
\address{
 London Institute for Mathematical Sciences, Royal Institution, London W1S 4BS, UK
}
\email{hey@maths.ox.ac.uk}

\author[Heyes]{Elli Heyes}
\address{
Department of Mathematics, City, University of London, EC1V 0HB, UK
}
\email{elli.heyes@city.ac.uk}

\author[Hirst]{Edward Hirst}
\address{
Centre for Theoretical Physics, 
Queen Mary University of London, 
E1 4NS, UK
}
\email{e.hirst@qmul.ac.uk}

\author[Riabchenko]{Dmitrii Riabchenko}
\address{
Department of Mathematics, City, University of London, EC1V 0HB, UK
}
\email{Dmitrii.Riabchenko@city.ac.uk}


\keywords{Exceptional symmetries,
invariants,
Cayley-Hamilton theorem,
Clifford algebras, 
Coxeter groups, 
root systems, 
Platonic solids.\\
\textbf{Report Number.} QMUL-PH-23-15}

\date{\#\#\#, 2023}

\dedicatory{To Jim Humphreys, Peter Neumann, and John McKay}

\begin{abstract}
There has been recent interest in novel Clifford geometric invariants of linear transformations. This motivates the investigation of such invariants for a certain type of geometric transformation of interest in the context of root systems, reflection groups, Lie groups and Lie algebras: the Coxeter transformations. We perform exhaustive calculations of all Coxeter transformations for $A_8$, $D_8$ and $E_8$ for a choice of basis of simple roots and compute their invariants, using high-performance computing. This computational algebra paradigm generates a dataset that can then be mined using techniques from data science such as supervised and unsupervised machine learning. In this paper we focus on neural network classification and principal component analysis. Since the output -- the invariants -- is fully determined by the choice of simple roots and the permutation order of the corresponding reflections in the Coxeter element, we expect huge degeneracy in the mapping. This provides the perfect setup for machine learning, and indeed we see that the datasets can be machine learned to very high accuracy. This paper is a pump-priming study in experimental mathematics using Clifford algebras, showing that such Clifford algebraic datasets are amenable to machine learning, and shedding light on relationships between these novel and other well-known geometric invariants and also giving rise to analytic results. 
\end{abstract}

\maketitle

\section{Introduction}\label{sec_Intro}


\Modification{Great interest in Clifford geometric invariants of linear transformations, originally proposed in \cite{hestenes2012clifford}, was sparked in recent work from a practical \cite{lasenby2022reconstructing, lasenby2022some} and theoretical  \cite{lasenby2022some, shirokov2021computing, abdulkhaev2021explicit, doran2003geometric} point of view.
Orthogonal transformations, such as rotations, and their invariants are important in engineering, e.g. moving cameras, robots etc. The types of transformations we are looking at in this work are also rotations, particularly interesting because of their symmetry structures.}
Linear transformation invariants are traditionally exemplified by the determinant and trace, which appear in the highest and lowest coefficients of the characteristic polynomial. Typically such linear transformations are described by matrices; however, in Clifford algebras one has the alternative to implement orthogonal transformations via versors. In Clifford algebras, algebraic objects have a clearer geometric interpretation than in the standard matrix approach. There is a systematic way of calculating multivector invariants of linear transformations via what are called `simplicial derivatives', which we will introduce further in the next section. These Clifford geometric invariants are then systematically related to geometric invariant spaces of the linear transformation and the coefficients in the characteristic polynomial and Cayley-Hamilton theorem\footnote{The decomposition of a linear transformation into orthogonal eigenspaces is also related to some interesting recent work by \cite{roelfs2023graded}.}. This serves as motivation to study this type of new geometric invariant of linear transformations.

From the perspective of some of our other work on root systems and reflection groups \cite{dechant2013affine, dechant2018trinity, dechant2017clifford} we are particularly interested in a certain type of linear transformations that occurs in this root system context: the `Coxeter elements' or `Coxeter transformations'. These are a particular type of orthogonal transformation in reflection/Coxeter groups \cite{Humphreys1990Coxeter}. They are the group elements of the highest order (called the `Coxeter number', $h$) and they are all conjugate to each other. High-dimensional root systems are notoriously difficult to visualise, and projection into a distinguished plane (called a `Coxeter plane') is a common way of visualising the geometry. In these planes, the Coxeter elements just act by $h$-fold rotations. Root systems are determined by a subset called the `simple roots', which act as a basis for the vector space and each determines a `simple reflection' in the hyperplane to which they are orthogonal\footnote{The existence of the Coxeter plane relies on the simple roots admitting a separation into two sets that are mutually orthogonal within each set, often visualised as a bipartite (alternating) colouring of the corresponding Dynkin diagram.}. A Coxeter element is then just given by multiplying each of the simple reflections once in some permutation order, which at the versor level is just encoded by multiplying together the root vectors in the Clifford algebra directly, doubly covering the orthogonal transformation. This set of permutations giving rise to  a set of Coxeter versors will be the focus of this paper, as these allow us to calculate the full set of invariants from them, which we will refer to as the set of characteristic multivectors (SOCM).

In previous work, the authors have established a paradigm for experimental mathematics: first, using computational algebra techniques and high-performance computing (HPC) one can generate a dataset of algebraic data; this dataset can then be mined by applying the standard data science toolkit, in order to find patterns that were not obvious from an analytical perspective
\cite{dechant2022cluster, cheung2022clustering, he2023machine, he2018calabi, he2022murmurations, he2021calabi}. Mathematicians often calculate examples of interest by hand to formulate or test hypotheses. Essentially, this computational algebra approach automates and scales up such an approach, \Modification{and turns the problem into a `data analysis' task}\footnote{Clifford algebra multivector computations can easily be performed in such a \texttt{python} HPC setup using the \texttt{galgebra} package \cite{Bromborsky2020}.}. One can either calculate a very large number of examples and analyse these statistically via \Modification{`data analysis'}; or in other cases of interest, it may be possible to calculate \emph{all} the cases exhaustively and analyse the patterns that emerge, which can help with hypothesis formulation and theorem proving.  

At the ICCA conference in Hefei in 2020 a talk on this approach sparked much interest, resulting in the creation of a Topical Collection (TC) on `Machine Learning Mathematical Structures' in the journal \emph{Advances in Applied Clifford Algebras} \cite{ablamowicz2021call}. This TC sits at the intersection of 3 different topics -- machine learning, mathematical structures, and Clifford algebras -- and was intended to stimulate new research across these interfaces. Whilst there have been many activities in pairwise combinations (e.g. Machine Learning and Clifford algebras \cite{buchholz2008clifford, vieira2023bicomplex, kobayashi2021synthesis, kuroe2011models, bayro2010clifford, pepe2022learning, dorst2014total, groenendijk2023geometric, pepe2022using, wang2016clifford, moya2021quaternion, aouiti2021finite, sriraman2022stability}, mathematical structures and Clifford algebras \cite{ablamowicz2021ternary, shirokov2019calculation, ablamowicz2018classification, da2020efficient, helmstetter2023various, roelfs2023geometric, wilson2021problem}, and of course our examples for machine learning algebraic structures earlier as well as others in this TC \cite{heal2022deep, bena2022algorithmically}), we are not aware of any research that actually sits at the intersection of \emph{all three}. 

In the interest of such a first non-trivial intersection we therefore see our work as sufficiently motivated: to investigate the machine learning of Clifford invariants of Coxeter elements. We consider the three 8-dimensional root systems $A_8$, $D_8$ and $E_8$, which are of course of wider interest in terms of exceptional $E_8$ and ADE patterns \cite{Dechant2016Birth, baez2017icosahedron, CDHM2024ADE}. Eight dimensions allow $8!$ permutations of the simple roots, which can all be explicitly calculated with HPC. They give rise to linear transformations (Coxeter transformations) whose simplicial derivatives can be taken in an automated way, and whose characteristic multivectors are computed as the output. The $40320$ input permutations are a large enough number to make these examples accessible to data science techniques such as machine learning classification tasks (e.g. distinguishing between the three ADE types), principal component analysis etc. Of course, the actual number of Coxeter versors will be lower, firstly because of the fact that the simple roots can be decomposed into two sets that are mutually orthogonal within sets (leading to a $k!$ reduction whenever $k$ orthogonal simple roots are grouped together), but secondly because such roots are also more widely orthogonal to other roots outside of the sets, as given by the adjacency in the Dynkin diagram such that two roots are orthogonal if there exists no edge between them. So in practice, there will be some degeneracy in the mapping from the permutations to the Coxeter versors which will result in a significantly reduced set of invariants. 
Such (anti)commutation properties in the Coxeter element can in principle be understood analytically. But in the interest of the experimental mathematics approach followed here for now we prefer to just calculate the permutations exhaustively using computational algebra and treat the repeats (whose structure is interesting in its own right, and learning it essentially means learning the root system geometry and Dynkin diagram) as the data science standard practice of data augmentation. Analytical considerations will largely be presented in a companion paper though we point out a few instances where exhaustive computation has led to analytical insights in this paper. 

We organise this paper as follows. In Section \ref{sec_back}, we introduce some of the detail on the aforementioned Clifford simplicial derivatives and invariants, as well as root systems and Coxeter transformations. We then discuss in Section \ref{sec:data} what datasets we are mining, and how they were generated using computational algebra. This section also contains some exploratory data analysis around numbers of distinct invariants as well as the connectivity structure of the bivector invariants. We then move onto Machine Learning in Section \ref{sec_ML}; in particular, we discuss predictive performance as well as ternary classification tasks, before moving onto gradient saliency sensitivity on the input and Principal Component Analysis. We conclude in Section \ref{sec_concl}. Our computer code scripts and data can be found on \href{https://github.com/DimaDroid/ML_Clifford_Invariants.git}{GitHub}\footnote{\url{https://github.com/DimaDroid/ML_Clifford_Invariants.git}}.

\section{Background}\label{sec_back}

Thorough introductions to root systems and Clifford algebras are available elsewhere \cite{dechant2017clifford} so here we will be succinct. A root system lives in the arena of a vector space with a scalar product (which immediately allows one to consider the corresponding Clifford algebra). It is a collection of vectors (called  `roots', and customarily denoted $\alpha$) in that vector space which is invariant under all the reflections in the hyperplanes to which the root vectors are perpendicular. We will only consider root systems with roots of the same length, which can be assumed to be normalised \footnote{Note this is different from the normalisation convention used in Lie theory}. Such reflections in the normal hyperplanes are given by $x \rightarrow x- 2(x\cdot n) n$, where $x$ is the vector to be transformed and $n$ is a unit normal to the hyperplane. 

\begin{figure}[!t]
    
	\begin{center}	

		\begin{tikzpicture}[scale=0.5,
		knoten/.style={        circle,      inner sep=.1cm,        draw}
		]
		\node at  (1,.7) (knoten1) [knoten,  color=white!0!black, ball color=white ] {};
		\node at  (3,.7) (knoten2) [knoten,  color=white!0!black, ball color=black] {};
		\node at  (5,.7) (knoten3) [knoten,  color=white!0!black, ball color=white ] {};
		\node at  (7,.7) (knoten4) [knoten,  color=white!0!black, ball color=black] {};

		\node at  (9,.7) (knoten5) [knoten,  color=white!0!black, ball color=white] {};
		\node at (11,.7) (knoten6) [knoten,  color=white!0!black, ball color=black] {};
		\node at (13,.7) (knoten7) [knoten,  color=white!0!black, ball color=white] {};
		\node at (15,0.7) (knoten8) [knoten,  color=white!0!black, ball color=black] {};

		\node at  (1,0)  (alpha1) {$\alpha_1$};
		\node at  (3,0)  (alpha2) {$\alpha_2$};
		\node at  (5,0)  (alpha3) {$\alpha_3$};
		\node at  (7,0)  (alpha4) {$\alpha_4$};
		\node at  (9,0)  (alpha7) {$\alpha_5$};
		\node at (11,0)  (alpha6) {$\alpha_6$};
		\node at (13,0)  (alpha5) {$\alpha_7$};
		\node at (15,0) (alpha8) {$\alpha_8$};

		\path  (knoten1) edge (knoten2);
		\path  (knoten2) edge (knoten3);
		\path  (knoten3) edge (knoten4);
		\path  (knoten4) edge (knoten5);
		\path  (knoten5) edge (knoten6);
		\path  (knoten6) edge (knoten7);
		\path  (knoten7) edge (knoten8);

		\end{tikzpicture}

\begin{tikzpicture}[scale=0.5,
		knoten/.style={        circle,      inner sep=.1cm,        draw}
		]
		\node at  (1,.7) (knoten1) [knoten,  color=white!0!black, ball color=white ] {};
		\node at  (3,.7) (knoten2) [knoten,  color=white!0!black, ball color=black] {};
		\node at  (5,.7) (knoten3) [knoten,  color=white!0!black, ball color=white ] {};
		\node at  (7,.7) (knoten4) [knoten,  color=white!0!black, ball color=black] {};

		\node at  (9,.7) (knoten5) [knoten,  color=white!0!black, ball color=white] {};
		\node at (11,.7) (knoten6) [knoten,  color=white!0!black, ball color=black] {};
		\node at (13,-.5) (knoten7) [knoten,  color=white!0!black, ball color=white] {};
		\node at (13,1.9) (knoten8) [knoten,  color=white!0!black, ball color=white] {};

		\node at  (1,0)  (alpha1) {$\alpha_1$};
		\node at  (3,0)  (alpha2) {$\alpha_2$};
		\node at  (5,0)  (alpha3) {$\alpha_3$};
		\node at  (7,0)  (alpha4) {$\alpha_4$};
		\node at  (9,0)  (alpha7) {$\alpha_5$};
		\node at (11,0)  (alpha6) {$\alpha_6$};
		\node at (13,0.1)  (alpha5) {$\alpha_7$};
		\node at (13,1.2) (alpha8) {$\alpha_8$};

		\path  (knoten1) edge (knoten2);
		\path  (knoten2) edge (knoten3);
		\path  (knoten3) edge (knoten4);
		\path  (knoten4) edge (knoten5);
		\path  (knoten5) edge (knoten6);
		\path  (knoten6) edge (knoten7);
		\path  (knoten6) edge (knoten8);

		\end{tikzpicture}
  
\begin{tikzpicture}[scale=0.5,
		knoten/.style={        circle,      inner sep=.1cm,        draw}
		]
		\node at  (1,.7) (knoten1) [knoten,  color=white!0!black, ball color=white ] {};
		\node at  (3,.7) (knoten2) [knoten,  color=white!0!black, ball color=black] {};
		\node at  (5,.7) (knoten3) [knoten,  color=white!0!black, ball color=white ] {};
		\node at  (7,.7) (knoten4) [knoten,  color=white!0!black, ball color=black] {};

		\node at  (9,.7) (knoten5) [knoten,  color=white!0!black, ball color=white] {};
		\node at (11,.7) (knoten6) [knoten,  color=white!0!black, ball color=black] {};
		\node at (13,.7) (knoten7) [knoten,  color=white!0!black, ball color=white] {};
		\node at (9,2.2) (knoten8) [knoten,  color=white!0!black, ball color=black] {};

		\node at  (1,0)  (alpha1) {$\alpha_1$};
		\node at  (3,0)  (alpha2) {$\alpha_2$};
		\node at  (5,0)  (alpha3) {$\alpha_3$};
		\node at  (7,0)  (alpha4) {$\alpha_4$};
		\node at  (9,0)  (alpha7) {$\alpha_5$};
		\node at (11,0)  (alpha6) {$\alpha_6$};
		\node at (13,0)  (alpha5) {$\alpha_7$};
		\node at (9,2.8) (alpha8) {$\alpha_8$};

		\path  (knoten1) edge (knoten2);
		\path  (knoten2) edge (knoten3);
		\path  (knoten3) edge (knoten4);
		\path  (knoten4) edge (knoten5);
		\path  (knoten5) edge (knoten6);
		\path  (knoten6) edge (knoten7);
		\path  (knoten5) edge (knoten8);

		\end{tikzpicture}

	\end{center}

    \caption{The diagrams of the 8-dimensional simply-laced root systems $A_8$, $D_8$ and $E_8$ (vertically downwards respectively), along with our labelling for the simple roots and a bipartite colouring. }\label{ADEdiagrams}
\end{figure}
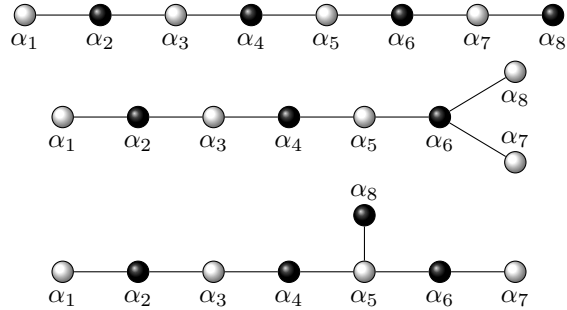

A subset called `simple roots' is sufficient to write all roots as (in our case) integer linear combinations of this basis of simple roots, whilst their corresponding reflections, the `simple reflections', generate the reflection group. Taking these simple reflections all exactly once leads to interesting types of group elements called `Coxeter elements'. They are of the same order $h$ (the `Coxeter number'), and have invariant planes, called `Coxeter planes', which are useful for visualising root systems in any dimension (via projection into these planes). These reflection groups  have interesting integer -- in fact prime -- invariants, that are characteristic of the geometry, called  `exponents' $m$. This name derives from the fact that Coxeter elements act on different invariant planes by $h$-fold rotations by $m$ times $2\pi/h$, which is usually interpreted as a complex eigenvalue of the Coxeter element (even though we are by assumption in a real vector space). The root system geometry can also be encoded in diagrammatic form (called `Coxeter-Dynkin diagram'), where each simple root corresponds to a node and orthogonal nodes are not linked, whilst roots at $2\pi/3$ angles are connected with a link (we will only be considering such `simply-laced' examples, see Fig. \ref{ADEdiagrams}). Likewise, our simply-laced examples are tree-like and admit an alternate colouring (or `bipartite', e.g. black and white). This effectively means that all black roots are orthogonal to each other, and likewise for the white roots. This colouring means that there are distinguished types of Coxeter elements where first all the black reflections are taken, and then all the white (or the other way round). We will call these `bipartite' Coxeter elements. This bipartite colouring also implies the existence of the Coxeter plane via a more complex argument, the details of which we will omit here, but which relies on the adjacency matrix of the Dynkin diagram having a distinguished largest eigenvalue and corresponding eigenvector, the Perron-Frobenius eigenvector (which \emph{will} make an appearance below). In our labelling of the 8 simple roots for $A_8$, $D_8$ and $E_8$, $\alpha_1$ to $\alpha_7$
make one long string. The different diagrams arise depending on where the 8th root $\alpha_8$ attaches: at the terminal node $\alpha_7$ for $A_8$ (leading to bilateral symmetry), at the penultimate node $\alpha_6$ for $D_8$ (leading to permutation symmetry of the terminal nodes), or $\alpha_5$ for $E_8$\footnote{Note that attaching to other roots is symmetry-equivalent to the options just mentioned with the exception of attaching to $\alpha_4$, which leads to something called affine $E_7$, or $\tilde{E}_7$.}.

As mentioned above, Clifford algebras can be constructed when one is working in an $n$-dimensional vector space with an inner product, giving rise to a $2^n$-dimensional algebra of `multivectors'. The scalar product is given as the symmetric part of the geometric product, i.e. $a\cdot b = \frac{1}{2}(ab+ba)$\footnote{The outer product $a\wedge b = \frac{1}{2}(ab-ba)$ is the antisymmetric part, is a bivector and determines the plane that two vectors generically span.}. Substituting this in the reflection formula above results in a cancellation which leads to the uniquely simple `sandwiching' reflection formula in Clifford algebras
\begin{equation}
x \rightarrow x- 2(x\cdot n) n = -nxn.
\end{equation}
Both $n$ and $-n$ doubly cover the same reflection. Via the Cartan-Dieudonn\'e theorem orthogonal transformations are just products of such reflections so that one can build up 
\begin{equation}
x \rightarrow \pm n_k\cdots n_1xn_1 \cdots n_k = \pm\tilde{A}xA
\end{equation}
such transformations via defining multivectors that are the products of normal vectors which encode the reflection hyperplanes, $A=n_1 \cdots n_k$ (called `versors'), and a tilde denotes reversing the order of these vectors in the product. These versors again doubly cover the transformation. 

We discuss here for a moment how this applies when the orthogonal transformation is a Coxeter element. In traditional root system notation, the simple reflections are denoted $s_i$ such that a Coxeter element is denoted $w = s_1\cdots s_n$. In the above versor framework, the reflections are encoded by the root vectors themselves (as a double cover), whilst the multivectors $W$ that one gets from multiplying the simple roots together $\alpha_1 \cdots \alpha_n$ doubly cover $w$
\begin{equation}
wx \rightarrow \pm \alpha_k\cdots\alpha_1x\alpha_1 \cdots \alpha_k = \pm\tilde{W}xW.
\end{equation}

We return now to the setting of linear transformations in Clifford algebras more generally again. Let us denote this linear transformation by $f(x)$. In order to calculate the desired invariants of this linear transformation (the SOCM), we define the concept of `simplicial derivatives'. 

First, let $\{a_k\}, k=1, \dots, n$ denote a frame, i.e. a basis. Often we use either a Euclidean basis $e_i$ or the basis of simple roots, $\alpha_i$. We denote by $\{a^k\}$ its reciprocal frame
such that $a^i\cdot a_j = \delta^i_j$. In a Euclidean basis this is effectively the basis itself; for a basis of simple roots the reciprocals are more commonly known as co-roots  (up to a different conventional normalisation factor). 
We also define $b_k  =f(a_k)$ as the transformation acting on the basis frame vectors. 
The $r$th simplicial derivative is then essentially defined as a combinatorial object
\begin{equation}
    \partial _{(r)}f _{(r)}=\sum (a^{j_r} \wedge \dots \wedge a ^{j_1})(b_{j_1} \wedge \dots \wedge b_{j_r})
\end{equation}
with sum over $0<j_1<\dots<j_r\le n$\footnote{This is due to the original notion of a multivector derivative essentially being equivalent to a projection.}. These simplicial derivatives are invariants of the linear transformation and are therefore `characteristic multivectors' with geometric significance. 

Now \cite{hestenes2012clifford} showed that it is the scalar parts of these geometric invariants (denoted by $\partial _{(s)}* f _{(s)}$) that constitute the coefficients in the 
Cayley-Hamilton theorem 
$$C_f(\lambda) = \sum_{s=0}^{m}(-\lambda)^{m-s}\partial _{(s)}* f _{(s)}$$
(where $\partial _{(0)}* f _{(0)}$ is interpreted as $1$)
and the characteristic polynomial
$$ \sum_{s=0}^{m}(-1)^{m-s}\partial _{(s)}* f _{(s)}f^{m-s}(a)=0$$
for any vector $a$ (where $f^0(a)$ is interpreted as $a$).

One can explicitly verify this for our examples. Using the \texttt{galgebra} package, one can perform calculations in the 256-dimensional multivector algebra, calculating Coxeter versors from permutations of the simple roots, and from that simplicial derivatives and geometric invariants. We will refer to the simplicial derivatives $\partial _{(r)}f _{(r)}$ as {the invariant of order $r$ or $\Inv_r$} (and to the full set as SOCM). 

Since we are considering an even orthogonal transformation in an 8-dimensional space we get some interesting structure in these invariants (see Table \ref{table:inv}): firstly, we note that  only even multivectors occur (in principle, this allows us to reduce the length of the 256-dimensional multivectors by half). Secondly, the lowest order invariant only has a scalar part (trivially), the next picks up a bivector term, the next one a quadrivector term, the next a sextivector, till finally $\Inv_4$ (generically) has a pseudoscalar term. Then it decreases again. In fact, thirdly, in our case we have a certain  `mirror symmetry', where the top half  in the Table is equal to the bottom half, though this is not generally the case. In fact, all these pieces, which we could denote by  $\Inv_r ^k$ are separately invariant under the Coxeter versor: $\tilde{W}\Inv_r ^kW=\Inv_r ^k$. So these $\Inv_r ^k$ are eigenmultivectors of the Coxeter element of grade $k$, but they do not have to be $k$-blades (i.e. be able to be written as the outer product of $k$ vectors\footnote{Something also noticed in the example in \cite{lasenby2022reconstructing}.}).

\begin{table}[h]
\centering
    \begin{tabular}{ |c||c|c|c|c|c| } 
        \hline
         & \multicolumn{5}{|c|}{Subinvariant} \\
        \hline
        Invariants by Order & scalar & bivector & quadrivector & sextivector & pseudoscalar \\
        \hline \hline
        $\Inv _0$ & X &   &  &   &  \\
        \hline
        $\Inv _1$ & X & X &  &   &  \\
        \hline
$        \Inv _2$ & X & X & X &   &  \\
        \hline
        $\Inv _3$ &X& X & X & X &  \\
        \hline
        $\Inv _4$ & X & X & X & X &  X \\
        \hline
        $\Inv _5 $& X & X & X & X &  \\
        \hline
      $ \Inv _ 6 $& X & X & X &   &  \\
        \hline
        $\Inv _7$ & X & X &  &   &  \\
        \hline
       $ \Inv _8$ & X &   &  &   &  \\
        \hline
        \end{tabular}
    \caption{Structure of the characteristic multivectors: non-zero grades are indicated by an X.}
    \label{table:inv}
\end{table}

So amongst other multivector components, e.g. for $E_8$ we in particular have 4 invariant bivectors from the invariants. It turns out that these are orthogonal. The Coxeter element also acts on 4 invariant orthogonal bivectors (giving planes, and they are blades by construction) via the Coxeter plane construction, so there is an immediate question of how our characteristic multivectors relate to exponents and degrees. In fact, we will say here already that for $E_8$ one can show that the two sets of 4 orthogonal eigenvectors (from the SOCM and the Coxeter construction) span the same 4d-subspace of the 28d bivector space. Reflection groups can also have other interesting invariant subspaces such as two $H_4$-invariant subspaces in $E_8$ 
\cite{dechant2013affine, Dechant2017e8}. We are exploring these more analytical questions more fully in the companion paper.

\section{Datasets}\label{sec:data}
We choose dimension 8 because of the following compromise: $8!=40320$ gives us something resembling `big data' which is accessible to data science techniques, whilst being computationally tractable\footnote{With the caveat that there is degeneracy in the permutations leading to the same or similar Coxeter elements and thus invariants, reducing the true number of different output vectors. \Modification{Although it was not obvious from the beginning, especially for $E_9$ and $D_8$.}}. It is also the last dimension in which there are three simply-laced root systems, with the exceptional $E_8$ adding some variety to the $A_n$ and $D_n$ families that exist in arbitrary dimensions. So we select $A_8$, $D_8$ and $E_8$, as this gives us scope for three-way (ternary) classification tasks and ADE patterns are of course of wider interest. 

The input vectors are the set of permutations in 8 elements, e.g. (0,1,2,3,4,5,6,7), labelling the simple roots $\alpha_1$ through to $\alpha_8$ and encoding in which order the simple roots are taken in for computing the Coxeter versor. The outputs are the 9 invariants $\{\Inv _0, \dots, \Inv _8\}$ as multivectors.
\begin{equation}
   \text{input} = (0,1,2,3,4,5,6,7) = \alpha_1\dots \alpha_8 \rightarrow \{\Inv _0, \dots, \Inv _8\} = \text{SOCM} = \text{output}
\end{equation} 
In 8-dimensions the multivector invariants have 256 components (some of which are trivial\footnote{Since the odd components are typically $0$ one could reduce this if needed, but for completeness and generalisability we haven't.}). 


\subsection{Data Generation}
\label{sec:data_gen}

The computational algebra approach followed here used \texttt{python} with the \texttt{galgebra} package for multivector computations \cite{Bromborsky2020}. Exploratory analysis for single permutations was performed in Jupyter notebooks but once parallelised the computations were run on clusters at Queen Mary, University of London, City, University of London, and University of Leeds. Data and Code can be found on \href{https://github.com/DimaDroid/ML_Clifford_Invariants.git}{GitHub}. We performed computations both in an Euclidean basis\footnote{\Modification{Roots in the root system are often defined as columns of components in the Euclidean orthonormal basis in some higher dimensional space. However, since the set of simple roots can generate the root system via addition, we can also take simple roots as a basis, although not orthonormal. While this may seem more complicated, it is more meaningful geometrically because everything we compute in geometric algebra using simple roots can be eventually expressed in the basis of simple roots.}}, which is a bit more straightforward, and the basis of simple roots via the multivector basis that it induces, which is more meaningful geometrically and less dependent on the choice of simple roots in the Euclidean basis. According to our earlier discussion around Table \ref{table:inv} we can also extract different grades of these invariants (e.g. scalar, bivector etc), which we refer to as `subinvariants' $\Inv_r ^k$, from the full set (SOCM).

\subsection{Frequency Analysis}
\label{sec:freq_anal}

As was mentioned previously, for each of the $A_8$, $D_8$ and $E_8$ root systems there are $8! = 40320$ permutations of 8 root vectors from which we can construct the corresponding Coxeter elements and the 9 geometric invariants. 

Each invariant is a sum of 8 subinvariants written in terms of the wedge product of the different number of basis vectors for the 8-dimensional vector space. These 8 subinvariants are scalar, vector, bivector, trivector, quadrivector, and so on up to 8-vector pseudoscalar. 
The components for each of the subinvariants can be written down in a chosen basis, e.g. in terms of simple roots. 
This allows us to compare SOCMs corresponding to different Coxeter elements, or invariants of a chosen order, or focus on subinvariants within the invariant of a chosen order.

\begin{figure}[!h]
    \centering
    \begin{subfigure}{0.45\textwidth}
        \centering
        \includegraphics[width=0.99\textwidth]{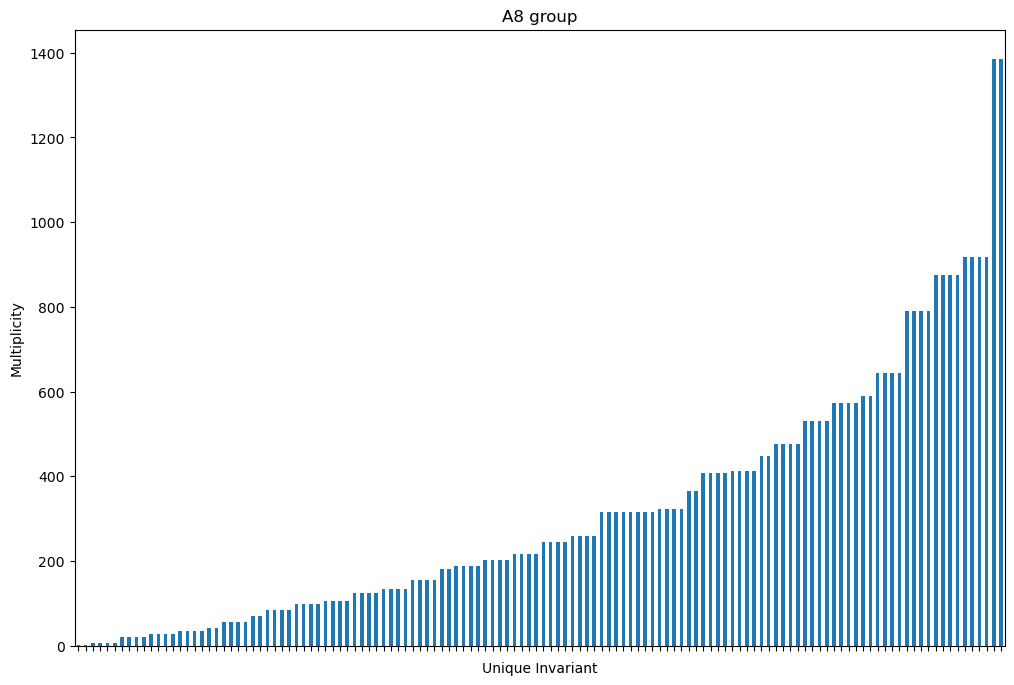}
        \caption{A8}
    \end{subfigure}
    \begin{subfigure}{0.45\textwidth}
        \centering
        \includegraphics[width=0.99\textwidth]{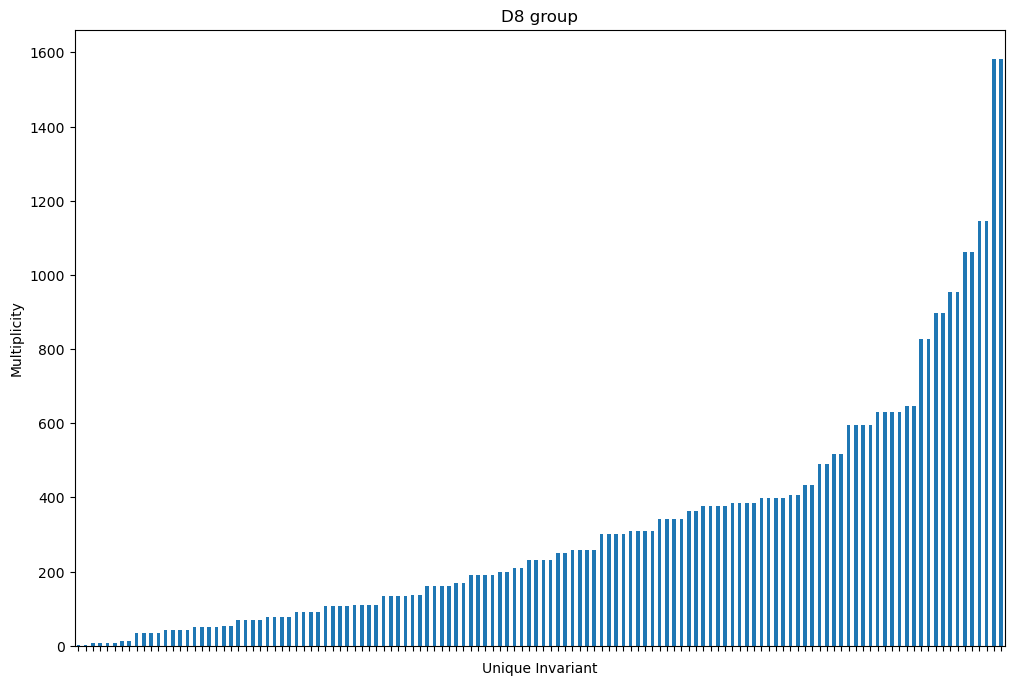}
        \caption{$D_8$}
    \end{subfigure} 
    \begin{subfigure}{0.45\textwidth}
        \centering
        \includegraphics[width=0.99\textwidth]{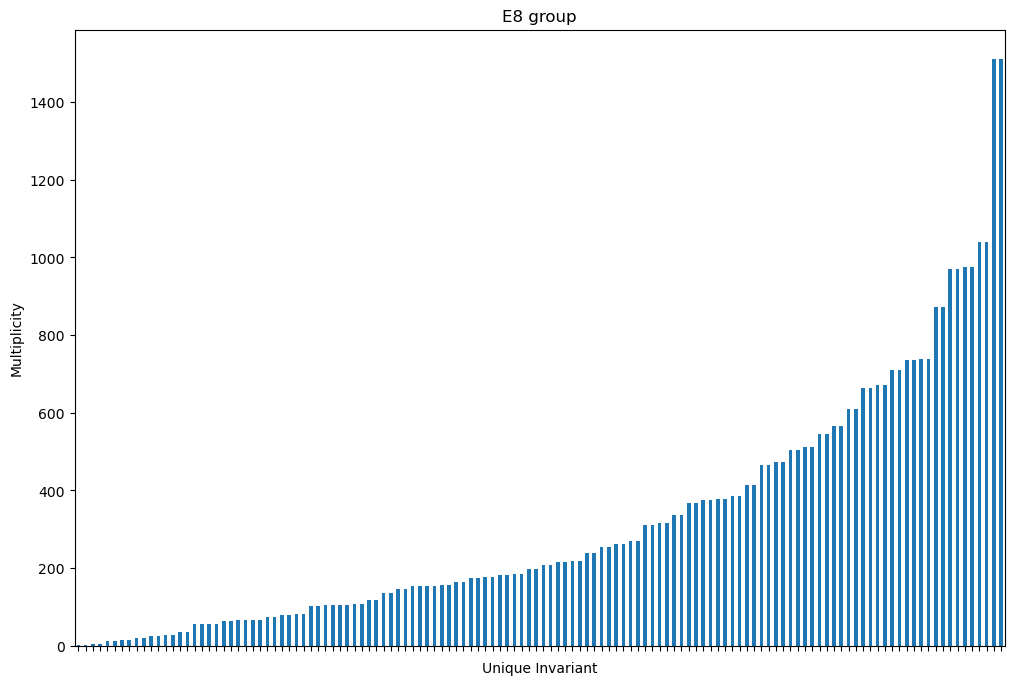}
        \caption{E8}
    \end{subfigure} 
    \caption{Sorted multiplicities of the 128 unique SOCMs, for each root system considered: $A_8$, $D_8$, $E_8$ respectively. $A_8$ is mostly quadruplets, $E_8$ mostly doublets and $D_8$ half and half. \Modification{See \href{https://github.com/DimaDroid/ML_Clifford_Invariants.git}{GitHub} for the full list of values.}}\label{InvariantMultiplicities}
\end{figure}

These exhaustive computational algebra calculations already show interesting results. On the highest level, we compare components of SOCMs and find that, although there are 40320 SOCMs we can construct, only 128 are distinct, and this is the same number for each of the algebras. We can think of these as `classes' of SOCMs with the same components. Each class has a certain number of representatives\footnote{They have different order of roots in a permutation encoding Coxeter versor.} in it which we call frequency. \Modification{The frequency of classes, which come in groups of 2, 4 or 8 and all have the same value, we call Doublets, Quadruplets and Octuplets, respectively.}
Although the number of classes for each of the algebras is the same, frequencies of individual classes differ, see Figure \ref{InvariantMultiplicities}. 

\noindent In the following, we will be using two types of operations on permutations:

\begin{itemize}
    \item Inversion - we say that two permutations are related by inversion if the order of simple roots for these permutations is reversed relative to each other, e.g. (0,1,2,3,4,5,6,7) and (7,6,5,4,3,2,1,0).

    
    \item  $B \leftrightarrow W$ - following the bipartite colouring of black roots and white roots, we can reduce (with some degeneracy) a permutation to a black and white `barcode'. For example, (2,4,6,8,1,3,5,7) becomes `$\bullet\bullet\bullet\bullet\circ\circ\circ \, \circ$'. We say that two permutations are related by  $B \leftrightarrow W$ if we replace black roots by white roots and vice versa. 
\end{itemize}

\noindent There are some common features among all three algebras:
\begin{itemize}
    \item \Modification{All the frequency values come in Doublets, Quadruplets and Octuplets;}

    \item The highest frequencies appear in Doublets;

    \item For Doublets, permutations of the class elements in one class are related by inversion to permutations of the class elements in another class. Quadruplets are essentially two Doublets with the same frequency and Octuplets are two Quadruplets with the same frequency.
\end{itemize}

\noindent Some other features which are different:
\begin{itemize}
    \item For $A_8$:
    
    \begin{itemize}
        \item Frequencies of all classes are odd numbers;

        \item There is a Doublet with the lowest frequency equal to 1 (i,e, two unique invariants). Classes in this doublet are represented by permutations (0,1,2,3,4,5,6,7) and (7,6,5,4,3,2,1,0) \footnote{One might call them `maximally non-commuting permutations', where none of the roots adjacent in the permutation are orthogonal.};

        \item There is a Doublet with the highest frequency equal to 1385. This Doublet consists of two invariants which are given by bipartite Coxeter elements: one of them is given by a Coxeter element with first 4 black roots with increased root number and then 4 white ones with increased root number as well; the second one is very similar and has first 4 white roots and then 4 black ones with increased root number in both subsets;

        \item For Quadruplets, there are pairs of classes that are related by inversion. In addition, these pairs within the Quadruplet are related to each other by $B \leftrightarrow W$. Presumably, having inverse barcodes signifies similar combinatorial properties that result in the same frequency. 

        \item As was mentioned before, in Doublets, two classes within it are related by inversion. In addition, the two classes are related to each other by $B \leftrightarrow W$ symmetry: if we assign a black or white colour to every simple root according to Figure \ref{ADEdiagrams}, we get a black and white `barcode' for a permutation encoding a Coxeter element. One can check that the barcode for the first class is the inversion (change black to white and vice versa) of the barcode for the other class within the Doublet, i.e. the Doublets are self-dual under $B \leftrightarrow W$.

        \item In some Quadruplets, there is even more $B \leftrightarrow W$ inversion symmetry: $B \leftrightarrow W$ symmetry between the pairs of classes that are related by inversion is enriched by the $B \leftrightarrow W$ symmetry within the pairs. This is because the barcode mapping is degenerate, i.e. non-equivalent permutations can give rise to the same barcode;

        \item An Octuplet appears as two Quadruplets with the same frequency;

        \item There are 8 Doublets, 26 Quadruplets and 1 Octuplet in total.
    \end{itemize}

    \item For $D_8$:
    
    \begin{itemize}
        \item All frequencies are even numbers; 
        
        \item There are no unique invariants, a Doublet with a frequency equal to 2, and a Doublet with the highest frequency equal to 1582;


        \item No signs of $B \leftrightarrow W$ symmetry; 

        \item There are 20 Doublets and 22 Quadruplets.
    \end{itemize}
    
    \item For $E_8$:
    
    \begin{itemize}
        \item All frequencies are odd numbers; 

        
        \item There are no unique invariants, a Doublet with the lowest frequency equal to 3 and a Doublet with the highest frequency equal to 1511;

        \item No signs of $B \leftrightarrow W$ symmetry;

        \item There are 58 Doublets and 3 Quadruplets.
    \end{itemize}
\end{itemize}


On the level of subinvariants within the invariants, one can perform the same analysis and find frequencies given in Tables \ref{table:AE8} and \ref{table:D8}. The tables for $A_8$ and $E_8$ are identical; all three groups have the same frequencies for bivector and quadrivector subinvariants. Empty cells denote the fact that all subinvariants for this order of invariants are trivially zero (some are also less-trivially zero).

\begin{table}[!t]
\centering
    \begin{tabular}{ |c||c|c|c|c|c| } 
        \hline
         & \multicolumn{5}{|c|}{Subinvariant} \\
        \hline
        Invariant Order & scalar & bivector & quadrivector & sextivector & pseudoscalar \\
        \hline \hline
        $\Inv_0$ & 1 &   &  &   &  \\
        \hline
        $\Inv_1$ & 1 & 128 &  &   &  \\
        \hline
        $\Inv_2$ & 1 & 128 & 64 &   &  \\
        \hline
        $\Inv_3$ & 1 & 128 & 64 & 128 &  \\
        \hline
        $\Inv_4$ & 1 & 128 & 64 & 128 &  2 \\
        \hline
        $\Inv_5$ & 1 & 128 & 64 & 128 &  \\
        \hline
        $\Inv_6$ & 1 & 128 & 64 &   &  \\
        \hline
        $\Inv_7$ & 1 & 128 &  &   &  \\
        \hline
        $\Inv_8$ & 1 &   &  &   &  \\
        \hline
        \end{tabular}
    \caption{Frequencies of subinvariants for $A_8$/$E_8$ group. Empty cells denote the fact that all subinvariants for this order of invariants are trivially zero.}
    \label{table:AE8}
\medskip
    \begin{tabular}{ |c||c|c|c|c|c| } 
        \hline
         & \multicolumn{5}{|c|}{Subinvariant} \\
        \hline
        Invariant Order & scalar & bivector & quadrivector & sextivector & pseudoscalar \\
        \hline \hline
        $\Inv_0$ & 1  &  &  &  &  \\
        \hline
        $\Inv_1$ & 1  & 128 &  &  &  \\
        \hline
        $\Inv_2$ &  0& 128 & 64 &  &  \\
        \hline
        $\Inv_3$ &  0& 128 & 64 & 32 &  \\
        \hline
        $\Inv_4$ &  0& 128 & 64 & 32 &  0\\
        \hline
        $\Inv_5$ &  0& 128 & 64 & 32 &  \\
        \hline
        $\Inv_6$ &  0& 128 & 64 &  &  \\
        \hline
        $\Inv_7$ & 1  & 128 &  &  &  \\
        \hline
        $\Inv_8$ & 1  &   &  &  &  \\
        \hline
        \end{tabular}
    \caption{Frequencies of subinvariants for $D_8$ group. Empty cells denote the fact that all subinvariants for this order of invariants are trivially zero. But there are also some non-trivial zeroes to do with the $D_8$ geometry, in which the factorisation of the Coxeter element into orthogonal eigenspaces contains two true reflections, also signalled by having two exponents of $h/2$. }
    \label{table:D8}
\end{table}

\begin{table}[!h]
\centering
    \begin{tabular}{ |c||c|c|c|c|c| } 
        \hline
         & \multicolumn{5}{|c|}{Subinvariant} \\
        \hline
        Invariant Order & scalar & bivector & quadrivector & sextivector & pseudoscalar \\
        \hline \hline
        $\Inv_0$ & 1   &       &       &       &    \\
        \hline
        $\Inv_1$ & 1   &  64   &       &       &    \\
        \hline
        $\Inv_2$ & 1   &  64   &  64   &       &    \\
        \hline
        $\Inv_3$ & 1   &  64   &  64   &  64   &    \\
        \hline
        $\Inv_4$ & 1   &  64   &  64   &  64   &   1 \\
        \hline
        $\Inv_5$ & 1   &  64   &  64   &  64   &    \\
        \hline
        $\Inv_6$ & 1   &  64   &  64   &       &    \\
        \hline
        $\Inv_7$ & 1   &  64   &       &       &    \\
        \hline
        $\Inv_8$ & 1   &       &       &       &    \\
        \hline
        \end{tabular}
    \caption{Frequencies of subinvariants for $A_8$ group up to an overall minus sign. Empty cells denote the fact that all subinvariants for this order of invariants are trivially zero.}
    \label{table:A8unSigned}
\medskip
    \begin{tabular}{ |c||c|c|c|c|c| } 
        \hline
         & \multicolumn{5}{|c|}{Subinvariant} \\
        \hline
        Invariant Order & scalar & bivector & quadrivector & sextivector & pseudoscalar \\
        \hline \hline
        $\Inv_0$ & 1   &       &       &       &     \\
        \hline
        $\Inv_1$ & 1   &  40   &       &       &     \\
        \hline
        $\Inv_2$ & 1   &  40   &  64   &       &     \\
        \hline
        $\Inv_3$ & 1   &  52   &  48   &  36   &     \\
        \hline
        $\Inv_4$ & 1   &  64   &  64   &  64   &   1 \\
        \hline
        $\Inv_5$ & 1   &  52   &  48   &  36   &     \\
        \hline
        $\Inv_6$ & 1   &  40   &  64   &       &     \\
        \hline
        $\Inv_7$ & 1   &  40   &       &       &     \\
        \hline
        $\Inv_8$ & 1   &       &       &       &     \\
        \hline
        \end{tabular}
    \caption{Frequencies of subinvariants for $E_8$ group up to an overall minus sign. Empty cells denote the fact that all subinvariants for this order of invariants are trivially zero.}
    \label{table:E8unSigned}
\medskip
    \begin{tabular}{ |c||c|c|c|c|c| } 
        \hline
         & \multicolumn{5}{|c|}{Subinvariant} \\
        \hline
        Invariant Order & scalar & bivector & quadrivector & sextivector & pseudoscalar \\
        \hline \hline

        $\Inv_0$ & 1   &       &       &       & \\
        \hline
        $\Inv_1$ & 1   &  64   &       &       &     \\
        \hline
        $\Inv_2$ & 0   &  64   &  64   &       &     \\
        \hline
        $\Inv_3$ & 0   &  64   &  64   &  16   &     \\
        \hline
        $\Inv_4$ & 0   &  64   &  64   &  16   &   0 \\
        \hline
        $\Inv_5$ & 0   &  64   &  64   &  16   &     \\
        \hline
        $\Inv_6$ & 0   &  64   &  64   &       &     \\
        \hline
        $\Inv_7$ & 1   &  64   &       &       &     \\
        \hline
        $\Inv_8$ & 1   &       &       &       &     \\
        \hline
        \end{tabular}
    \caption{Frequencies of subinvariants for $D_8$ group up to an overall minus sign. Empty cells denote the fact that all subinvariants for this order of invariants are trivially zero. But there are also some non-trivial zeroes to do with the $D_8$ geometry, in which the factorisation of the Coxeter element into orthogonal eigenspaces contains two true reflections, also signalled by having two exponents of $h/2$. }
    \label{table:D8unSigned}
\end{table} 


Another interesting thing to look at is the frequencies of invariants and subinvariants with the identification of objects that differ up to an overall minus sign. We find that this modification does not alter the frequencies of the full invariants, however, it does change the frequencies of subinvariants, see Tables \ref{table:A8unSigned}, \ref{table:E8unSigned}, \ref{table:D8unSigned}. We see that frequencies for bivector and sextivector (and pseudoscalar for $A_8$) subinvariants for $A_8$ and $D_8$ are halved, meaning that half of these subinvariants differ from the other half by a minus sign. At the same time, scalar and quadrivector subinvariants are unchanged. The picture is different for $E_8$, where bivector, quadrivector and sextivector frequencies change non-trivially under sign identification. 

\Modification{The idea behind these observations is to understand the symmetries of the root system. The frequency of multiplicities for the three algebras should be determined by the permutation of these symmetries. It is rather simple to determine the number of unique invariants and explain the existence of Doublets and Quadruplets for the $A_8$ algebra due to its simple Dynkin diagram, but much harder for the $D_8$ and $E_8$ algebras.}

\subsection{Bivector Subinvariants}
Now we are restricting our focus to the bivector parts of the invariants, which as subinvariants are of particular interest since bivectors generate planes for rotation (such as the Coxeter plane central to the study of these root systems).
Each of the bivector subinvariants has 28 entries, corresponding to the $\tiny{\begin{pmatrix} 8 \\ 2 \end{pmatrix}}$ combinations that form a basis for the bivector subspace, whether this is in a Euclidean basis or in the basis of simple roots. Here, we will be working in the basis of simple roots. Each bivector subinvariant hence takes the form: $\sum_{i,j=1|i<j}^8 c_{ij} (\alpha_i \wedge \alpha_j)$, for the 8 simple root basis vectors $\alpha_i$, and general coefficients $c_{ij}$, which turn out to be even integers.  This is motivated by the observation that rather intriguingly, the bivector part of the bipartite $E_8$ Coxeter element gives precisely rise to the $E_8$ diagram etc. 

\subsubsection{Interpretation as Graphs}\mbox{}\\
From each bivector subinvariant, one can construct a graph. This is done by associating a vertex to each simple root, and including the edge between vertices $i$ and $j$ if $c_{ij} \neq 0$. This construction method manifestly creates undirected unweighted simple graphs (with no loops as $c_{ii}=0 \ \forall i$, and at most one edge between any pair of vertices).
The generated graph is practically constructed via a symmetric adjacency matrix, with binary entries, such that $c_{ij}$ is taken as the upper triangle of a symmetric matrix with any non-zero entries converted to 1.
Since the adjacency matrices are symmetric their eigenvalues are all real, and one can begin to analyse their eigenspectra\footnote{Note that one may also create a directed weighted graph by setting the adjacency matrix upper triangle to be $c_{ij}$; however as the matrix is anti-symmetric, eigenvalues are complex, and hence cannot be sorted sensibly for analysis.}.

The Perron-Frobenius theorem \cite{Perron1907,frobenius1912matrizen} asserts that square matrices with positive integer coefficients have a unique largest real eigenvalue. 
In particular, for undirected graph adjacency matrices this maximum eigenvalue takes value in the range $(0,n-1]$, for graphs with $n$ vertices\footnote{The upper bound is saturated by the complete graph on $n$ vertices.}.
Furthermore, it turns out that the A, D, \& E Dynkin diagrams are particularly special in the space of undirected graphs, in that they are the \textit{only} connected graphs whose maximum eigenvalue $< 2$ by Smith's theorem \cite{smith1970some}. In fact, the Coxeter elements of bipartite form were observed to just give the Coxeter-Dynkin diagram of the $A_8$, $D_8$ and $E_8$ root systems. 
Since   bivector graphs can be more generally  induced  by all forms of Coxeter elements, this motivates the study of the maximum eigenvalues for all the respective undirected graphs.

Returning to our databases, except for the trivial zero invariant formed from the bivector subinvariant of $\Inv_0$ and $\Inv_8$, an initial unanticipated observation is that each of the graphs constructed from each of the bivector subinvariants across the 3 databases are all connected.
In addition to this, there is no overlap of bivector subinvariants between algebras (excluding the trivial zero invariant).
Also there is no overlap of bivector subinvariants between the orders of invariants they come from, within each of the algebras.
However, there is small repetition of adjacency matrices (i.e. after reducing non-zero entries to 1), and also graphs. 
Specifically, there are $A_8$: (0, 38, 12), $D_8$: (0, 1, 6), $E_8$: (0, 0, 25) repeated (subinvariants, adjacencies, graphs) between orders 1 to 4, for each algebra respectively.

Analysing the multiplicities of the bivector subinvariants, for ($A_8$, $D_8$, $E_8$) there are (513, 513, 513) distinct subinvariants across all orders for each algebra respectively.
When considering the undirected adjacency matrices constructed from these (out of $2^{28} \sim 2.7 \times 10^9$ possible undirected adjacency matrices), these bivector subinvariants reduce to (219, 256, 251) distinct adjacency matrices respectively.
These matrices then further reduce to respectively (88, 144, 137) non-isomorphic graphs (out of $11117$  possible non-isomorphic graphs \cite{oeis_connectedgraphs})\footnote{Noting that the trivial zero invariant (all 28 coefficients $c_{ij} = 0$) contributes a subinvariant, an adjacency matrix, and an empty graph to the counts for each algebra.}.

\begin{figure}[t]
    \centering
    \includegraphics[width=0.55\textwidth]{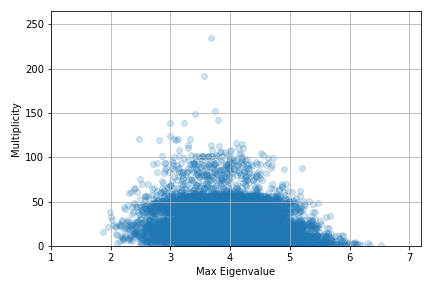}
    \caption{Distributions of the maximum eigenvalues for 282240 random connected matrices (of which 282086 are unique matrices, overall having 9741 unique eigenvalues).}\label{RandomMatrixMaxEvs}
\end{figure}

\begin{figure}[t]
    \centering
    \begin{subfigure}{0.32\textwidth}
        \centering
        \includegraphics[width=0.99\textwidth]{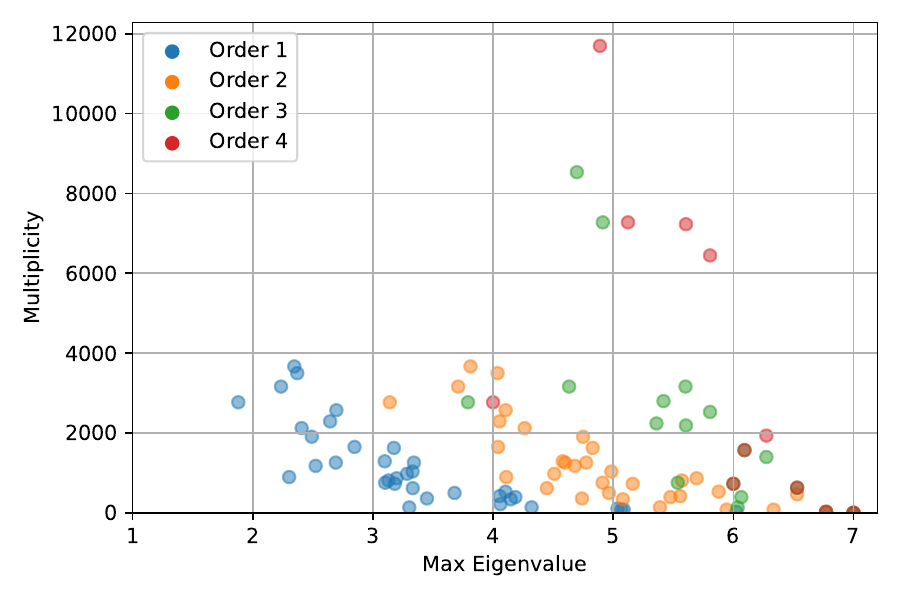}
        \caption{$A_8$ Subinvariants}
    \end{subfigure}
    \begin{subfigure}{0.32\textwidth}
        \centering
        \includegraphics[width=0.99\textwidth]{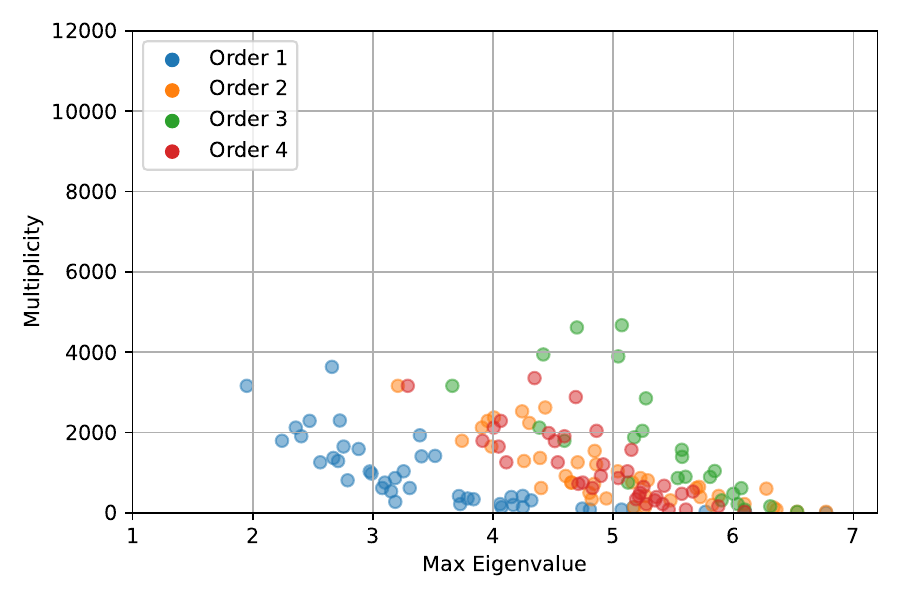}
        \caption{$D_8$ Subinvariants}
    \end{subfigure} 
    \begin{subfigure}{0.32\textwidth}
        \centering
        \includegraphics[width=0.99\textwidth]{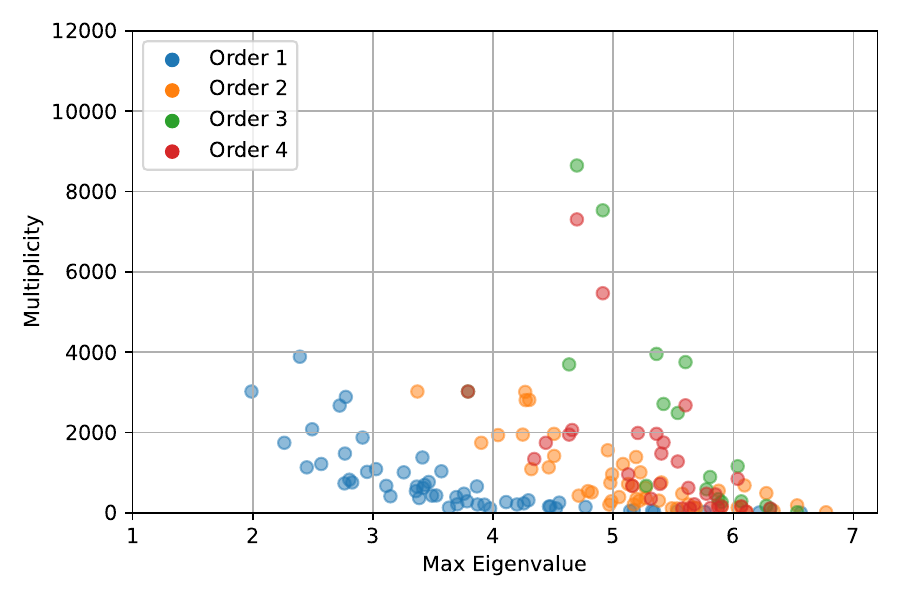}
        \caption{$E_8$ Subinvariants}
    \end{subfigure}
    \caption{Distributions of the maximum eigenvalues for each of the bivector subinvariants for each of the considered  algebras: $A_8$, $D_8$, $E_8$ respectively. Data includes all 282240 non-empty bivector subinvariants, coloured according to which order invariant they correspond to.}\label{BivectorMaxEvDistributions}
\end{figure}

Now in examining the distribution of the maximum eigenvalues, we first note that the trivial zero invariant, which is equivalent to the empty graph, has all eigenvalues zero, so is omitted in the following analysis.
To set a baseline for comparison we sample as many random connected adjacency matrices as non-zero bivector subinvariants occur in each dataset (282240), compute their maximum eigenvalues, and plot the respective histogram of multiplicities in Figure \ref{RandomMatrixMaxEvs}.
The maximum eigenvalues for the adjacency matrices constructed from each bivector subinvariant were then computed for each algebra's dataset, and histograms of their distributions for each algebra are shown in Figure \ref{BivectorMaxEvDistributions}, coloured according to the invariant order that they came from. 
In each class corresponding to invariant orders 1 to 4 (i.e. $\Inv_1$ to $\Inv_4)$, there are $A_8$: [36, 34, 18, 11], $D_8$: [43, 41, 24, 36], $E_8$: [54, 49, 18, 30] distinct eigenvalues respectively with multiplicities as shown in the plots.
Note that all these multiplicities reduce to 1 when considering the unique non-isomorphic graphs at each order.

From the plots, it can be seen that the distributions do appear to roughly follow a partition according to the order of the invariant they come from. {This is perhaps hinting at how these different order subinvariants span different subspaces of the full space of subinvariants, as dictated by their eigendecompositions.}
Additionally, the actual $A_8$, $D_8$, and $E_8$ graphs occur as bivector subinvariant graphs for a large number of the $\Inv_1$'s \textit{only} in each respective root system's dataset (the point with maximum eigenvalue below 2 in the $A_8$ plot is the $A_8$ graph of Figure \ref{ADEdiagrams}, etc). Presumably, these are due to Coxeter elements in bipartite form, and are in accordance with Smith's theorem.

\subsubsection{Eigenvector Centrality}\mbox{}\\
The maximum eigenvalue of a non-negative matrix has a corresponding eigenvector with exclusively non-negative entries, as also dictated by the Perron-Frobenius theorem. 
One can then associate each of these normalised non-negative entries to a centrality score for the graph node with corresponding index. This is known as \textit{eigenvector centrality} \cite{bonacich_1987}. 

For these bivector subinvariant graphs, there are 8 nodes (corresponding to the simple roots) and hence 8 respective centrality scores which can be computed for each graph for each invariant order across each root system.
Since the centrality scores are normalised, when examining the distribution of measures across the nodes, it is most interesting to consider: (1) the most central node with the highest score; as well as (2) the score distribution variance.
Respectively, these then indicate which parts of the graph are most important to the connectivity (and hence the most significant bivector contribution to its graph structure); and the extent to which this significance is polarised towards the requirement on this most central node to ensure connectivity. For instance, for the $D_8$ and $E_8$ Dynkin diagrams, the triply connected simple root is $6$ and $5$ respectively, so we would expect these to have highest centrality. Likewise, the middle roots $4$ and $5$ in $A_8$ should be the most central. But these Dynkin diagrams are in some way minimal, and other bivector diagrams will be `more fully connected', so we expect centrality of different roots to change for more general Coxeter elements. 

\begin{figure}[t]
    \centering
    \begin{subfigure}{0.32\textwidth}
        \centering
        \includegraphics[width=0.99\textwidth]{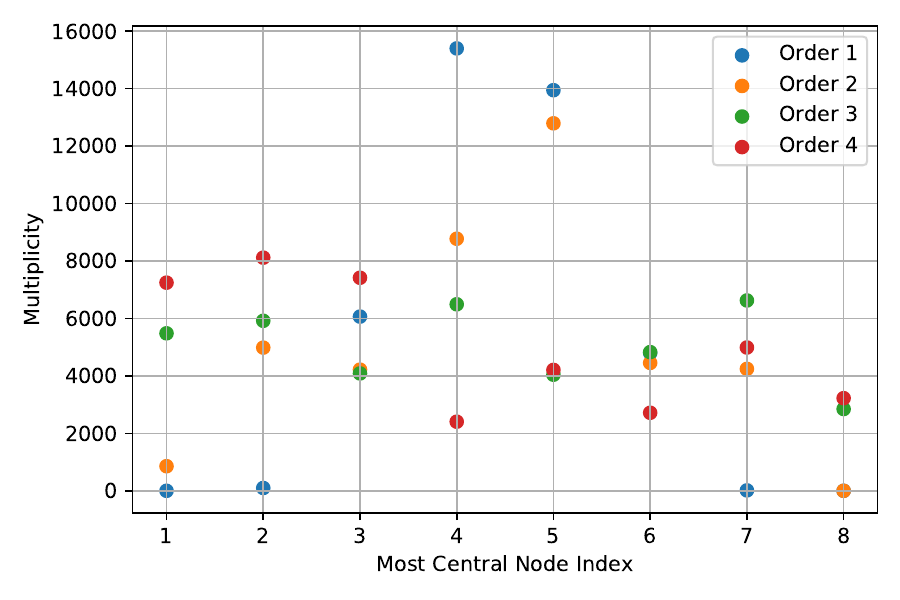}
        \caption{$A_8$ Subinvariants}
    \end{subfigure}
    \begin{subfigure}{0.32\textwidth}
        \centering
        \includegraphics[width=0.99\textwidth]{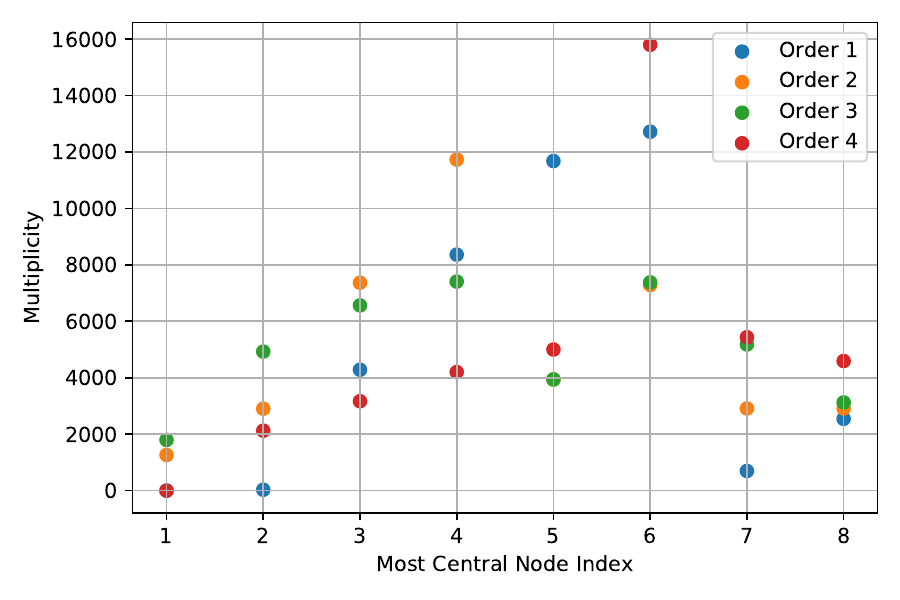}
        \caption{$D_8$ Subinvariants}
    \end{subfigure} 
    \begin{subfigure}{0.32\textwidth}
        \centering
        \includegraphics[width=0.99\textwidth]{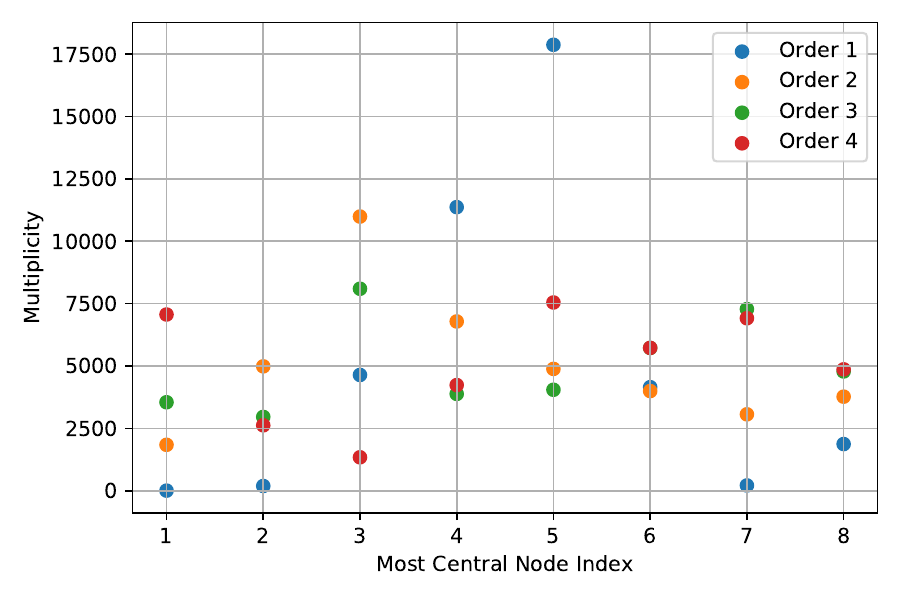}
        \caption{$E_8$ Subinvariants}
    \end{subfigure}
    \caption{The multiplicities that each of the 8 graph nodes (ie. simple roots $\alpha_i$) exists as the most central node in a bivector subinvariant graph, for all graphs across all invariant orders for each of the considered root systems: $A_8$, $D_8$, $E_8$ respectively.}\label{BivectorMostCentralNode}
\end{figure}

\begin{figure}[t]
    \centering
    \begin{subfigure}{0.32\textwidth}
        \centering
        \includegraphics[width=0.99\textwidth]{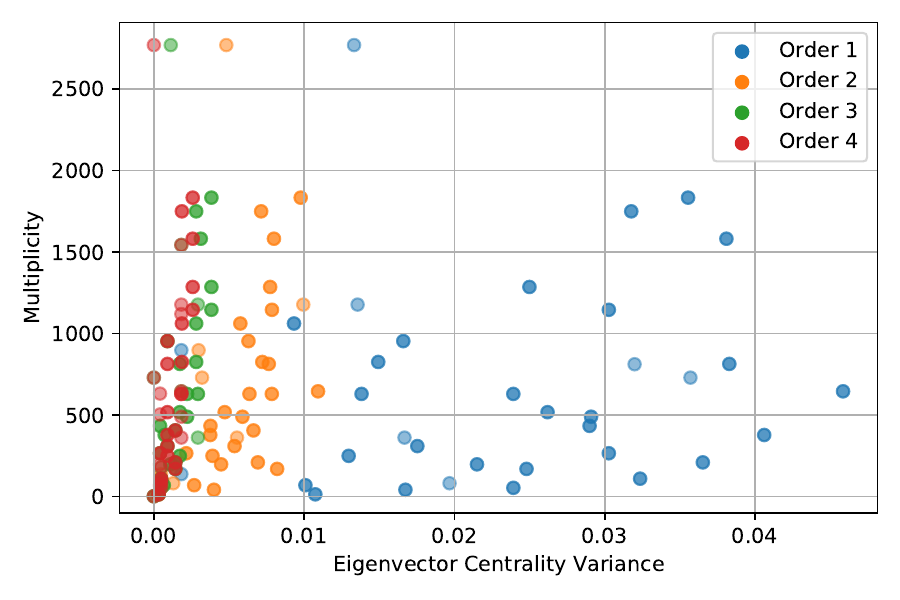}
        \caption{$A_8$ Subinvariants}
    \end{subfigure}
    \begin{subfigure}{0.32\textwidth}
        \centering
        \includegraphics[width=0.99\textwidth]{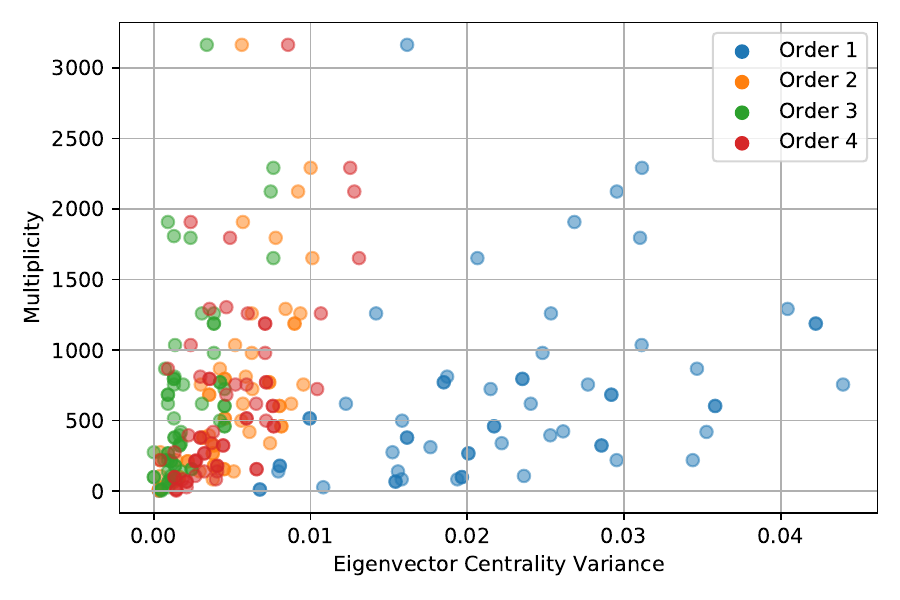}
        \caption{$D_8$ Subinvariants}
    \end{subfigure} 
    \begin{subfigure}{0.32\textwidth}
        \centering
        \includegraphics[width=0.99\textwidth]{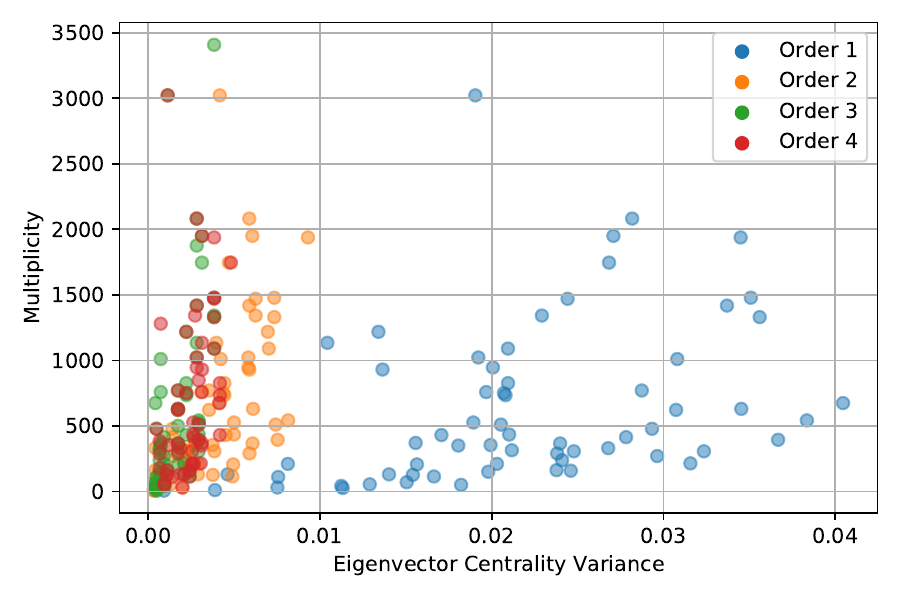}
        \caption{$E_8$ Subinvariants}
    \end{subfigure}
    \caption{The multiplicities of the variances across the distribution of eigenvector centrality scores for each bivector subinvariant graph. These variances were computed for each graph across all invariant orders for each of the considered root systems: $A_8, D_8, E_8$ respectively.}\label{BivectorCentralityVariances}
\end{figure}

To examine this behaviour in the root systems considered, the eigenvector corresponding to the previously studied largest eigenvalue was computed for each bivector subinvariant across all invariant orders for each root system.
The node index of the most central node was then identified, and the variance of the centrality measure distribution calculated. 
The multiplicity distributions of these centrality distribution measures are shown in Figure \ref{BivectorMostCentralNode} for the node index of the most central node, and Figure \ref{BivectorCentralityVariances} for the variance in centrality scores across each bivector subinvariant.

The results in Figure \ref{BivectorMostCentralNode} show some consistency of which nodes are the most central between the root systems.
In all cases the order 1  invariants ($\Inv_1$) have the most skewed distributions, with the first 2 nodes never being the most central, whilst the last nodes are the most central infrequently\footnote{We note here that graph nodes have no intrinsic order; the one chosen here matches the basis order.}. 
From this one can deduce that the basis invariants $\{a_4,a_5,a_6\}$ are most significant to the $\Inv_1$'s graph's connectivity, as in our labelling of the simple roots $\{a_1,a_2,a_3\}$ are the start of a long string, and are thus somewhat less central.
But the relative multiplicities between the nodes can be used to differentiate the root systems.
Within the set of each root systems's $\Inv_1$ invariants, the Dynkin diagrams themselves are included as graphs.
It is hence not surprising to see the $D_8$ and $E_8$ most central nodes have indeed maximum multiplicity as the most central node for $a_6$ and $a_5$ respectively, where the Dynkin diagram nodes have degree 3.
Equivalently the $A_8$ $\Inv_1$ invariants have a more symmetric distribution of most central node, with the highest multiplicity for $a_4$ (and with some numerical differences, $a_5$) matching the $A_8$ Dynkin diagram.
These results corroborate nicely the similarity of the graphs within each order and our earlier assertion that the graphs for higher order invariants `become more connected'.

Considering the higher order bivector subinvariant graphs, the range of multiplicities is noticeably lower.
However, range does not strictly decrease as order increases.
Order 2 graphs have a similarly noticeable skew towards more nodes closer to the middle of the basis order being more central, which either does not occur or is not significant enough to conclude for orders 3 and 4.
For $A_8$ and $E_8$ the order 3 and 4 graphs have a somewhat consistent distribution of which node is the most central.

In a similar manner, the variances of the centrality measures shown in Figure \ref{BivectorCentralityVariances} are larger for the order 1 bivector subinvariants $\Inv_1^2$, extending the analysis of Figure \ref{BivectorMostCentralNode} to the consideration of all the nodes' centrality scores (not just the most central one).
The overlap in variance values for the higher orders indicates that their respective graphs have similar connectivity properties, although there are potential bounds which could separate some order 2  invariants ($\Inv_2$), particularly for the $A_8$ and $E_8$ algebras.

Overall the analysis of the bivector subinvariant graphs' eigenvector centralities indicates that the order 1 graphs are distinctly different to those coming from higher order invariants. 
The $\Inv_1$ invariants tend to be more consistently structured (with the same basis elements creating the most central node) and more skewed in centrality with central nodes more dominantly central. Generally, going to higher-order invariants increases the connectivity, at least for $A_8$ and $E_8$, with some reasonably-well separated clustering of the orders. For $D_8$, the $\Inv_3$ invariants seem the most connected; this is likely due to the unique geometry of $D_8$, manifested e.g. also by the non-trivially zero scalar and pseudoscalar terms in Table \ref{table:D8}.

\section{Machine Learning}\label{sec_ML}
Statistical methods have always held a strong footing in the realm of exploratory mathematical analysis.
They are particularly useful for identifying patterns, which in turn lead to uncovering true mathematical structures, and guiding conjectures into proven theorems.

Only in the last few decades have computational resources seen such explosionary growth that many resource-heavy statistical methods have become feasible to implement. 
This has allowed the development and application of large-scale computational statistical methods, a range of techniques known commonly as \textit{machine learning} (ML).

ML methods have been used broadly across a diverse range of fields, with outstanding and surprising success.
Within our area of mathematics and mathematical physics research, the plethora of ML techniques have been largely unutilised, presenting many opportunities for new application and insight.
A selection of example successes of ML methods on mathematical data include: cluster algebras \cite{dechant2022cluster,cheung2022clustering,Bao:2020nbi}, dessins d'enfants \cite{He:2020eva}, tropical geometry \cite{Bao:2021olg,Chen:2022jwd}, knots \cite{Gukov:2020qaj}, and various string theory-inspired algebraic geometry datasets \cite{Bao:2022rup,Berglund:2021ztg,Bull:2018uow,Berman:2021mcw,Berglund:2023ztk,Bao:2021auj,Bao:2021ofk,coates2022machine,Abel:2021ddu,Niarchos:2023lot}.

\Modification{ML as a field is subdivided into three core categories,
: \textit{supervised}, \textit{unsupervised} and \textit{reinforcement} learning. The focus of this study is on the first two.} In \textit{supervised} learning, many-parameter functions are fitted to large datasets of (input, output) pairs.
In \textit{unsupervised} learning there is no respective output data, and traditional data analysis methods for feature extraction and clustering are applied directly to the datasets.

\subsection{Supervised: Neural Networks}
Supervised ML concerns itself with learning functions which map inputs to outputs.
The most common supervised architectures for this, largely due to their versatility as universal approximators \cite{Hornik1989MultilayerFN,LESHNO1993861}, are \textit{neural networks} (NNs).
NNs are built from layers of neurons, where the action at each neuron starts with an input vector $\textbf{x}$, which is acted upon linearly by weights $\textbf{W}$ and biases $b$, whose resulting number is then passed to a non-linear `activation' function $a$, producing the neuron output.
Overall this action looks like $neuron: \textbf{x} \longmapsto a(\textbf{W}\cdot\textbf{x}+b)$.
These neurons are arranged in layers such that the input is passed to each neuron in the first layer; their  outputs are then compounded into a vector to pass to the next layer etc, which cumulatively build up to a non-linear function\footnote{Note this architecture is the most general in form, known as fully-connected or dense. One may restrict which neurons receive which outputs from the previous layer in a variety of systematic ways to design more specific architectures.}.

In order to approximate a given or implied non-linear function relating the input to the output, the weights of a neural network can be adjusted, or `trained', using data and an optimisation algorithm (with respect to a `loss function'), in order to provide  an in some sense `optimal' approximation to  the original potentially highly complex non-linear function relating input and output \cite{anderson1995introduction}.
Given a dataset to learn with NNs, the data is first partitioned into training and test subsets.
The train data is fed into the NN in batches, and an optimiser then updates the NN parameters of weights and biases to minimise the specified loss function over this batch. 
This process is then iterated for each batch over the training dataset, then repeated for as many epochs as specified.
The unseen test dataset inputs are then passed into the trained NN to predict outputs, which are compared to the true outputs with various learning measures to assess performance. 
The entire process may be repeated $k$ times in $k$-fold cross-validation, with $k$ different partitions into train and test datasets to provide $k$ output performance measures, which can be averaged to provide statistical confidence.

\subsubsection{Binary Classification of Invariants: real vs fake data}\label{sec:bc_invariants}\mbox{}\\
It is expected that geometric invariants can be constructed, via some formula, from Coxeter elements that carry information about the root system. So one might hope to find specific features of invariants that depend on the root system/Lie algebra. 

The dataset used consists of 3 subsets, one for each of the algebras $A_8$/$E_8$/$D_8$, with 40320 entries and 2304 components in each subset. To further enlarge the amount of data, we generated a `fake' dataset for each of the $A_8$/$E_8$/$D_8$ algebra. In the first iteration, `fake' datasets were generated by constructing empirical distributions for each of the 2304 components from all available invariants for the corresponding algebra and then sampling from this distribution to create 40,000 unique elements in new datasets. One can notice that in the `real' data, the number of zeros in each entry is constant and specific for different algebras \footnote{1942 for $E_8$, 2083 for $D_8$ and 1805 for $A_8$.}. We implemented this feature in fake datasets we used by eliminating fake data entries which do not satisfy this condition.


We started with one of the simplest supervised learning approaches, binary classification by a dense neural network with an idea to train 3 NNs to distinguish invariants for one of the algebras from other algebras and `fake' invariants. Overall, we have 6 datasets of geometric invariant components for $A_8$, $D_8$, and $E_8$ algebras, as well as `fake' datasets for each. 
From these, three final training/test datasets were constructed, one for each algebra, where we labelled `real' $A_8$ or $D_8$ or $E_8$ algebra invariant components with 1, two remaining `real' algebras invariant components with 0, and 3 `fake' datasets labelled as 0 as well. This is what we will imply when we say that we create training/test datasets to distinguish one of the $A_8$, $D_8$, and $E_8$ invariants from other `real' invariants and the `fake' ones.

\Modification{One should notice however that training NNs to distinguish one of the $A_8/D_8/E_8$ invariants from others and fake invariants using all datasets is invalid. For all three of them, the prediction on test datasets would be 100\% accurate because, as described in section \ref{sec:freq_anal}, there are only 128 unique invariants for each of the ADE datasets, leading to a large repetition of them (although with unequal frequencies). Hence, there is a high chance that in a randomly chosen training subset, we would find all 128 unique invariants. This would make the test stage biased, as it likely contains repetitions of the training data.}
Effectively, these NNs are learning to reproduce this dataset of invariants perfectly, but would not be able to generalise beyond it.

A more meaningful problem is to remove degeneracy in datasets for `real' $A_8/D_8/E_8$ invariants, leaving just 128 elements in each. Then, we can mix original and fake invariants, which ensures that we have some real invariants in the training set and others in the test set only. The proportion of data in training and test datasets was again set to 80\% and 20\%. However, this makes the whole dataset skewed as there are around 40000 fake invariants and only $3 \times 128$ real invariants. We kept data unbalanced in the training set, but in the test set, to make it easier to interpret results, we cut the number of fake invariants to be the same as the number of real invariants, meaning we had  $3 \times \{128/k$ real and $128/k$ fake\} invariants. 

In this setup, we tried NN with architectures varying from 1 hidden layer with 32 units to 2 hidden layers with 256 units in each. \Modification{In all of them, the ReLU activation function was used. During training, we used the Adam optimizer (with a learning rate of 0.001) to optimize the log-loss function. The train and test datasets represented an 80\% / 20\% split of the total data.} The best performance was demonstrated by 1 hidden layer 64, 128 and 256 units NNs with accuracies in the range of 0.90-0.92. From this one might speculate that there should be some relatively simple invariant quantity (similar to the genus of a surface) that was learned by the NNs to distinguish real invariants coming from different algebras and fake invariants.

\subsubsection{Regressing Invariants from Permutations}\label{sec:regresssubinvariants}\mbox{}\\
As introduced in Section \ref{sec_back}, Clifford algebras and the simplicial derivatives/characteristic multivectors provide us with a systematic way of computing the geometric invariants (SOCM) occurring e.g. in the Cayley-Hamilton theorem. In particular, this depends on the permutation order of the simple reflections in a Coxeter element. It is, therefore, expected that there is an analytical formula that directly predicts the corresponding geometric invariant from the order of the permutation of the simple reflections. Since making conjectures or derivations from scratch is challenging in this task, in the spirit of experimental mathematics, we hope that the use of supervised learning algorithms, which train on labelled datasets to make predictions, can shed some light on this expected relation.

In this paper, we used dense NNs coded in \texttt{python} where the input is the (one-hot encoded \cite{anderson1995introduction}) permutation and the output is the coefficients of the invariants. For clearer results, we first partitioned each dataset for $A_8,D_8,E_8$ into 9 subdatasets for each of the 9 invariant orders, and then look at the coefficients of both the full invariant and each subinvariant for each order subdataset, i.e., the scalar, bivector, quadrivector, sextivector, and pseudoscalar for each invariant order $\Inv_0$-$\Inv_8$ (SOCM). The NN model includes four dense layers of 256 units with ReLU activation function. In training, we used the Adam optimizer (with learning rate 0.001) to minimise a mean-squared error loss. 5-fold cross validation was also used, with test data subset being $20\%$ of the full dataset. To calculate accuracy, predictions were rounded to the nearest integer, and a prediction was considered correct if all of its coefficients were predicted correctly after rounding. 

Our ML results are summarised in Tables \ref{tab:NN A8} to \ref{tab:NN E8} for the $A_8,D_8,E_8$ data in the simple root basis. The results show near-perfect prediction of all invariants and subinvariants across all algebras.
The trivial scalar invariants are unsurprisingly all learnt perfectly, but in many other cases there is perfect learning also.
The lowest performance occurs for the sextivectors, where there is less data to learn from.
These results indicate that the NNs are capable of well approximating the complicated algorithm carried out to compute these invariants, as well as accommodating for the basis permutations.

\begin{table}[!ht]
    \centering
    \begin{tabular}{|c||>{\centering\arraybackslash}m{5.1em}||>{\centering\arraybackslash}m{5.1em}|>{\centering\arraybackslash}m{5.1em}|>{\centering\arraybackslash}m{5.1em}|>{\centering\arraybackslash}m{5.1em}|>{\centering\arraybackslash}m{5.1em}|}
    \hline
    & Acc($\Inv_i$) & Acc($\Inv_i^0$) & Acc($\Inv_i^2$) & Acc($\Inv_i^4$) & Acc($\Inv_i^6$) & Acc($\Inv_i^8$)\\
    \hline
    $\Inv_0$ &\small{\(1.0000\)}&\small{\(1.0000\)} & & & &  \\
    \hline
    $\Inv_1$ &\small{\(1.0000\)}&\small{\(1.0000\)} &\small{\(0.9996\)}& & & \\
    \hline
    $\Inv_2$ &\small{\(0.9996\)}&\small{ \(1.0000\)}&\small{\(0.9999\)} & \small{$0.9999$} & & \\
    \hline
    $\Inv_3$ &\small{\(0.9980\)}&\small{\(1.0000\)} &\small{$0.9973$} &\small{$0.9480$} & \small{$0.9999$} & \\
    \hline
    $\Inv_4$ &\small{\(0.9958\)}&\small{$1.0000$ }&  \small{$0.9999$}& \small{$0.9999$} & \small{$0.9117$}& \small{$0.9596$}\\
    \hline
    $\Inv_5$ &\small{\(0.9986\)}& \small{$1.0000$} & \small{$0.9995$}& \small{$0.9999$} & \small{$1.0000$} & \\
    \hline
    $\Inv_6$ &\small{\(1.0000\)}& \small{$1.0000$} &  \small{$0.9948$}  & \small{$0.9999$} & &\\
    \hline
    $\Inv_7$ &\small{\(0.9999\)}&\small{ $1.0000$}& \small{$1.0000$} & & &\\
    \hline
    $\Inv_8$ &\small{\(1.0000\)}& \small{$1.0000$}& & & & \\
    \hline
    \end{tabular}
    \caption{Summary of the final test accuracy (Acc) for the full invariants and each subinvariant of the 9 invariants for $A_8$ simple root data.}
    \label{tab:NN A8}
\medskip
    \begin{tabular}{|c||>{\centering\arraybackslash}m{5.1em}||>{\centering\arraybackslash}m{5.1em}|>{\centering\arraybackslash}m{5.1em}|>{\centering\arraybackslash}m{5.1em}|>{\centering\arraybackslash}m{5.1em}|>{\centering\arraybackslash}m{5.1em}|}
    \hline
    & Acc($\Inv_i$) & Acc($\Inv_i^0$) & Acc($\Inv_i^2$) & Acc($\Inv_i^4$) & Acc($\Inv_i^6$) & Acc($\Inv_i^8$)\\
    \hline
    $\Inv_0$ &\small{\(1.0000\)} &\small{\(1.0000\)} & & & &  \\
    \hline
    $\Inv_1$ &\small{\(1.0000\)} &\small{\(1.0000\)} &\small{\(0.9955\)}& & & \\
    \hline
    $\Inv_2$ &\small{\(1.0000\)} &\small{ \(1.0000\)}&\small{\(0.9912\)} & \small{$0.9999$} & & \\
    \hline
    $\Inv_3$ &\small{$0.9993$} &\small{\(1.0000\)} &\small{$0.9995$} &\small{$0.9999$} & \small{$1.0000$} & \\
    \hline
    $\Inv_4$ &\small{$0.9995$}&\small{$1.0000$ }&  \small{$0.9988$}& \small{$0.9891$} & \small{$0.9998$}& \small{$1.0000$}\\
    \hline
    $\Inv_5$ &\small{$0.9995$} & \small{$1.0000$} & \small{$0.9986$}& \small{$1.0000$} & \small{$1.0000$} & \\
    \hline
    $\Inv_6$ & \small{$1.0000$}& \small{$1.0000$} &  \small{$1.0000$}  & \small{$1.0000$} & &\\
    \hline
    $\Inv_7$ &\small{ $1.0000$}&\small{ $1.0000$}& \small{$0.9999$} & & &\\
    \hline
    $\Inv_8$ &\small{ $1.0000$} & \small{$1.0000$}& & & & \\
    \hline
    \end{tabular}
    \caption{Summary of the final test accuracy (Acc) for the full invariants and each subinvariant of the 9 invariants for $D_8$ simple root data.}
    \label{tab:NN D8}
\medskip
    \begin{tabular}{|c||>{\centering\arraybackslash}m{5.1em}||>{\centering\arraybackslash}m{5.1em}|>{\centering\arraybackslash}m{5.1em}|>{\centering\arraybackslash}m{5.1em}|>{\centering\arraybackslash}m{5.1em}|>{\centering\arraybackslash}m{5.1em}|}
    \hline
    & Acc($\Inv_i$) & Acc($\Inv_i^0$) & Acc($\Inv_i^2$) & Acc($\Inv_i^4$) & Acc($\Inv_i^6$) & Acc($\Inv_i^8$)\\
    \hline
    $\Inv_0$ &\small{\(1.0000\)} &\small{\(1.0000\)} & & & &  \\
    \hline
    $\Inv_1$ &\small{\(1.0000\)} &\small{\(1.0000\)} &\small{\(1.0000\)}& & & \\
    \hline
    $\Inv_2$ &\small{$0.9999$}&\small{ \(1.0000\)}&\small{\(1.0000\)} & \small{$1.0000$} & & \\
    \hline
    $\Inv_3$ &\small{$0.9994$}&\small{\(1.0000\)} &\small{$0.9906$} &\small{$0.9999$} & \small{$0.9891$} & \\
    \hline
    $\Inv_4$ &\small{$0.9969$}&\small{$1.0000$ }&  \small{$1.0000$}& \small{$0.9963$} & \small{$0.9793$}& \small{$0.9223$}\\
    \hline
    $\Inv_5$ &\small{$0.9994$} & \small{$1.0000$} & \small{$0.9996$}& \small{$0.9999$} & \small{$0.9990$} & \\
    \hline
    $\Inv_6$ & \small{$1.0000$} & \small{$1.0000$} &  \small{$1.0000$}  & \small{$1.0000$} & &\\
    \hline
    $\Inv_7$ & \small{$1.0000$} &\small{ $1.0000$}& \small{$0.9998$} & & &\\
    \hline
    $\Inv_8$ & \small{$1.0000$} & \small{$1.0000$}& & & & \\
    \hline
    \end{tabular}
    \caption{Summary of the final test accuracy (Acc) for the full invariants and each subinvariant of the 9 invariants for $E_8$ simple root data.}
    \label{tab:NN E8}
\end{table}

\subsubsection{Gradient Saliency Analysis}
\begin{figure}[t]
    \centering
    \begin{subfigure}{0.19\textwidth}
        \centering
        \includegraphics[width=0.99\textwidth]{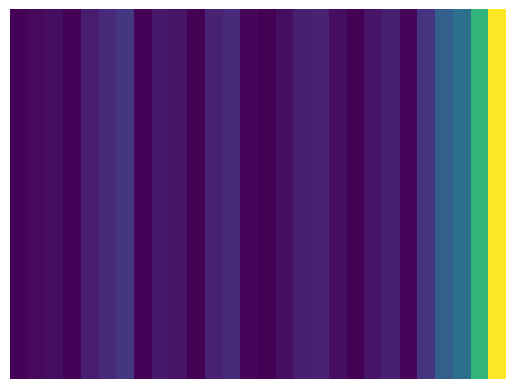}
        \caption{$\Inv_1^2$}
    \end{subfigure}
    \begin{subfigure}{0.19\textwidth}
        \centering
        \includegraphics[width=0.99\textwidth]{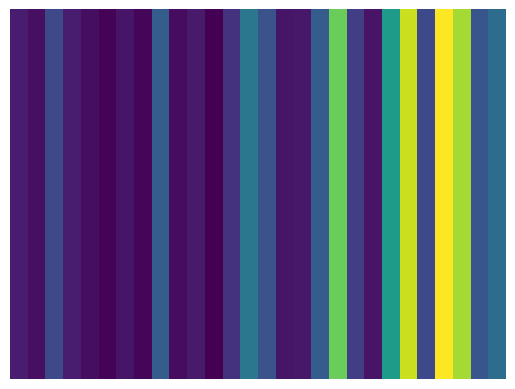}
        \caption{$\Inv_2^2$ }
    \end{subfigure} 
    \begin{subfigure}{0.19\textwidth}
        \centering
        \includegraphics[width=0.99\textwidth]{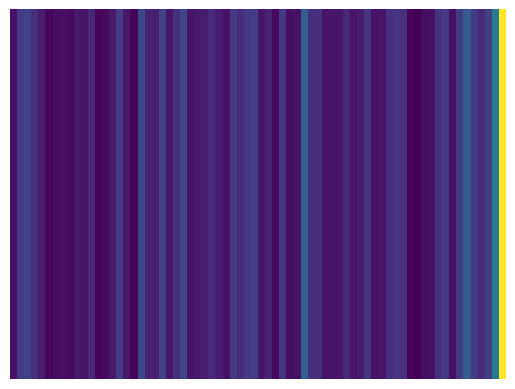}
        \caption{\!$\Inv_2^4$\! }
    \end{subfigure}
    \begin{subfigure}{0.19\textwidth}
        \centering
        \includegraphics[width=0.99\textwidth]{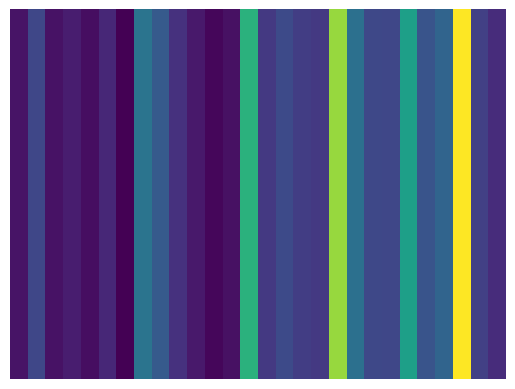}
        \caption{$\Inv_3^2$ }
    \end{subfigure}
    \begin{subfigure}{0.19\textwidth}
        \centering
        \includegraphics[width=0.99\textwidth]{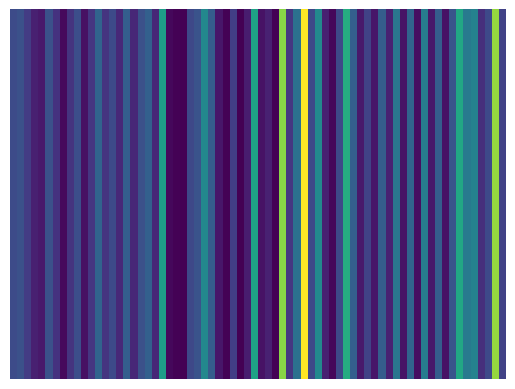}
        \caption{\!$\Inv_3^4$\!}
    \end{subfigure} 
    \begin{subfigure}{0.19\textwidth}
        \centering
        \includegraphics[width=0.99\textwidth]{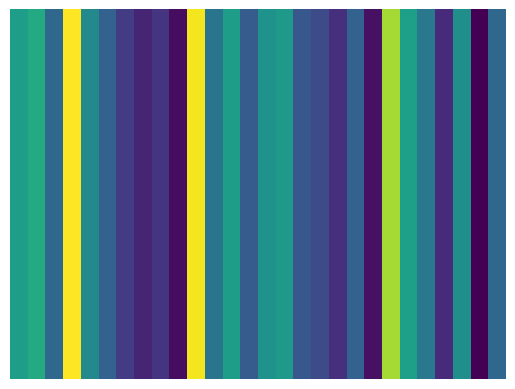}
        \caption{$\Inv_3^6$ }
    \end{subfigure}
    \begin{subfigure}{0.19\textwidth}
        \centering
        \includegraphics[width=0.99\textwidth]{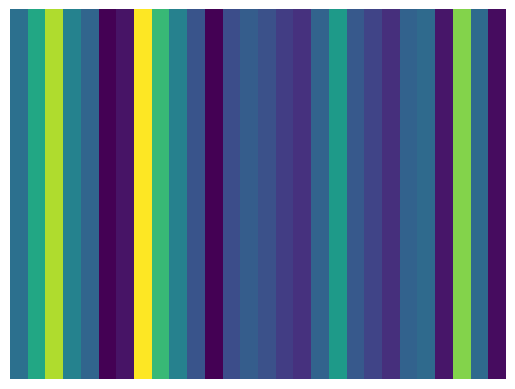}
        \caption{$\Inv_4^2$ }
    \end{subfigure}
    \begin{subfigure}{0.19\textwidth}
        \centering
        \includegraphics[width=0.99\textwidth]{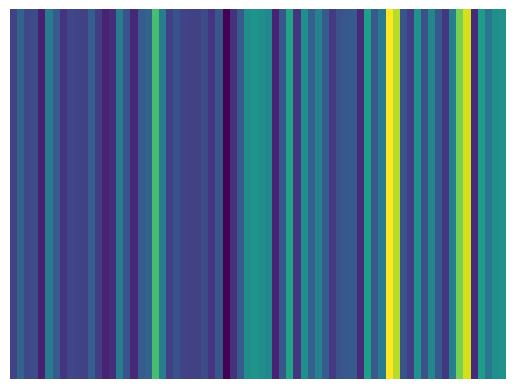}
        \caption{\!$\Inv_4^4$\! }
    \end{subfigure} 
    \begin{subfigure}{0.19\textwidth}
        \centering
        \includegraphics[width=0.99\textwidth]{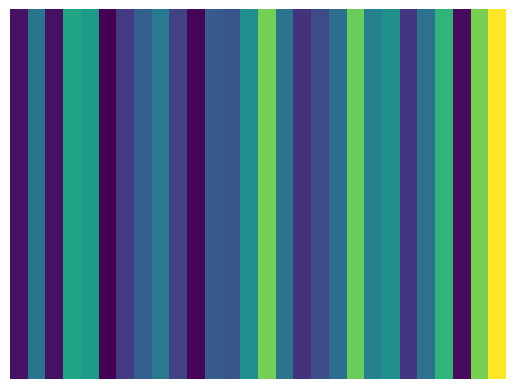}
        \caption{$\Inv_4^6$ }
    \end{subfigure}
    \begin{subfigure}{0.19\textwidth}
        \centering
        \includegraphics[width=0.99\textwidth]{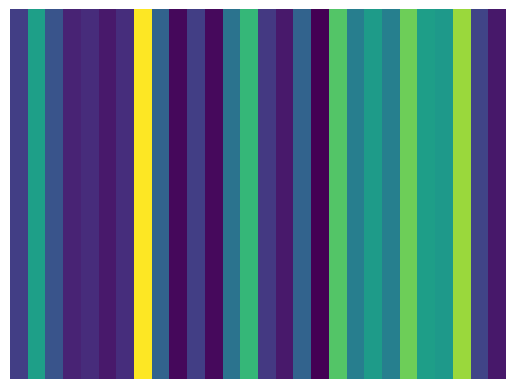}
        \caption{$\Inv_5^2$}
    \end{subfigure} 
    \begin{subfigure}{0.19\textwidth}
        \centering
        \includegraphics[width=0.99\textwidth]{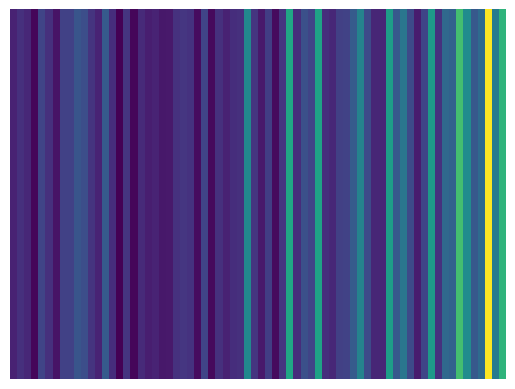}
        \caption{\!$\Inv_5^4$\!}
    \end{subfigure}
    \begin{subfigure}{0.19\textwidth}
        \centering
        \includegraphics[width=0.99\textwidth]{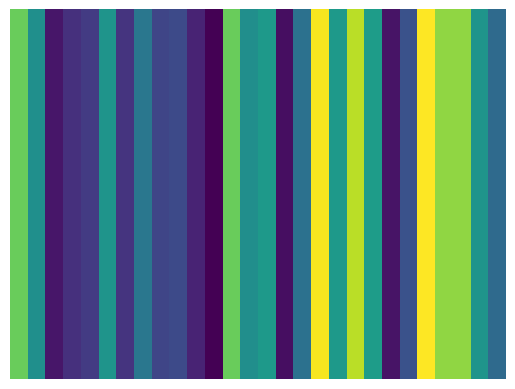}
        \caption{$\Inv_5^6$ }
    \end{subfigure} 
    \begin{subfigure}{0.19\textwidth}
        \centering
        \includegraphics[width=0.99\textwidth]{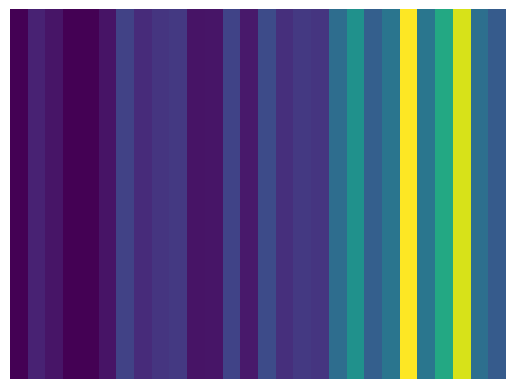}
        \caption{$\Inv_6^2$}
    \end{subfigure}
    \begin{subfigure}{0.19\textwidth}
        \centering
        \includegraphics[width=0.99\textwidth]{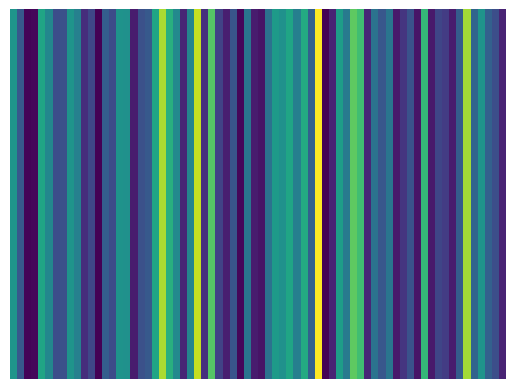}
        \caption{\!$\Inv_6^4$\!}
    \end{subfigure} 
    \begin{subfigure}{0.19\textwidth}
        \centering
        \includegraphics[width=0.99\textwidth]{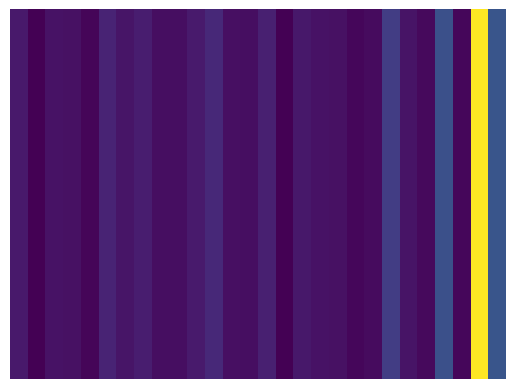}
        \caption{$\Inv_7^2$}
    \end{subfigure}
    \caption{Gradient saliency maps (barcodes) for ternary classification NN model. The NN function takes as input the coefficients of the subinvariant and outputs one-hot encoded: $A_8$, $D_8$, or $E_8$. The saliency hence represents the relative importance of each input coefficient (i.e. combination of simple roots) to determining the classification output. Lighter colours indicate larger gradients and greater importance.}\label{fig: GS NN classification}
\end{figure}

\mbox{}\\
To better interpret the decision-making of our NN models -- which are black-box models -- we also performed gradient saliency analysis \cite{sundararajan2017axiomatic}. In general, the magnitude of the elements in a weight vector in the model tells us the importance of the corresponding input element for a particular output. 
We can extend this to consider the sensitivity of the entire NN function to the inputs by computing the gradient of a given output with respect to the input via backpropagation. The magnitude of the gradient indicates how sensitive the output is to a change in the input variable. The results for the average gradient magnitudes across the test sets (and 100 cross-validation runs) are shown in Figure \ref{fig: GS NN classification} for the multiclassification investigation equivalent to Section \ref{sec:bc_invariants} but without considering the fake data (performance was equivalently perfect), and Tables \ref{tab:GSA8regression}, \ref{tab:GSD8regression}, \ref{tab:GSE8regression} in appendix \ref{app_GS} for the subinvariant regression in Section \ref{sec:regresssubinvariants}.

The Figure \ref{fig: GS NN classification} gradient saliency barcode maps show the relative importance of the subinvariant inputs for algebra classification; hence the bivector $\tiny{\begin{pmatrix} 8 \\ 2 \end{pmatrix}} = 28$ ($= \tiny{\begin{pmatrix} 8 \\ 6 \end{pmatrix}}$ for the sextivector also) coefficients are represented by 28 vertical lines, whilst the quadrivector $\tiny{\begin{pmatrix} 8 \\ 4 \end{pmatrix}} = 70$ coefficients are represented by 70 vertical lines.
Note the trivial scalar as well as the pseudoscalar subinvariants are omitted, and the remaining plots approximately satisfy the observed mirror symmetry between invariant orders (i.e. \(\Inv_1^2\) $\sim$ \(\Inv_7^2\), \(\Inv_3^4\) $\sim$ \(\Inv_5^4\), etc.).

For \(\Inv_1^2\) and \(\Inv_7^2\), the NNs rely almost exclusively on the final components of the coefficient vector, with a similar skewed behaviour for  \(\Inv_2^2\) and  \(\Inv_6^2\) towards the final coefficients implying the span of these final coefficient values is most disparate between the algebras and thus can be used for classification.
The  \(\Inv_3^2,\Inv_4^2,\Inv_5^2\) and  \(\Inv_3^6,\Inv_4^6,\Inv_5^6\) barcodes have less discernible patterns, but in each case do appear to prioritise specific coefficient entries, emphasising that the entries in the subinvariant coefficients vectors are not equally important as one may naively assume.
The  \(\Inv_2\) (and  \(\Inv_5\)) quadrivector maintains the \(\Inv_2\) bivector bias towards reliance on the final coefficient entries, however the other quadrivector barcodes have a smoother spread of importance and more complicated (NN-approximate) function differentiable structure.

These results provide insight into the relative importance of each coefficient to uniquely identifying the respective algebra, and continues to guide our analytical study of these subinvariants in our companion paper by indicating which subinvariants have the most clear coefficient dependence (\(\Inv_1\) and \(\Inv_7\) bivector, and \(\Inv_2\) quadrivector) for us to focus on, in revealing the underlying behaviour of these invariants.

In Tables \ref{tab:GSA8regression}, \ref{tab:GSD8regression}, \ref{tab:GSE8regression}, the saliency barcodes probe the NN learning of the subinvariants explicitly from the Coxeter element root permutation (i.e. the order of the 8 roots in the Coxeter element) -- hence having 8 bars in each barcode.
Satisfyingly, the mirror symmetry between opposite order invariants is again approximately obeyed for each of the 3 algebras considered.

The trivial order 0 and 8 full invariants (with only scalar part such that they start with a single 1 followed by 255 0s), as well as all the scalar invariants, are as expected perfectly learned and have random saliency behaviour since the learning of a constant function is trivial and independent of the inputs. 
One may expect perfectly equal barcodes, but the rounding of the outputs allows the final neuron output to vary, so the stochastic search of the optimiser has a large range of functions which give perfect results that it will be random walking in the function space throughout the training. 
This makes the final function fairly arbitrary in this class of suitable functions and hence the barcodes random, providing some measure of the level of noise in the learning.

Conversely the pseudoscalar for the order 4 invariants has different saliency properties between the algebras, with focus on different parts of the permutation vector; where the $A_8$, $D_8$, $E_8$ order 4 invariant pseudoscalars respectively are computed primarily from the end, middle, start of the permutation vector.
Note that the pseudoscalar component is 0 for $D_8$ and therefore that barcode is essentially noise, alike the scalar barcodes.

The remaining barcodes all have similar behaviour across the algebras.
This is for the bivector, quadrivector and sextivector subinvariants, which dominate the full invariant coefficient vector and thus unsurprisingly lead to similar behaviour for the full invariant.
This behaviour put focus on both ends of the permutation barcodes, indicating the information about which roots are positioned at the ends of the Coxeter element is the most important for determining the structure of each of these respective subinvariants (as well as the full invariant).

Intuition one may extract from this is that at the ends of the permutation vectors the roots have only one direction they can be permuted, and thus the root they are adjacent to is paramount to determining whether that root can commute further into the permutation vector without changing the Coxeter element; as dictated by whether the roots are connected in the Dynkin diagram.
How the 40320 permutation orders split into the 128 Coxeter elements for each algebra may then be well correlated with the end roots of the permutation, providing the NNs with important primary information in directing the first steps of their functional algorithm for information flow through their architecture, leading to correct calculations of the invariants.
Further analytic analysis with focus on permutation partitions grouped by their end roots should hope to reveal the dominant factors for the distributions of these invariants.

\subsection{Unsupervised: PCA}
Principal component analysis (PCA) is a widely used machine learning technique for dimensionality reduction and exploratory data analysis \cite{Jolliffe_2016}. 
In short, one computes the principal components, which are linear combinations of the initial variables, and performs a change of basis on the data. One then projects the data onto only the first few principal components to obtain a lower-dimensional data representation.

The first principal component is the normalised linear combination of initial variables that explains the largest variance in the data. The second principal component is uncorrelated with (i.e. perpendicular to) the first principal component and explains the next highest variance, and so on.
The computation of the principal components can be broken down into the following steps:
\begin{enumerate}
    \item The covariance matrix is computed. This is a symmetric matrix whose entries are the covariances associated with all possible pairs of variables.
    \item The next step is to compute the eigenvectors and eigenvalues of the covariance matrix. These eigenvectors are the principal components, and the eigenvalues describe the amount of variance carried in each principal component.
\end{enumerate}
Note that often in general data science an extra step 0 is included whereby one standardises the variables so that each contributes equally, which is especially important when different input features have different units of measurement.
However since all variables are unit-less and take values in a similar range, we don't standardise here. 
By ranking the eigenvectors in order of their eigenvalues, highest to lowest, one gets the principal components in order of significance.
Finally, to transform the data into the new representation one performs a change of basis on the standardised data using the principal components, followed by projection.  

\begin{figure}[t]
    \centering
    \begin{subfigure}{0.32\textwidth}
        \centering
        \includegraphics[width=0.99\textwidth]{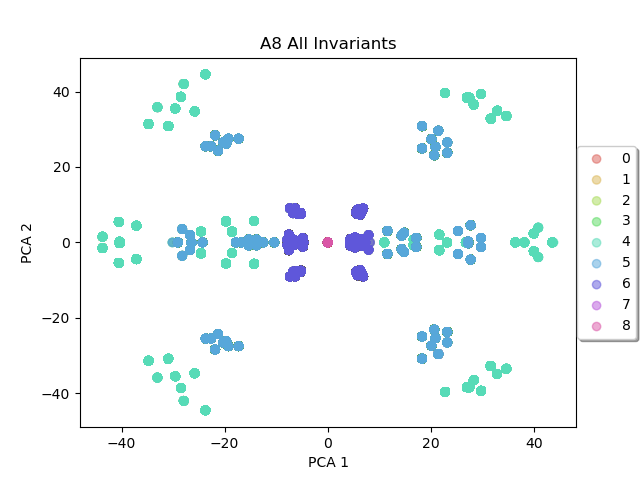}
        \caption{$A_{8}$}
    \end{subfigure}
    \begin{subfigure}{0.32\textwidth}
        \centering
        \includegraphics[width=0.99\textwidth]{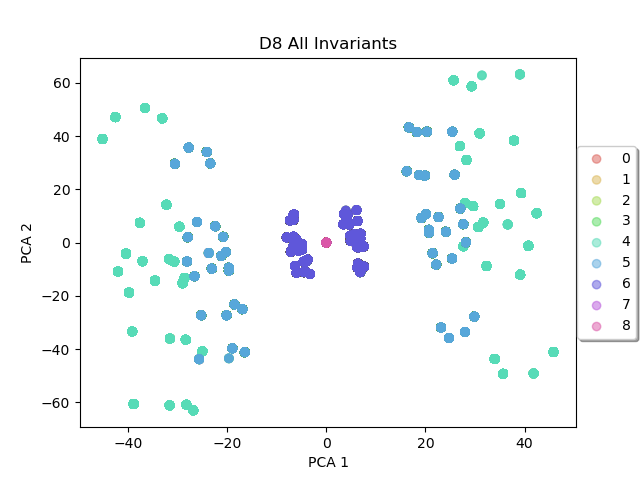}
        \caption{$D_{8}$}
    \end{subfigure} 
    \begin{subfigure}{0.32\textwidth}
        \centering
        \includegraphics[width=0.99\textwidth]{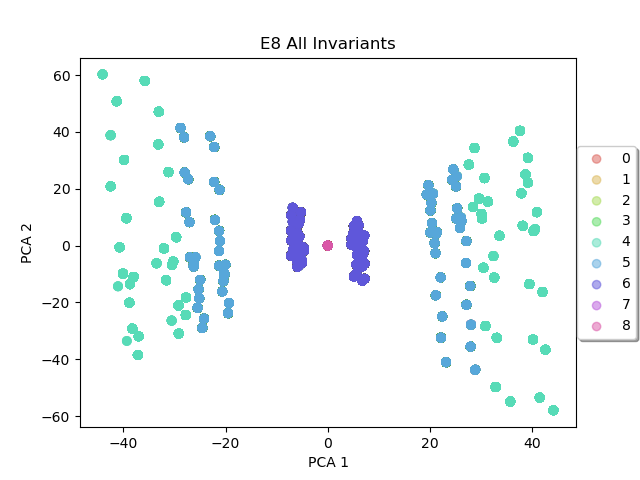}
        \caption{$E_{8}$}
    \end{subfigure}
    \caption{PCA plots of all 9 order invariants (SOCM) simultaneously for $A_{8},D_{8}$ and $E_{8}$.  Note that labels 5-8 don't appear, as these invariants are mirror symmetric. This plot and analysis are a good check of this fact. }\label{fig:PCA_all}
\end{figure}

Using the Coxeter element invariant data in the simple root basis as described in Section \ref{sec:data_gen}, we perform PCA on the 9 order invariants both individually and combined, for $A_{8},D_{8}$ and $E_{8}$. With this we plot the data in the first two principal components. 
The results for the 9 individual invariants of $A_{8},D_{8}$ and $E_{8}$ are shown in Figures \ref{fig:PCA_A8}, \ref{fig:PCA_D8} and \ref{fig:PCA_E8} respectively and the PCA results on the combined invariant data are shown in Figure \ref{fig:PCA_all}. 
We can see clearly from Figures \ref{fig:PCA_A8}-\ref{fig:PCA_E8}, that $\Inv_0$ and $\Inv_8$ just give the trivial invariant, and furthermore $\Inv_1$ matches $\Inv_7$, $\Inv_2$ matches $\Inv_6$ and $\Inv_3$ matches $\Inv_5$. 
This aligns with the connection we made earlier in Section \ref{sec_back}. 
Comparing the individual plots for $A_{8},D_{8}$ and $E_{8}$ we see that the plots for $D_{8}$ and $E_{8}$ roughly align, while the plots for $A_{8}$ are different. 
In $D_{8},E_{8}$, for example, the plots for $\Inv_3,\Inv_4$ and $\Inv_5$ share a gap in the middle and all the data points are roughly scattered on either side. 
On the other hand, the $\Inv_4$ plot for $A_{8}$ presents a circular pattern around the center, and whilst separated down the middle, the data points in the $\Inv_3$ and $\Inv_5$ plots of $A_{8}$ are tightly clustered. 
For the $A_8$ plots, there appears to be a 2-fold reflection symmetry in the $\Inv_1-\Inv_7$ plots, with the $\Inv_1$ and $\Inv_7$ plots having a second orthogonal 2-fold reflection symmetry. 
For $D_{8}$ and $E_{8}$, there is instead an approximate 2-fold rotational symmetry.

\begin{figure}[t]
    \centering
    \begin{subfigure}{0.48\textwidth}
        \centering
        \includegraphics[width=0.8\textwidth]{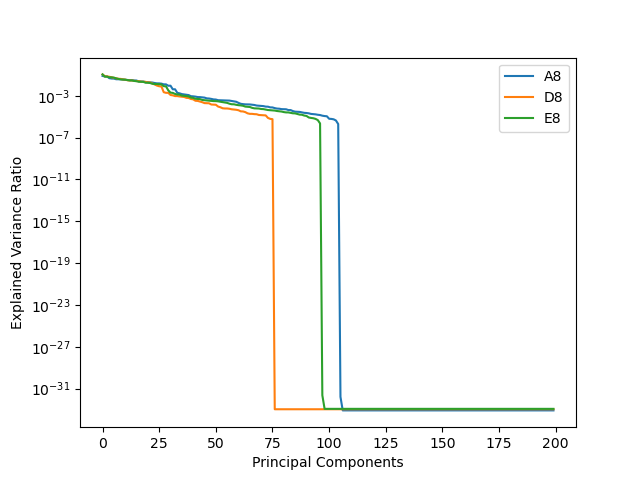}
        \caption{}
        \label{fig:elbow}
    \end{subfigure}
    \begin{subfigure}{0.48\textwidth}
        \centering
        \includegraphics[width=0.8\textwidth]{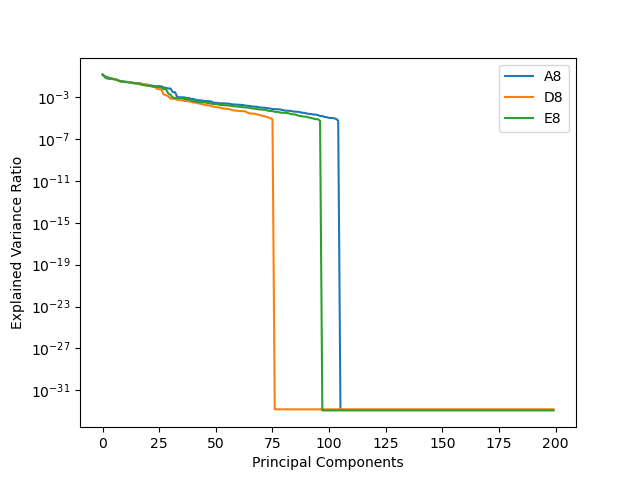}
        \caption{}
        \label{fig:elbow_unique}
    \end{subfigure} 
    \caption{Elbow plot of explained variance ratio against principal component number for (a) full $A_{8}$, $D_{8}$ and $E_{8}$ datasets and (b) the reduced $A_{8}$, $D_{8}$ and $E_{8}$ datasets with duplicates removed.}
\end{figure}

Figure \ref{fig:PCA_all} shows the 2-dimensional PCA projections when fitting the principal components for all the invariant orders considered together. 
The orders form distinct clusters and the two 2-fold reflection symmetry of $A_8$ and 2-fold rotation symmetries of $D_8$ and $E_8$ are approximately preserved. Figure \ref{fig:elbow} also shows the elbow plot of the PCA ratios against the number of principal components for PCA performed on the combined dataset of all orders. The explained variance ratio is a measure of the proportion of the total variance in the original dataset that is explained by each principal component. This is equal to the ratio of its eigenvalue to the sum of the eigenvalues of all the principal components. The $x$-axis in Figure \ref{fig:elbow} is the order number of the first principal components and the $y$-axis is log of the explained variance ratio. For $A_{8}$ and $E_{8}$ we see a characteristic sharp drop in the ratio at around the 100th principal component and for $D_{8}$ at around the 75th principal component. This means that in all cases the 256-dimensional vectors describing the invariants (of which of course only $128$ are not trivially zero) in fact can be reduced whilst preserving the majority of information. Furthermore, it is interesting that $D_{8}$ requires fewer principal components than the other two, which may be related to the vanishing pseudoscalar as well as some scalar parts in this case, and a quarter as many unique sextivector parts also as shown in Table \ref{table:D8}. 

\begin{figure}[t]
    \centering
    \begin{subfigure}{0.32\textwidth}
        \centering
        \includegraphics[width=0.99\textwidth]{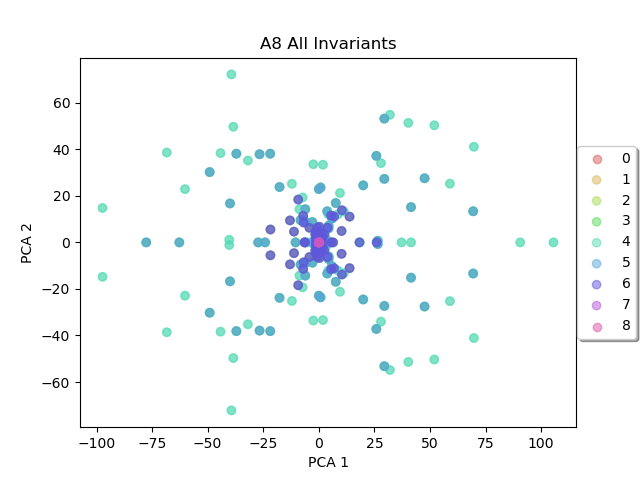}
        \caption{$A_{8}$}
    \end{subfigure}
    \begin{subfigure}{0.32\textwidth}
        \centering
        \includegraphics[width=0.99\textwidth]{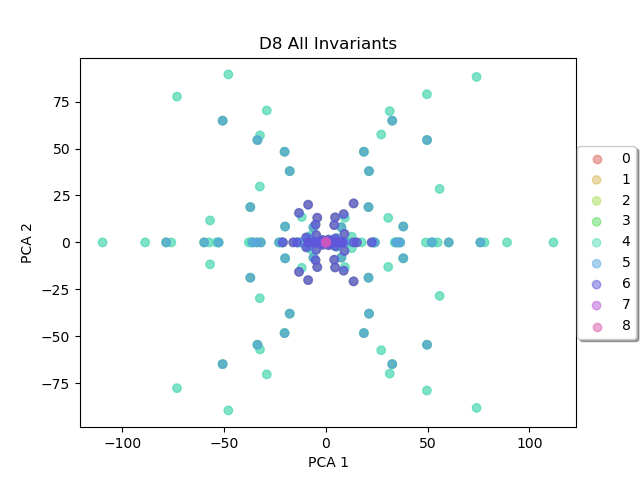}
        \caption{$D_{8}$}
    \end{subfigure} 
    \begin{subfigure}{0.32\textwidth}
        \centering
        \includegraphics[width=0.99\textwidth]{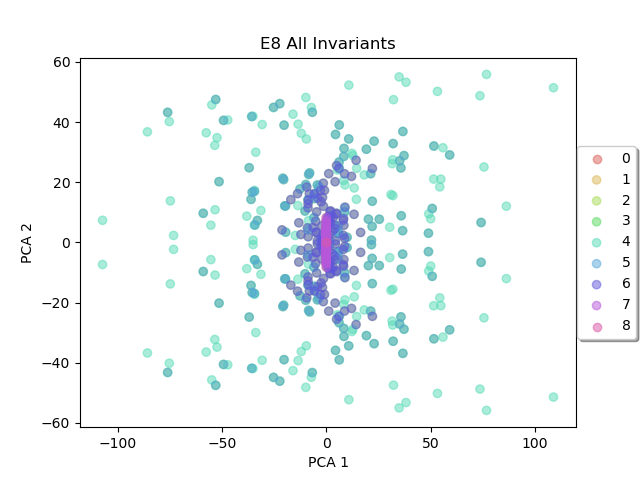}
        \caption{$E_{8}$}
    \end{subfigure}
    \caption{PCA plots of reduced datasets (with duplicates deleted) of all 9 order invariants (SOCM) simultaneously for $A_{8},D_{8}$ and $E_{8}$.}\label{fig:PCA_all_unique}
\end{figure}

As we saw in Section \ref{sec:freq_anal} the 40320 permutations give rise to only 128 unique Coxeter elements for $A_{8}$, $D_{8}$ and $E_{8}$ and the frequency of these 128 vary greatly. The frequency of invariants will have a significant effect on the PCA results and therefore, for comparison, we repeat the PCA but on the reduced dataset of 128 invariants. Again we perform the analysis on the 9 orders of invariant individually, and combined. The individual PCA plots are shown in Figures \ref{fig:PCA_A8_unique}-\ref{fig:PCA_E8_unique} and the combined plots are given in Figure \ref{fig:PCA_all_unique}. Figure \ref{fig:elbow_unique} also shows the elbow plot of the explained variance ratios for the combined PCA.

The discussed reflection and rotation symmetries of Figures \ref{fig:PCA_A8}-\ref{fig:PCA_E8} become clearer to see in Figures \ref{fig:PCA_A8_unique}-\ref{fig:PCA_E8_unique}, as presumably uneven multiplicities no longer weight the projections asymmetrically. 
Also for the $A_8$ and $D_8$ algebras, the $\Inv_4$ invariant projections become very nearly identical to the respective $\Inv_3/\Inv_5$ projections, emphasising a negligible impact of the inclusion of the pseudoscalar between these invariant orders on the first two principal components (which turns out to be $0$ for $D_8$ but not $A_8$, see Tables \ref{table:AE8} and \ref{table:D8}).
This, surprisingly, does not happen for $E_8$, indicating the pseudoscalar contribution to the principal components is more significant here. 

Whereas we saw distinct clustering of the different orders in the combined PCA plots in Figure \ref{fig:PCA_all}, we do not see this in the equivalent plots in Figure \ref{fig:PCA_all_unique} from PCA on the combined datasets with duplicates deleted. Comparing the combined plots to the unique plots for the reduced datasets we see the patterns from the unique order plots in Figures \ref{fig:PCA_A8_unique}-\ref{fig:PCA_E8_unique} emerging in the combined plots in \ref{fig:PCA_all_unique}. It appears as if all the unique plots have simply been overlaid on top of one another. This suggests that principal components for all of the 9 orders are the same and also match the principal components from the combined PCA. 

The elbow plot in Figure \ref{fig:elbow_unique} match almost identically that in Figure \ref{fig:elbow} and the same conclusions hold.

\section{Conclusions}\label{sec_concl}
This work is a pump-priming study in experimental mathematics within the field of Clifford algebras. 
This new paradigm combines an HPC computational algebra approach generating \Modification{a significant amount of algebraic} data with a data science analysis of the resulting dataset. 
Performing exhaustive calculations opens up a new angle to conjecture formulation and theorem proving, and sheds light on the new geometric invariants for the important class of examples of Coxeter transformations. 
This detailed example can be used as a foundation to explore the behaviour of invariants in other dimensions or with different types of linear transformations. 
\Modification{It was expected that there is a large degeneracy in the mapping between input permutations, resulting in a smaller number of unique elements in the dataset. This assumption was verified through the application of data analysis techniques. Moreover, many unexpected features were discovered, such as the equality of the number of unique invariants among all three algebras. The relatively small size of unique} output invariants \Modification{dataset in} this example \Modification{made it} perfectly suited to machine learning tasks, which perform very impressively. 
One can of course go to arbitrary dimension to get larger data sets for $A_n$ and $D_n$, but at the expense of missing out on the $E$-type. 
The patterns observed in the reduced set of Coxeter elements, the invariant bivectors, as well as the approach of `explainable AI' using gradient saliency, have certainly pointed in the direction of analytical results that generalise these computational observations here, some of which we have mentioned above and some we will present in the companion paper. 

\Modification{
The conjectural style of writing in this paper implies that there is still work to be done in the future. Currently, many of the statements are preliminary and have not been checked rigorously. The companion paper would aim to theoretically explain (at least some of) the features observed in this work. Some of the most interesting questions it could answer are why the number of unique invariants among three algebras is the same, how the symmetry determines the distribution of Doublets, Quadruplets and Octuplets among the invariants, what the reason for variations in clustering of the different orders in the combined PCA is, what causes the difference in the choice of important parts of the permutation vector by NN for the classification task (which was revealed by the gradient saliency analysis), and so on. Ultimately, the hope is to understand better the structure of the symmetry structures, the geometric invariants and their geometrical meaning.}


\subsection*{Acknowledgment}
This paper is dedicated to the memory of the late John McKay, Peter Neumann and Jim Humphreys. Jim's books have had profound impact on those studying Lie theory and reflection/Coxeter groups, and certainly his exposition of invariants, exponents and the Coxeter plane has p\^iqued our interest. John was a master at finding connections between different areas of Mathematics. We hope that he would approve of our interdisciplinary approach as well as the choice of ADE as examples, which is the topic of our forthcoming book with him and Peter Cameron, whose PhD advisor was Peter Neumann. \\
PPD is grateful to the London Mathematical Society for grants 42035 and 42111, which helped establish this collaboration through previous works, and to the London Institute for Mathematical Sciences for its hospitality. 
YHH would like to thank STFC for grant ST/J00037X/2.
E.~Hirst acknowledges support from Pierre Andurand over the course of this research. 
E.~Heyes is funded by School of Science and Technology - City, University of London and the States of Jersey postgraduate grant.
DR is supported by the School of Science and Technology - City, University of London and Science and Technology Facilities Council Doctoral Studentship. \\
This research utilised Queen Mary's Apocrita HPC facility, supported by QMUL Research-IT \cite{apocrita}.
Aspects of this work were undertaken on ARC3, part of the High Performance Computing facilities at the University of Leeds, UK.
For the purpose of open access, the authors have applied a Creative Commons Attribution (CC BY) licence to any Author Accepted Manuscript version arising from this submission. 

\section*{Declarations}

\subsection*{Competing interests}

YHH and PPD are Guest Editors of the ``Topical Collection: Machine-Learning Mathematical Structure''.

\bibliography{references}

\begin{thebibliography}{10}
\providecommand{\url}[1]{{#1}}
\providecommand{\urlprefix}{URL }
\expandafter\ifx\csname urlstyle\endcsname\relax
  \providecommand{\doi}[1]{DOI~\discretionary{}{}{}#1}\else
  \providecommand{\doi}{DOI~\discretionary{}{}{}\begingroup
  \urlstyle{rm}\Url}\fi

\bibitem{abdulkhaev2021explicit}
Abdulkhaev, K., Shirokov, D.: On explicit formulas for characteristic
  polynomial coefficients in {Geometric} {Algebras}.
\newblock In: Computer Graphics International Conference, pp. 670--681.
  Springer (2021)

\bibitem{Abel:2021ddu}
Abel, S., Constantin, A., Harvey, T.R., Lukas, A.: {String Model Building,
  Reinforcement Learning and Genetic Algorithms}.
\newblock In: {Nankai Symposium on Mathematical Dialogues}: {In celebration of
  S.S.Chern's 110th anniversary} (2021)

\bibitem{ablamowicz2021call}
Ab{\l}amowicz, R.: Call for papers: {TC} machine-learning mathematical
  structures.
\newblock Advances in Applied Clifford Algebras \textbf{31}(1), 9 (2021)

\bibitem{ablamowicz2021ternary}
Ab{\l}amowicz, R.: On ternary {Clifford} algebras on two generators defined by
  extra-special 3-groups of order 27.
\newblock Advances in Applied Clifford Algebras \textbf{31}(4), 62 (2021)

\bibitem{ablamowicz2018classification}
Ab{\l}amowicz, R., Varahagiri, M., Walley, A.M.: A classification of {Clifford}
  algebras as images of group algebras of {Salingaros} vee groups.
\newblock Advances in Applied Clifford Algebras \textbf{28}, 1--34 (2018)

\bibitem{anderson1995introduction}
Anderson, J.A.: An introduction to neural networks.
\newblock MIT press (1995)

\bibitem{aouiti2021finite}
Aouiti, C., Bessifi, M.: Finite-time and fixed-time synchronization of fuzzy
  {Clifford}-valued {Cohen}-{Grossberg} neural networks with discontinuous
  activations and time-varying delays.
\newblock International Journal of Adaptive Control and Signal Processing
  \textbf{35}(12), 2499--2520 (2021)

\bibitem{baez2017icosahedron}
Baez, J.C.: From the icosahedron to {$E_8$}.
\newblock arXiv preprint arXiv:1712.06436  (2017)

\bibitem{Bao:2020nbi}
Bao, J., Franco, S., He, Y.H., Hirst, E., Musiker, G., Xiao, Y.: {Quiver
  Mutations, Seiberg Duality and Machine Learning}.
\newblock Phys. Rev. D \textbf{102}(8), 086,013 (2020).
\newblock \doi{10.1103/PhysRevD.102.086013}

\bibitem{Bao:2022rup}
Bao, J., He, Y.H., Heyes, E., Hirst, E.: {Machine Learning Algebraic Geometry
  for Physics}  (2022)

\bibitem{Bao:2021olg}
Bao, J., He, Y.H., Hirst, E.: {Neurons on Amoebae}.
\newblock J. Symb. Comput. \textbf{116}, 1--38 (2022).
\newblock \doi{10.1016/j.jsc.2022.08.021}

\bibitem{Bao:2021ofk}
Bao, J., He, Y.H., Hirst, E., Hofscheier, J., Kasprzyk, A., Majumder, S.:
  {Polytopes and Machine Learning}  (2021)

\bibitem{Bao:2021auj}
Bao, J., He, Y.H., Hirst, E., Hofscheier, J., Kasprzyk, A., Majumder, S.:
  {Hilbert series, machine learning, and applications to physics}.
\newblock Phys. Lett. B \textbf{827}, 136,966 (2022).
\newblock \doi{10.1016/j.physletb.2022.136966}

\bibitem{bayro2010clifford}
Bayro-Corrochano, E.J., Arana-Daniel, N.: Clifford support vector machines for
  classification, regression, and recurrence.
\newblock IEEE Transactions on Neural Networks \textbf{21}(11), 1731--1746
  (2010)

\bibitem{bena2022algorithmically}
Bena, I., Bl{\aa}b{\"a}ck, J., Gra{\~n}a, M., L{\"u}st, S.: Algorithmically
  solving the tadpole problem.
\newblock Advances in Applied Clifford Algebras \textbf{32}(1), 7 (2022)

\bibitem{Berglund:2021ztg}
Berglund, P., Campbell, B., Jejjala, V.: {Machine Learning Kreuzer-Skarke
  Calabi-Yau Threefolds}  (2021)

\bibitem{Berglund:2023ztk}
Berglund, P., He, Y.H., Heyes, E., Hirst, E., Jejjala, V., Lukas, A.: {New
  Calabi-Yau Manifolds from Genetic Algorithms}  (2023)

\bibitem{Berman:2021mcw}
Berman, D.S., He, Y.H., Hirst, E.: {Machine learning Calabi-Yau hypersurfaces}.
\newblock Phys. Rev. D \textbf{105}(6), 066,002 (2022).
\newblock \doi{10.1103/PhysRevD.105.066002}

\bibitem{bonacich_1987}
Bonacich, P.: Power and centrality: A family of measures.
\newblock American Journal of Sociology \textbf{92}(5), 1170--1182 (1987).
\newblock \urlprefix\url{http://www.jstor.org/stable/2780000}

\bibitem{Bromborsky2020}
Bromborsky, A., Song, U., Wieser, E., Hadfield, H., {The Pygae Team}:
  pygae/galgebra: v0.5.0.
\newblock Zenodo  (2020).
\newblock \doi{10.5281/zenodo.3875882}

\bibitem{buchholz2008clifford}
Buchholz, S., Sommer, G.: On {Clifford} neurons and {Clifford} multi-layer
  perceptrons.
\newblock Neural Networks \textbf{21}(7), 925--935 (2008)

\bibitem{Bull:2018uow}
Bull, K., He, Y.H., Jejjala, V., Mishra, C.: {Machine Learning CICY
  Threefolds}.
\newblock Phys. Lett. B \textbf{785}, 65--72 (2018).
\newblock \doi{10.1016/j.physletb.2018.08.008}

\bibitem{CDHM2024ADE}
Cameron, P., Dechant, P.P., He, Y.H., McKay, J.: {ADE} - patterns in
  Mathematics.
\newblock Cambridge University Press (2024)

\bibitem{Chen:2022jwd}
Chen, S., He, Y.H., Hirst, E., Nestor, A., Zahabi, A.: {Mahler Measuring the
  Genetic Code of Amoebae}  (2022)

\bibitem{cheung2022clustering}
Cheung, M.W., Dechant, P.P., He, Y.H., Heyes, E., Hirst, E., Li, J.R.:
  Clustering cluster algebras with clusters.
\newblock accepted in Advances in Theoretical and Mathematical Physics; arXiv
  preprint arXiv:2212.09771  (2022)

\bibitem{coates2022machine}
Coates, T., Hofscheier, J., Kasprzyk, A.: Machine learning the dimension of a
  polytope (2022)

\bibitem{Dechant2016Birth}
Dechant, P.P.: The birth of {$E_8$} out of the spinors of the icosahedron.
\newblock Proceedings of the Royal Society A 20150504  (2016).
\newblock \urlprefix\url{http://dx.doi.org/10.1098/rspa.2015.0504}

\bibitem{dechant2017clifford}
Dechant, P.P.: Clifford algebra is the natural framework for root systems and
  {Coxeter} groups. group theory: {Coxeter}, conformal and modular groups.
\newblock Advances in Applied Clifford Algebras \textbf{27}, 17--31 (2017)

\bibitem{Dechant2017e8}
Dechant, P.P.: The {$E_8$} geometry from a {Clifford} perspective.
\newblock Advances in Applied Clifford Algebras \textbf{27}(1), 397--421 (2017)

\bibitem{dechant2018trinity}
Dechant, P.P.: From the trinity {($A_3$, $B_3$, $H_3$)} to an {ADE}
  correspondence.
\newblock Proceedings of the Royal Society A \textbf{474}(2220), 20180,034
  (2018)

\bibitem{dechant2013affine}
Dechant, P.P., B{\oe}hm, C., Twarock, R.: Affine extensions of
  non-crystallographic {Coxeter} groups induced by projection.
\newblock Journal of Mathematical Physics \textbf{54}(9) (2013)

\bibitem{dechant2022cluster}
Dechant, P.P., He, Y.H., Heyes, E., Hirst, E.: Cluster algebras: Network
  science and machine learning.
\newblock accepted in Journal of Computational Algebra; arXiv preprint
  arXiv:2203.13847  (2022)

\bibitem{doran2003geometric}
Doran, C., Lasenby, A.: Geometric algebra for physicists.
\newblock Cambridge University Press (2003)

\bibitem{dorst2014total}
Dorst, L.: Total least squares fitting of k-spheres in n-d euclidean space
  using an (n+ 2)-d isometric representation.
\newblock Journal of mathematical imaging and vision \textbf{50}, 214--234
  (2014)

\bibitem{frobenius1912matrizen}
Frobenius, G.: {\"U}ber {Matrizen} aus nicht negativen {Elementen}.
\newblock Preussische Akademie der Wissenschaften Berlin: Sitzungsberichte der
  Preu{\ss}ischen Akademie der Wissenschaften zu Berlin. Reichsdr. (1912).
\newblock \urlprefix\url{https://books.google.co.uk/books?id=fuK3PgAACAAJ}

\bibitem{groenendijk2023geometric}
Groenendijk, R., Dorst, L., Gevers, T.: Geometric back-propagation in
  morphological neural networks.
\newblock IEEE Transactions on Pattern Analysis and Machine Intelligence
  (2023)

\bibitem{Gukov:2020qaj}
Gukov, S., Halverson, J., Ruehle, F., Su\l{}kowski, P.: {Learning to Unknot}.
\newblock Mach. Learn. Sci. Tech. \textbf{2}(2), 025,035 (2021).
\newblock \doi{10.1088/2632-2153/abe91f}

\bibitem{he2021calabi}
He, Y.H.: The {Calabi}--{Yau} landscape: From geometry, to physics, to machine
  learning, vol. 2293.
\newblock Springer Nature (2021)

\bibitem{he2023machine}
He, Y.H., Heyes, E., Hirst, E.: Machine learning in physics and geometry.
\newblock arXiv preprint arXiv:2303.12626  (2023)

\bibitem{He:2020eva}
He, Y.H., Hirst, E., Peterken, T.: {Machine-learning dessins
  d\textquoteright{}enfants: explorations via modular and
  Seiberg\textendash{}Witten curves}.
\newblock J. Phys. A \textbf{54}(7), 075,401 (2021).
\newblock \doi{10.1088/1751-8121/abbc4f}

\bibitem{he2022murmurations}
He, Y.H., Lee, K.H., Oliver, T., Pozdnyakov, A.: Murmurations of elliptic
  curves.
\newblock arXiv preprint arXiv:2204.10140  (2022)

\bibitem{he2018calabi}
He, Y.H., Seong, R.K., Yau, S.T.: {Calabi}--{Yau} volumes and reflexive
  polytopes.
\newblock Communications in Mathematical Physics \textbf{361}, 155--204 (2018)

\bibitem{heal2022deep}
Heal, K., Kulkarni, A., Sert{\"o}z, E.C.: Deep learning {G}auss--{M}anin
  connections.
\newblock Advances in Applied Clifford Algebras \textbf{32}(2), 24 (2022)

\bibitem{helmstetter2023various}
Helmstetter, J.: Various characteristic properties of {L}ipschitzian elements
  in {C}lifford algebras.
\newblock Advances in Applied Clifford Algebras \textbf{33}(4), 43 (2023)

\bibitem{hestenes2012clifford}
Hestenes, D., Sobczyk, G.: {Clifford} algebra to geometric calculus: a unified
  language for mathematics and physics, vol.~5.
\newblock Springer Science \& Business Media (2012)

\bibitem{Hornik1989MultilayerFN}
Hornik, K., Stinchcombe, M.B., White, H.L.: Multilayer feedforward networks are
  universal approximators.
\newblock Neural Networks \textbf{2}, 359--366 (1989)

\bibitem{Humphreys1990Coxeter}
Humphreys, J.E.: Reflection groups and {Coxeter} groups.
\newblock Cambridge University Press, Cambridge (1990)

\bibitem{Jolliffe_2016}
Jolliffe, I.T., Cadima, J.: Principal component analysis: a review and recent
  developments.
\newblock Philosophical Transactions of the Royal Society A: Mathematical,
  Physical and Engineering Sciences \textbf{374} (2016).
\newblock \doi{doi.org/10.1098/rsta.2015.0202}

\bibitem{apocrita}
King, T., Butcher, S., Zalewski, L.: {Apocrita - High Performance Computing
  Cluster for Queen Mary University of London} (2017).
\newblock \doi{10.5281/zenodo.438045}.
\newblock \urlprefix\url{https://doi.org/10.5281/zenodo.438045}

\bibitem{kobayashi2021synthesis}
Kobayashi, M.: Synthesis of complex-and hyperbolic-valued {Hopfield} neural
  networks.
\newblock Neurocomputing \textbf{423}, 80--88 (2021)

\bibitem{kuroe2011models}
Kuroe, Y., Tanigawa, S., Iima, H.: Models of {Hopfield}-type {Clifford} neural
  networks and their energy functions-hyperbolic and dual valued networks.
\newblock In: Neural Information Processing: 18th International Conference,
  ICONIP 2011, Shanghai, China, November 13-17, 2011, Proceedings, Part I 18,
  pp. 560--569. Springer (2011)

\bibitem{lasenby2022some}
Lasenby, A.: Some recent results for {$SU(3)$} and octonions within the
  {Geometric Algebra} approach to the fundamental forces of nature.
\newblock Mathematical Methods in the Applied Sciences  (2022)

\bibitem{lasenby2022reconstructing}
Lasenby, A., Lasenby, J., Matsantonis, C.: Reconstructing a rotor from initial
  and final frames using characteristic multivectors: With applications in
  orthogonal transformations.
\newblock Mathematical Methods in the Applied Sciences  (2022)

\bibitem{LESHNO1993861}
Leshno, M., Lin, V.Y., Pinkus, A., Schocken, S.: Multilayer feedforward
  networks with a nonpolynomial activation function can approximate any
  function.
\newblock Neural Networks \textbf{6}(6), 861--867 (1993).
\newblock \doi{doi.org/10.1016/S0893-6080(05)80131-5}

\bibitem{moya2021quaternion}
Moya-S{\'a}nchez, E.U., Xamb{\'o}-Descamps, S., Salazar~Colores, S.,
  S{\'a}nchez~P{\'e}rez, A., Cort{\'e}s, U.: A quaternion deterministic
  monogenic {CNN} layer for contrast invariance.
\newblock In: Systems, Patterns and Data Engineering with Geometric Calculi,
  pp. 133--152. Springer (2021)

\bibitem{Niarchos:2023lot}
Niarchos, V., Papageorgakis, C., Richmond, P., Stapleton, A.G., Woolley, M.:
  {Bootstrability in Line-Defect CFT with Improved Truncation Methods}  (2023)

\bibitem{pepe2022learning}
Pepe, A., Lasenby, J., Chac{\'o}n, P.: Learning rotations.
\newblock Mathematical Methods in the Applied Sciences  (2022)

\bibitem{pepe2022using}
Pepe, A., Lasenby, J., Chac{\'o}n, P.: Using a graph transformer network to
  predict 3d coordinates of proteins via {Geometric Algebra} modelling.
\newblock In: International Workshop on Empowering Novel Geometric Algebra for
  Graphics and Engineering, pp. 83--95. Springer (2022)

\bibitem{Perron1907}
Perron, O.: Zur {Theorie} der {Matrices}.
\newblock Mathematische Annalen \textbf{64}(2), 248--263 (1907).
\newblock \doi{10.1007/BF01449896}.
\newblock \urlprefix\url{https://doi.org/10.1007/BF01449896}

\bibitem{roelfs2023geometric}
Roelfs, M.: Geometric invariant decomposition of {$SU(3)$}.
\newblock Advances in Applied Clifford Algebras \textbf{33}(1), 5 (2023)

\bibitem{roelfs2023graded}
Roelfs, M., De~Keninck, S.: Graded symmetry groups: plane and simple.
\newblock Advances in Applied Clifford Algebras \textbf{33}(3), 30 (2023)

\bibitem{shirokov2019calculation}
Shirokov, D.: Calculation of elements of spin groups using method of averaging
  in {Clifford}’s {Geometric} {Algebra}.
\newblock Advances in Applied Clifford Algebras \textbf{29}(3), 1--12 (2019)

\bibitem{shirokov2021computing}
Shirokov, D.: On computing the determinant, other characteristic polynomial
  coefficients, and inverse in {Clifford} algebras of arbitrary dimension.
\newblock Computational and Applied Mathematics \textbf{40}(5), 1--29 (2021)

\bibitem{da2020efficient}
da~Silva, D., de~Araujo, C.P., Chow, E.: An efficient homomorphic data encoding
  with multiple secret {H}ensel codes.
\newblock International Journal of Information and Electronics Engineering
  \textbf{10}(1) (2020)

\bibitem{oeis_connectedgraphs}
Sloane, N.J.A.: {The On-Line Encyclopedia of Integer Sequences, A001349}.
\newblock \urlprefix\url{https://oeis.org/A001349}.
\newblock Number of connected graphs with $n$ nodes.

\bibitem{smith1970some}
Smith, J.H.: Some properties of the spectrum of a graph.
\newblock Combinatorial Structures and their applications pp. 403--406 (1970)

\bibitem{sriraman2022stability}
Sriraman, R., Rajchakit, G., Kwon, O.M., Lee, S.M.: Stability analysis for
  delayed {Cohen}--{Grossberg} {Clifford}-valued neutral-type neural networks.
\newblock Mathematical Methods in the Applied Sciences \textbf{45}(17),
  10,925--10,945 (2022)

\bibitem{sundararajan2017axiomatic}
Sundararajan, M., Taly, A., Yan, Q.: Axiomatic attribution for deep networks.
\newblock In: International conference on machine learning, pp. 3319--3328.
  PMLR (2017)

\bibitem{vieira2023bicomplex}
Vieira, N.: Bicomplex neural networks with hypergeometric activation functions.
\newblock Advances in Applied Clifford Algebras \textbf{33}(2), 20 (2023)

\bibitem{wang2016clifford}
Wang, R., Zhang, X., Cao, W.: Clifford fuzzy support vector machines for
  classification.
\newblock Advances in applied Clifford algebras \textbf{26}, 825--846 (2016)

\bibitem{wilson2021problem}
Wilson, R.A.: On the problem of choosing subgroups of {Clifford} algebras for
  applications in fundamental physics.
\newblock Advances in Applied Clifford Algebras \textbf{31}(4), 59 (2021)

\end{thebibliography}

\appendix

\section{NN Gradient Saliency Results}\label{app_GS}
The gradient saliency results, showing the relative importance of the input parts of the Coxeter element permutation for predicting the subinvariants at each order, are presented in the subsequent Tables \ref{tab:GSA8regression}, \ref{tab:GSD8regression}, \ref{tab:GSE8regression}.

\begin{table}[!ht]
    \centering
    \begin{tabular}{|c||>{\centering\arraybackslash}m{5em}|>
    {\centering\arraybackslash}m{5em}|>{\centering\arraybackslash}m{5em}|>{\centering\arraybackslash}m{5em}|>{\centering\arraybackslash}m{5em}|>{\centering\arraybackslash}m{5em}|}
    \hline
    & $\Inv_{i}$ & $\Inv_{i}^{0}$ & $\Inv_{i}^{2}$ & $\Inv_{i}^{4}$ & $\Inv_{i}^{6}$ & $\Inv_{i}^{8}$\\
    \hline
    $\Inv_0$ & \raisebox{-.5\height}{ \includegraphics[height=10mm]{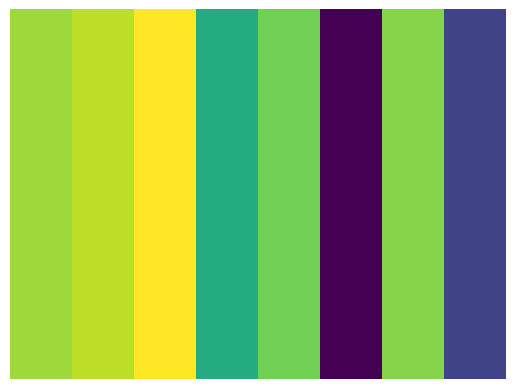}}& \raisebox{-.5\height}{ \includegraphics[height=10mm]{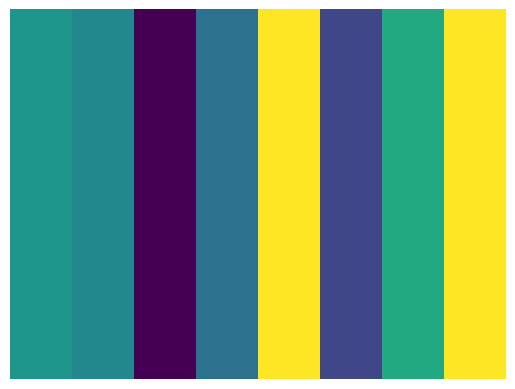}} & & & &  \\
    \hline
    $\Inv_1$ & \raisebox{-.5\height}{ \includegraphics[height=10mm]{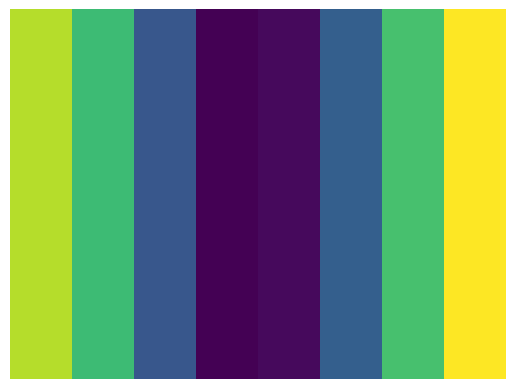}}& \raisebox{-.5\height}{ \includegraphics[height=10mm]{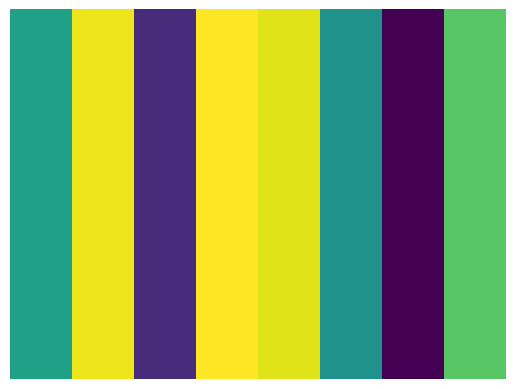}} &\raisebox{-.5\height}{ \includegraphics[height=10mm]{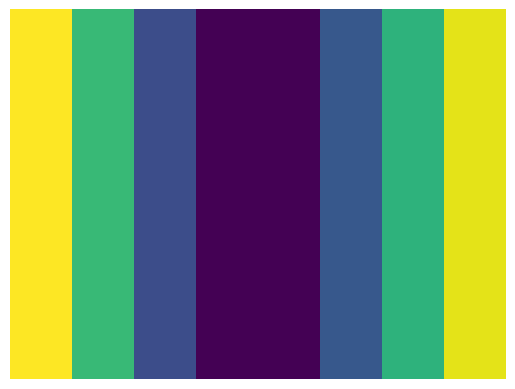}}& & & \\
    \hline
    $\Inv_2$ & \raisebox{-.5\height}{ \includegraphics[height=10mm]{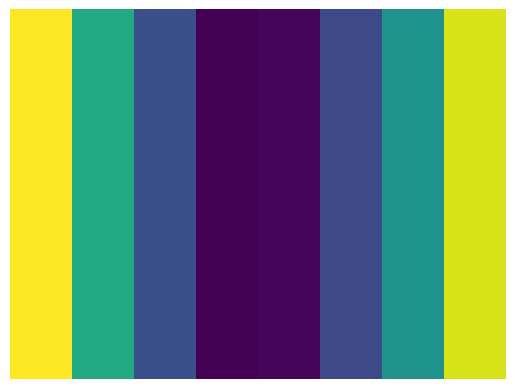}}&\raisebox{-.5\height}{ \includegraphics[height=10mm]{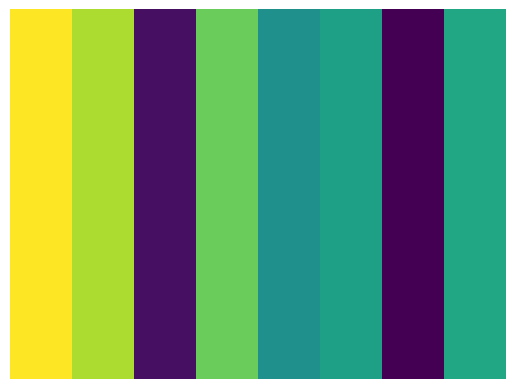}}&\raisebox{-.5\height}{ \includegraphics[height=10mm]{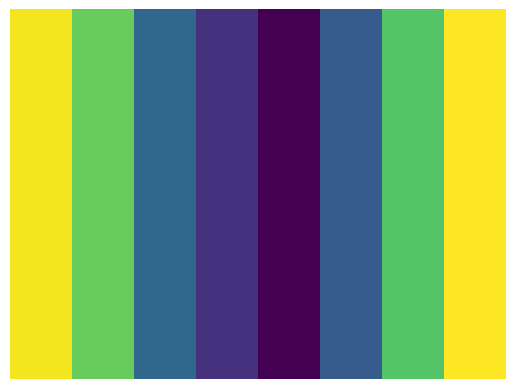}}& \raisebox{-.5\height}{ \includegraphics[height=10mm]{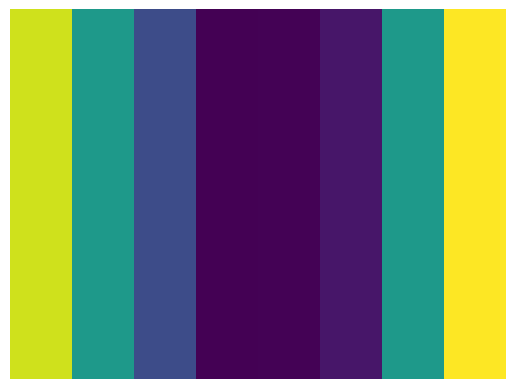}} & & \\
    \hline
    $\Inv_3$ & \raisebox{-.5\height}{ \includegraphics[height=10mm]{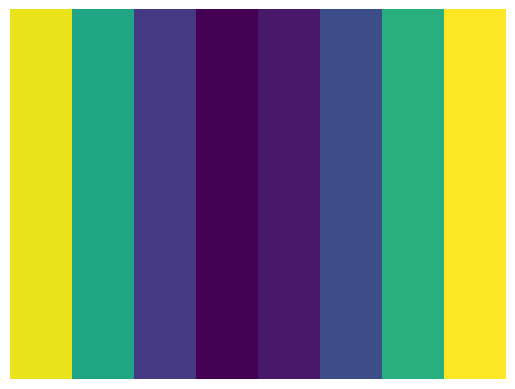}}&\raisebox{-.5\height}{ \includegraphics[height=10mm]{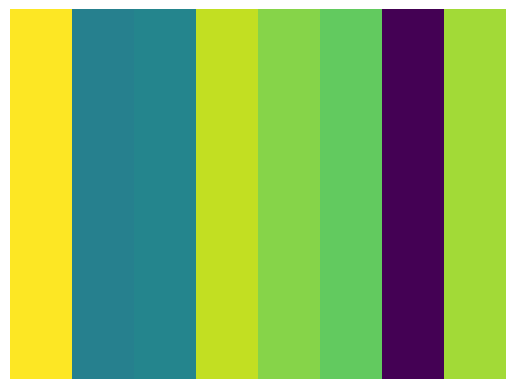}} &\raisebox{-.5\height}{ \includegraphics[height=10mm]{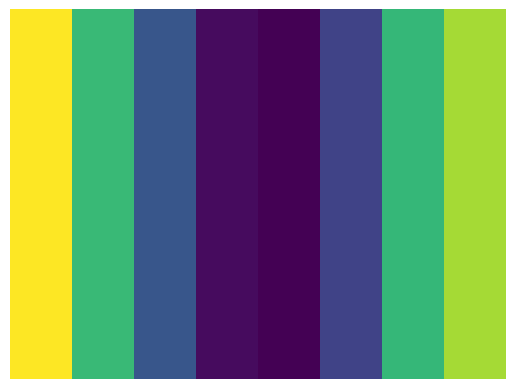}} &\raisebox{-.5\height}{ \includegraphics[height=10mm]{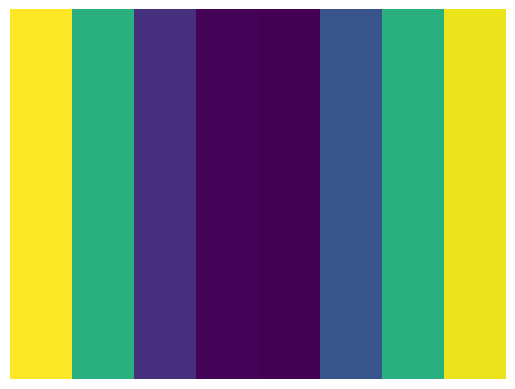}} & \raisebox{-.5\height}{ \includegraphics[height=10mm]{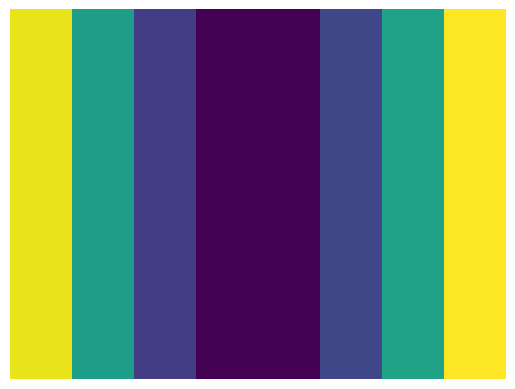}} & \\
    \hline
    $\Inv_4$ & \raisebox{-.5\height}{ \includegraphics[height=10mm]{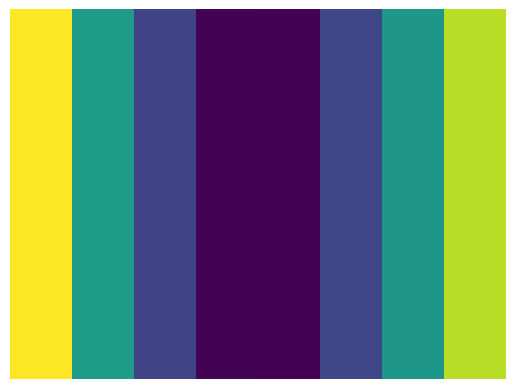}}&\raisebox{-.5\height}{ \includegraphics[height=10mm]{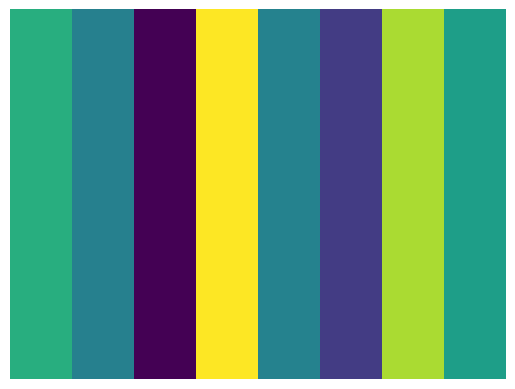}}&  \raisebox{-.5\height}{ \includegraphics[height=10mm]{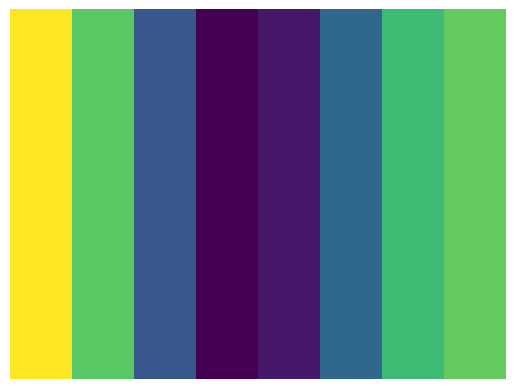}}& \raisebox{-.5\height}{ \includegraphics[height=10mm]{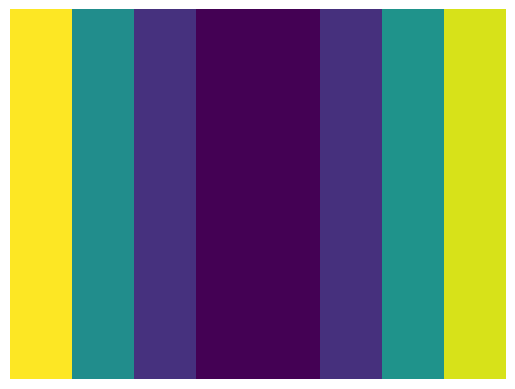}} & \raisebox{-.5\height}{ \includegraphics[height=10mm]{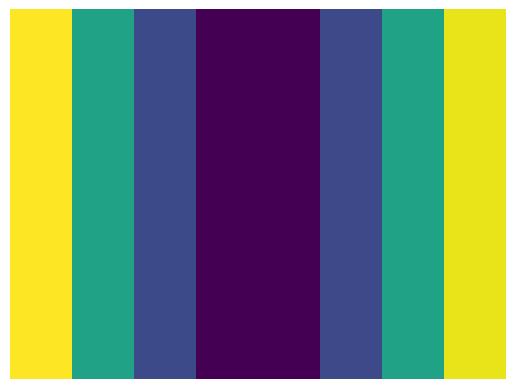}}& \raisebox{-.5\height}{ \includegraphics[height=10mm]{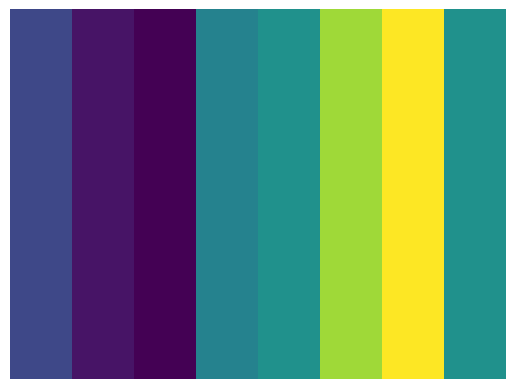}}\\
    \hline
    $\Inv_5$ & \raisebox{-.5\height}{ \includegraphics[height=10mm]{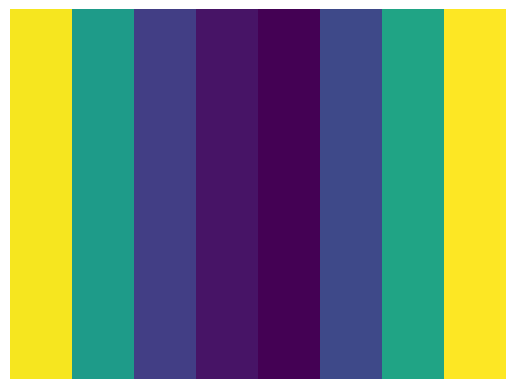}}& \raisebox{-.5\height}{ \includegraphics[height=10mm]{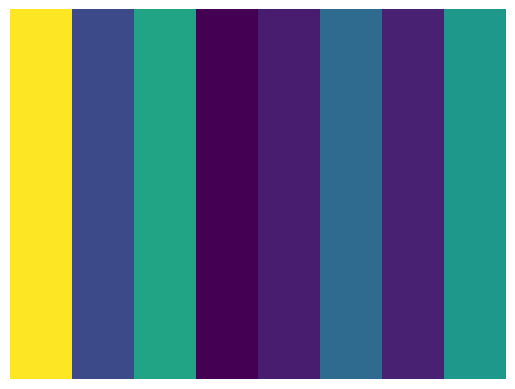}} & \raisebox{-.5\height}{ \includegraphics[height=10mm]{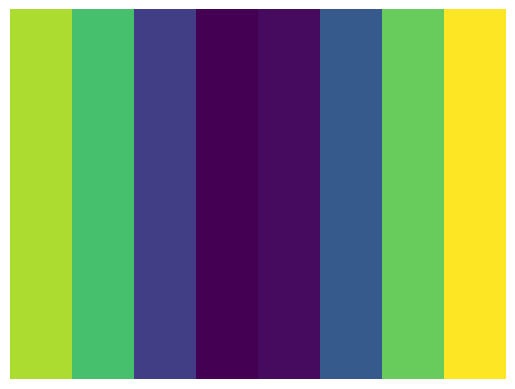}}& \raisebox{-.5\height}{ \includegraphics[height=10mm]{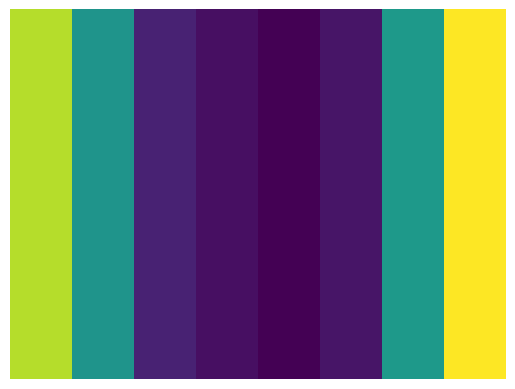}} & \raisebox{-.5\height}{ \includegraphics[height=10mm]{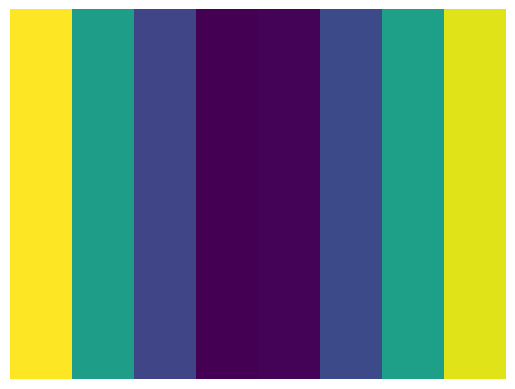}}& \\
    \hline
    $\Inv_6$ & \raisebox{-.5\height}{ \includegraphics[height=10mm]{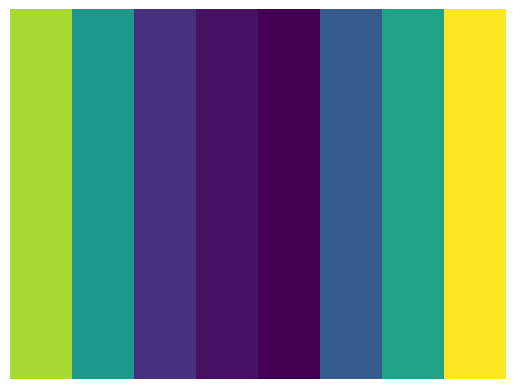}}& \raisebox{-.5\height}{ \includegraphics[height=10mm]{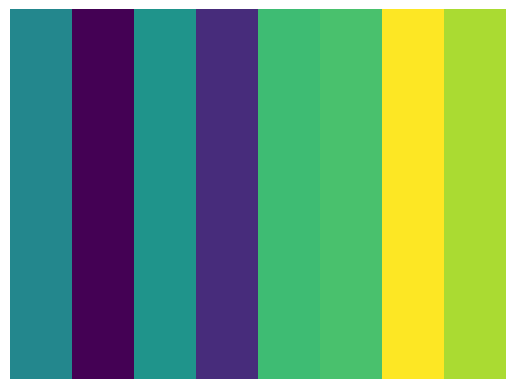}}&  \raisebox{-.5\height}{ \includegraphics[height=10mm]{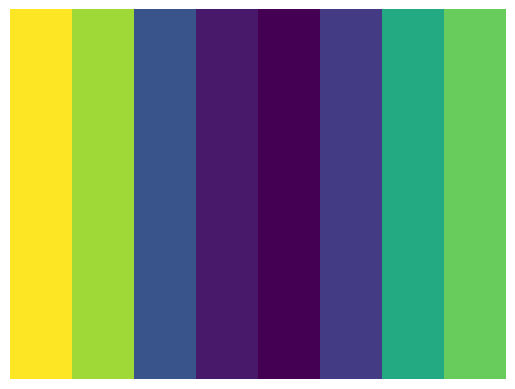}} & \raisebox{-.5\height}{ \includegraphics[height=10mm]{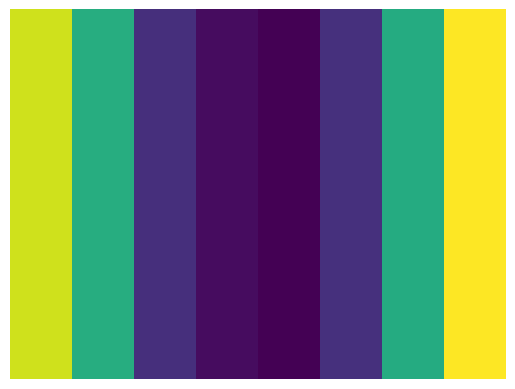}} & &\\
    \hline
    $\Inv_7$ & \raisebox{-.5\height}{ \includegraphics[height=10mm]{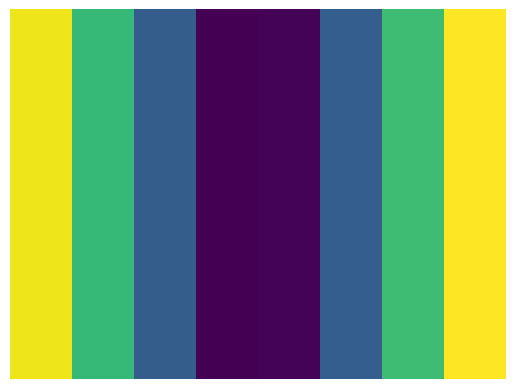}}&\raisebox{-.5\height}{ \includegraphics[height=10mm]{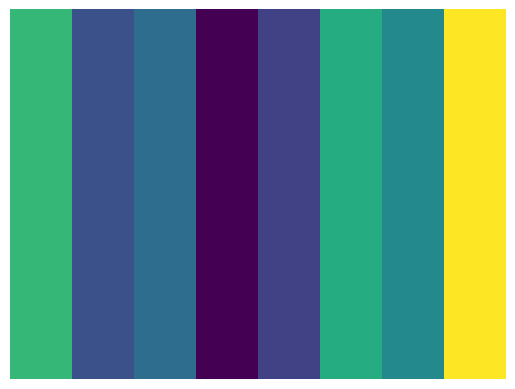}}& \raisebox{-.5\height}{ \includegraphics[height=10mm]{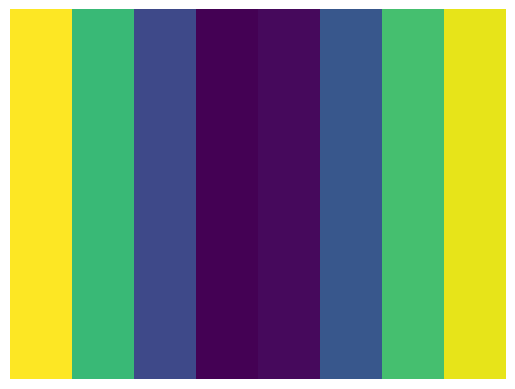}}& & &\\
    \hline
    $\Inv_8$ & \raisebox{-.5\height}{ \includegraphics[height=10mm]{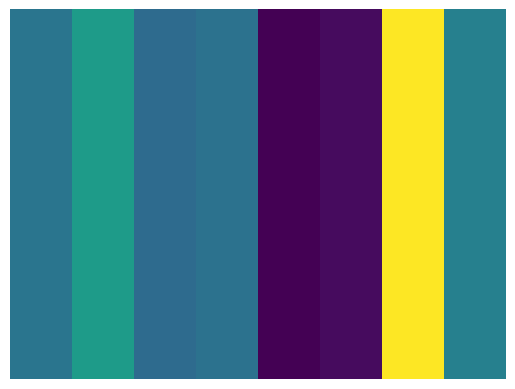}}& \raisebox{-.5\height}{ \includegraphics[height=10mm]{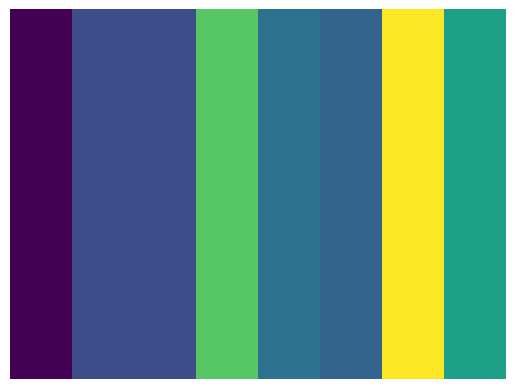}}& & & & \\
    \hline
    \end{tabular}
    \caption{Summary of gradient saliency analysis for each invariant and subinvariant for $A_8$ simple root data. The NN function takes as input the permutation performed on the original Coxeter element and outputs the coefficients of the respective subinvariant. The saliency barcodes hence represent the relative importance of each root in the permutation for computing the subinvariant coefficients. Lighter colours indicate larger gradients and greater importance.}
    \label{tab:GSA8regression}
\end{table}

\begin{table}[!ht]
    \centering
    \begin{tabular}{|c||>{\centering\arraybackslash}m{5em}|>
    {\centering\arraybackslash}m{5em}|>{\centering\arraybackslash}m{5em}|>{\centering\arraybackslash}m{5em}|>{\centering\arraybackslash}m{5em}|>{\centering\arraybackslash}m{5em}|}
    \hline
    & $\Inv_{i}$ & $\Inv_{i}^{0}$ & $\Inv_{i}^{2}$ & $\Inv_{i}^{4}$ & $\Inv_{i}^{6}$ & $\Inv_{i}^{8}$\\
    \hline
    $\Inv_0$ & \raisebox{-.5\height}{ \includegraphics[height=10mm]{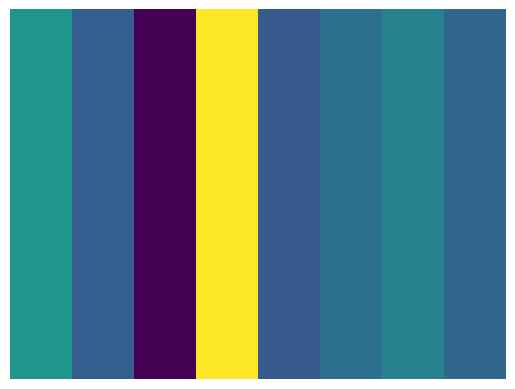}} & \raisebox{-.5\height}{ \includegraphics[height=10mm]{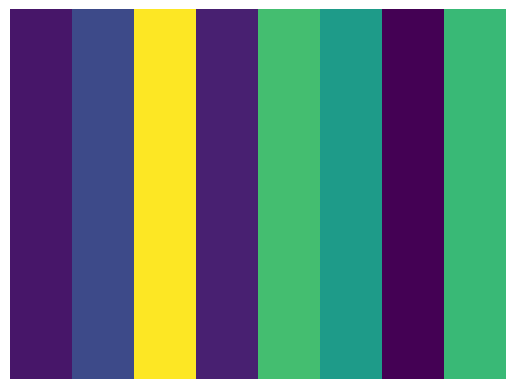}} & & & &  \\
    \hline
    $\Inv_1$ & \raisebox{-.5\height}{ \includegraphics[height=10mm]{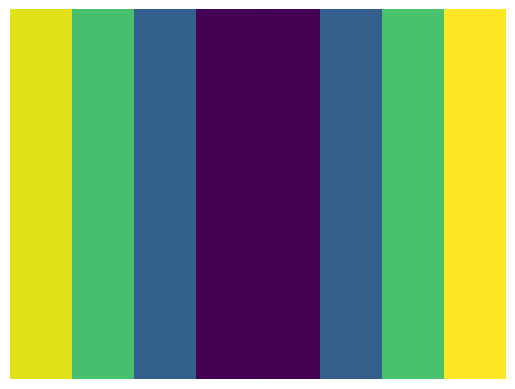}} & \raisebox{-.5\height}{ \includegraphics[height=10mm]{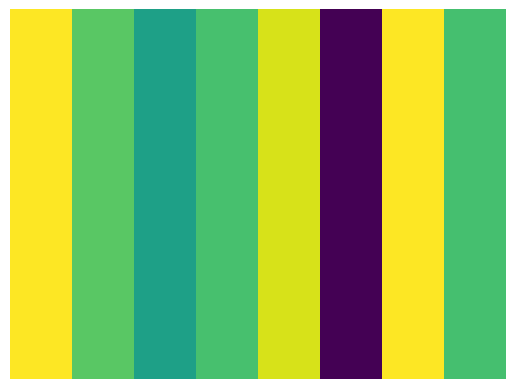}} &\raisebox{-.5\height}{ \includegraphics[height=10mm]{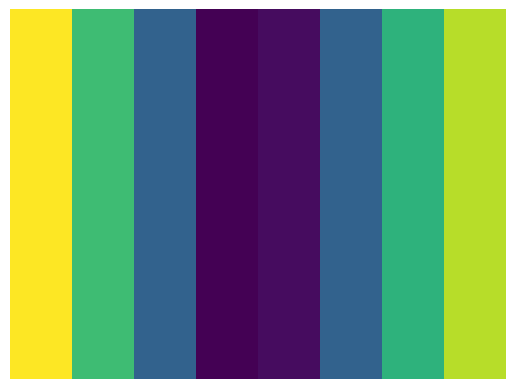}}& & & \\
    \hline
    $\Inv_2$ & \raisebox{-.5\height}{ \includegraphics[height=10mm]{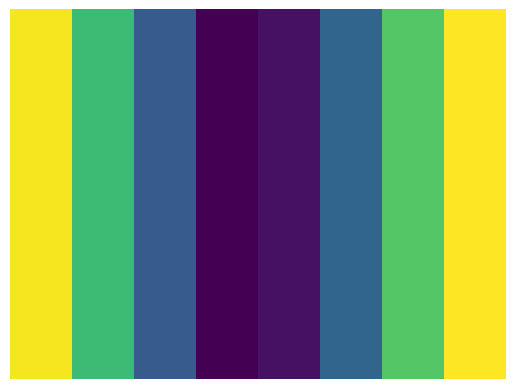}} &\raisebox{-.5\height}{ \includegraphics[height=10mm]{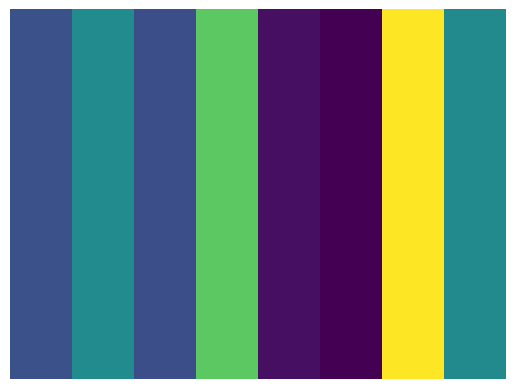}}&\raisebox{-.5\height}{ \includegraphics[height=10mm]{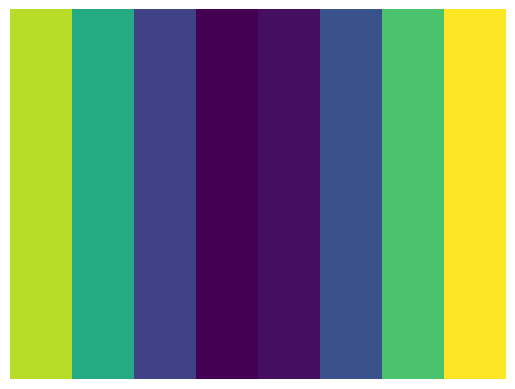}}& \raisebox{-.5\height}{ \includegraphics[height=10mm]{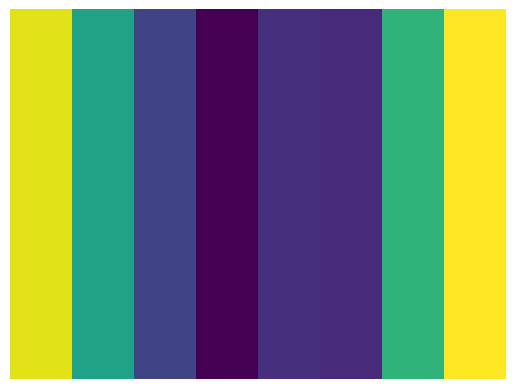}} & & \\
    \hline
    $\Inv_3$ & \raisebox{-.5\height}{ \includegraphics[height=10mm]{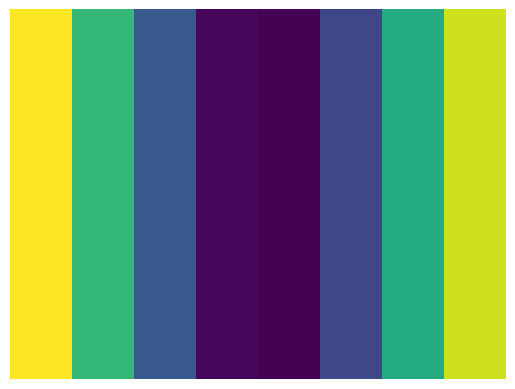}} &\raisebox{-.5\height}{ \includegraphics[height=10mm]{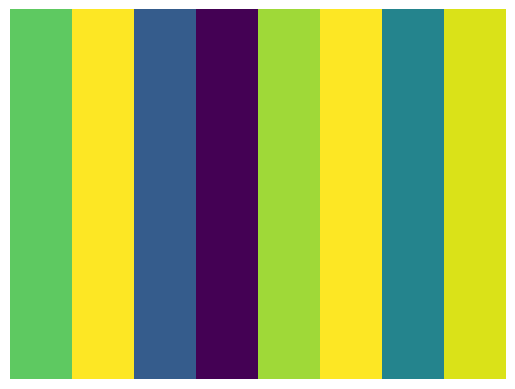}} &\raisebox{-.5\height}{ \includegraphics[height=10mm]{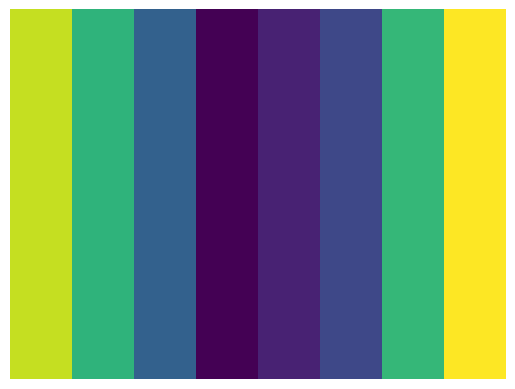}} &\raisebox{-.5\height}{ \includegraphics[height=10mm]{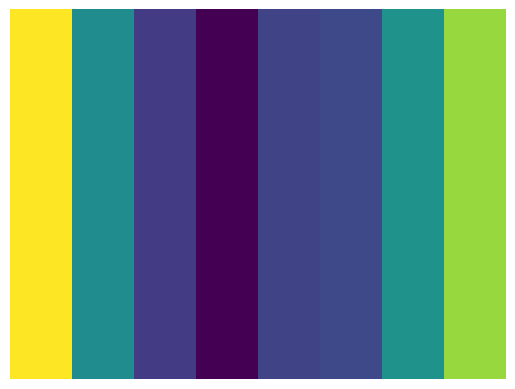}} & \raisebox{-.5\height}{ \includegraphics[height=10mm]{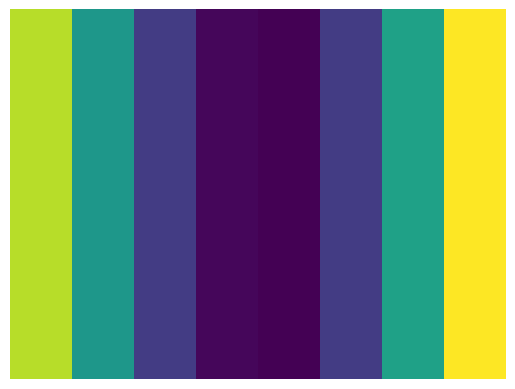}} & \\
    \hline
    $\Inv_4$ & \raisebox{-.5\height}{ \includegraphics[height=10mm]{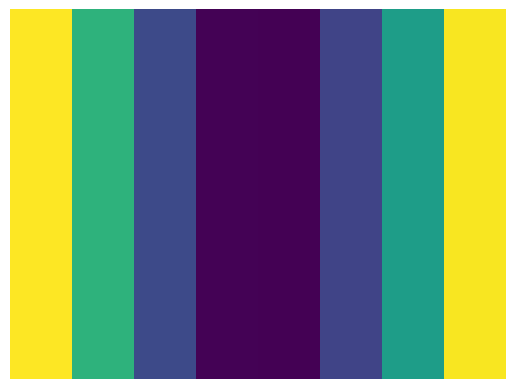}} &\raisebox{-.5\height}{ \includegraphics[height=10mm]{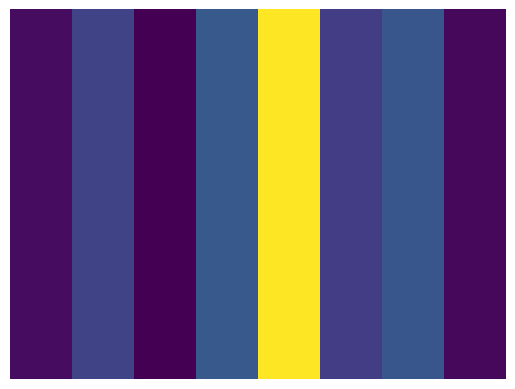}}&  \raisebox{-.5\height}{ \includegraphics[height=10mm]{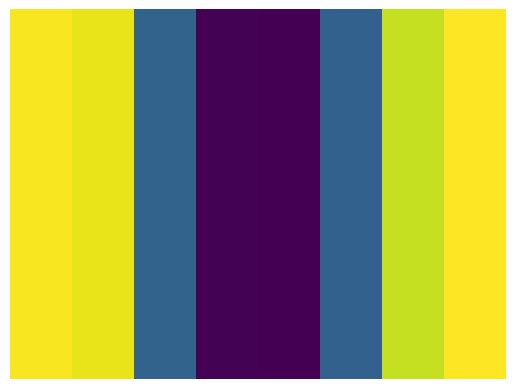}}& \raisebox{-.5\height}{ \includegraphics[height=10mm]{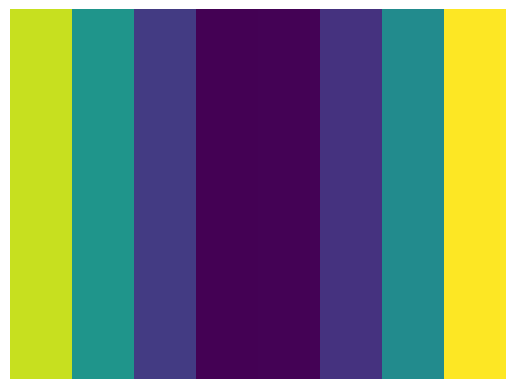}} & \raisebox{-.5\height}{ \includegraphics[height=10mm]{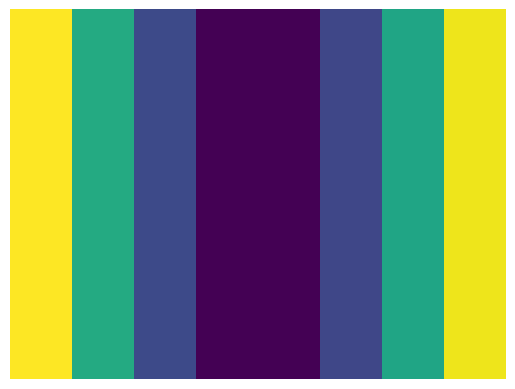}}& \raisebox{-.5\height}{ \includegraphics[height=10mm]{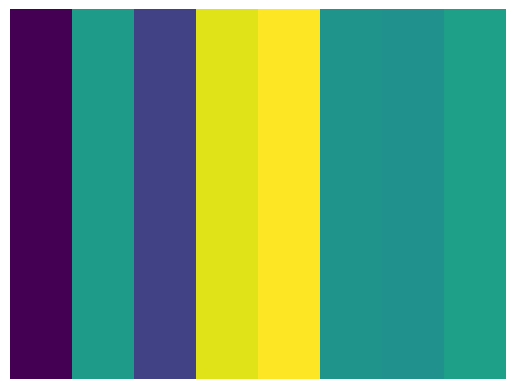}}\\
    \hline
    $\Inv_5$ & \raisebox{-.5\height}{ \includegraphics[height=10mm]{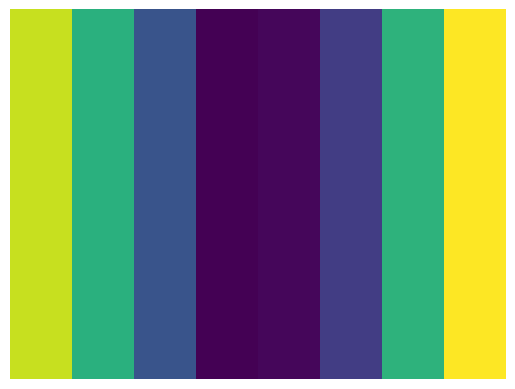}} & \raisebox{-.5\height}{ \includegraphics[height=10mm]{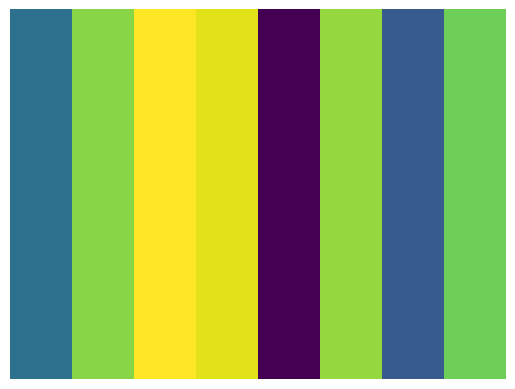}} & \raisebox{-.5\height}{ \includegraphics[height=10mm]{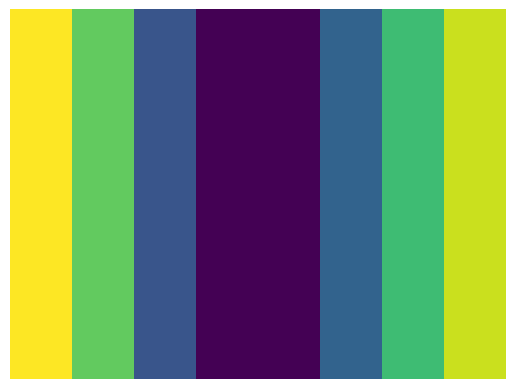}}& \raisebox{-.5\height}{ \includegraphics[height=10mm]{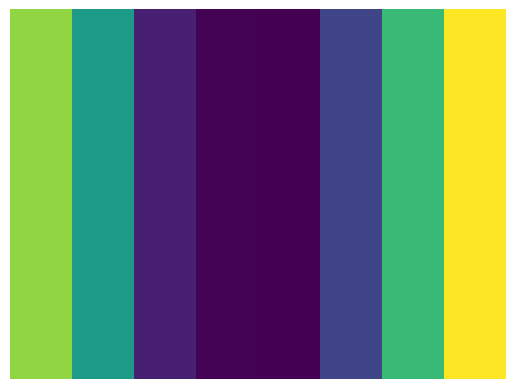}} & \raisebox{-.5\height}{ \includegraphics[height=10mm]{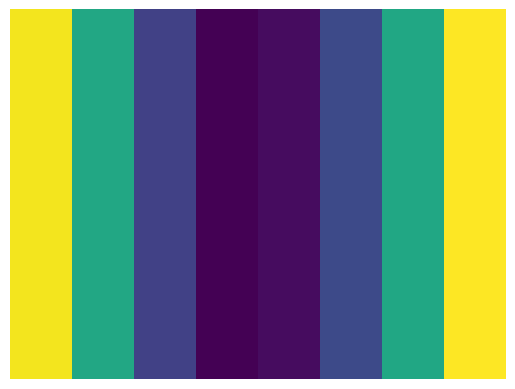}}& \\
    \hline
    $\Inv_6$ & \raisebox{-.5\height}{ \includegraphics[height=10mm]{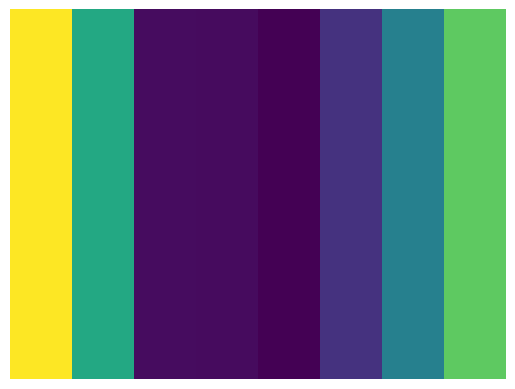}} & \raisebox{-.5\height}{ \includegraphics[height=10mm]{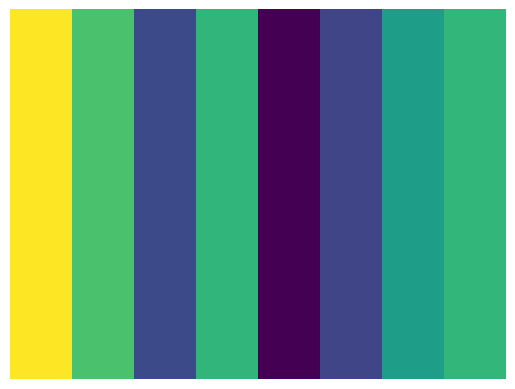}}&  \raisebox{-.5\height}{ \includegraphics[height=10mm]{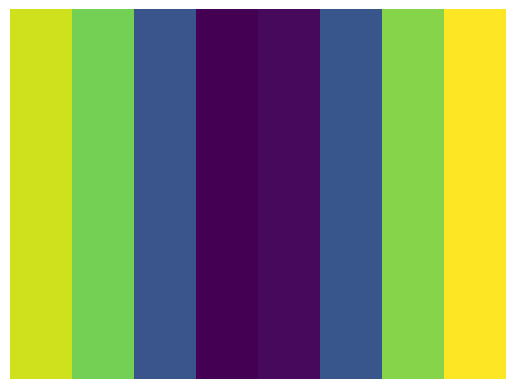}} & \raisebox{-.5\height}{ \includegraphics[height=10mm]{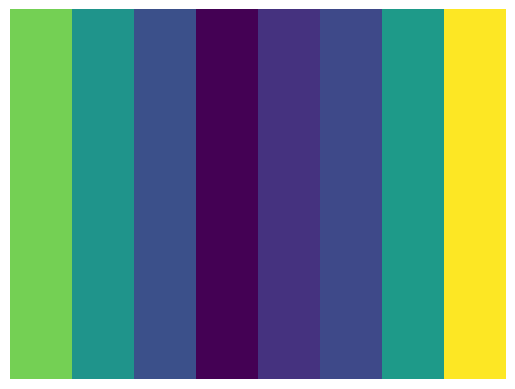}} & &\\
    \hline
    $\Inv_7$ & \raisebox{-.5\height}{ \includegraphics[height=10mm]{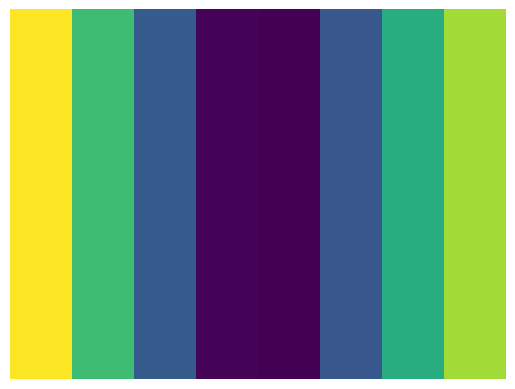}} &\raisebox{-.5\height}{ \includegraphics[height=10mm]{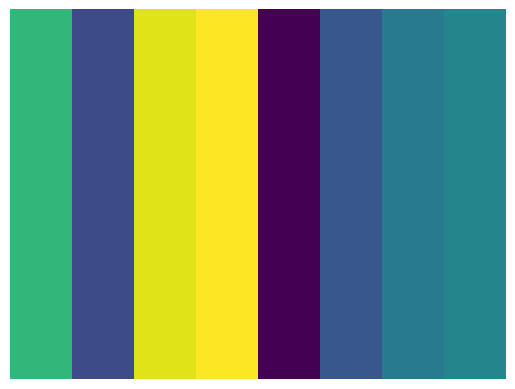}}& \raisebox{-.5\height}{ \includegraphics[height=10mm]{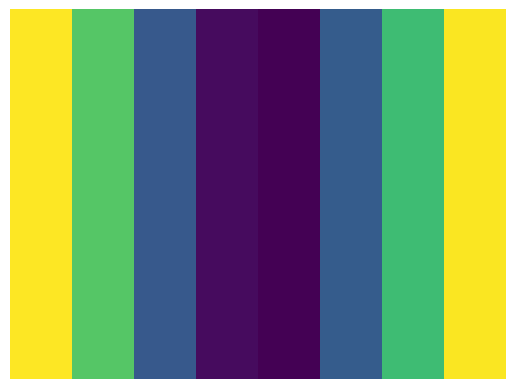}}& & &\\
    \hline
    $\Inv_8$ & \raisebox{-.5\height}{ \includegraphics[height=10mm]{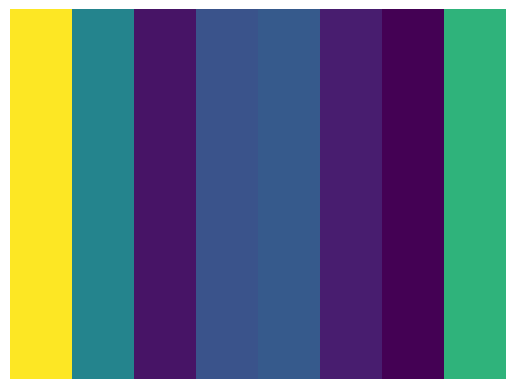}} & \raisebox{-.5\height}{ \includegraphics[height=10mm]{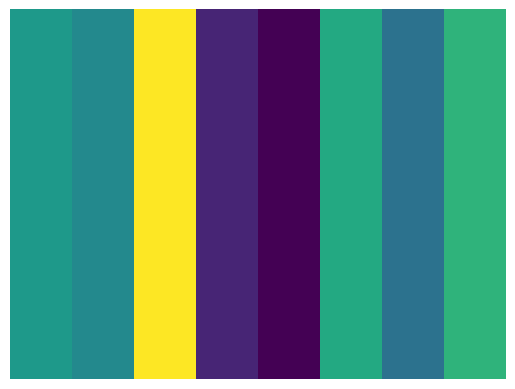}}& & & & \\
    \hline
    \end{tabular}
    \caption{Summary of gradient saliency analysis for each invariant and subinvariant for $D_8$ simple root data. The NN function takes as input the permutation performed on the original Coxeter element and outputs the coefficients of the respective subinvariant. The saliency barcodes hence represent the relative importance of each root in the permutation for computing the subinvariant coefficients. Lighter colours indicate larger gradients and greater importance.}
    \label{tab:GSD8regression}
\end{table}

\begin{table}[!ht]
    \centering
    \begin{tabular}{|c||>{\centering\arraybackslash}m{5em}|>
    {\centering\arraybackslash}m{5em}|>{\centering\arraybackslash}m{5em}|>{\centering\arraybackslash}m{5em}|>{\centering\arraybackslash}m{5em}|>{\centering\arraybackslash}m{5em}|}
    \hline
    & $\Inv_{i}$ & $\Inv_{i}^{0}$ & $\Inv_{i}^{2}$ & $\Inv_{i}^{4}$ & $\Inv_{i}^{6}$ & $\Inv_{i}^{8}$\\
    \hline
    $\Inv_0$ & \raisebox{-.5\height}{ \includegraphics[height=10mm]{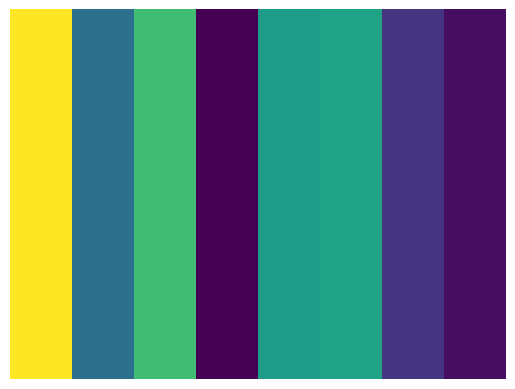}}& \raisebox{-.5\height}{ \includegraphics[height=10mm]{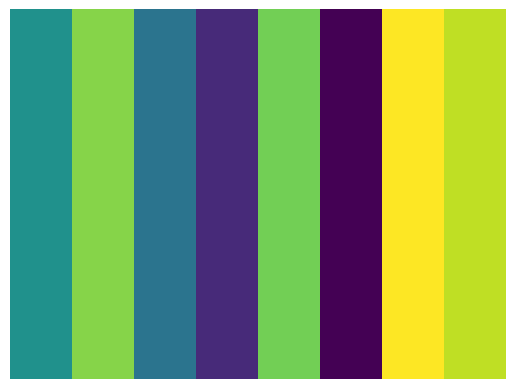}} & & & &  \\
    \hline
    $\Inv_1$ & \raisebox{-.5\height}{ \includegraphics[height=10mm]{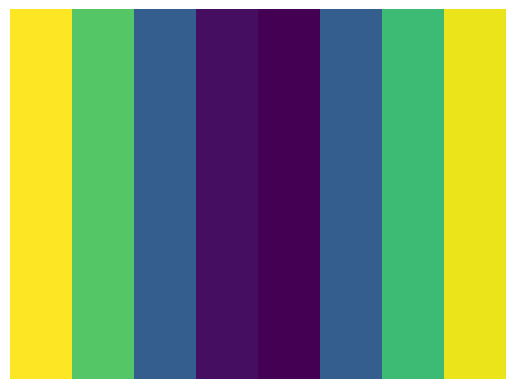}}& \raisebox{-.5\height}{ \includegraphics[height=10mm]{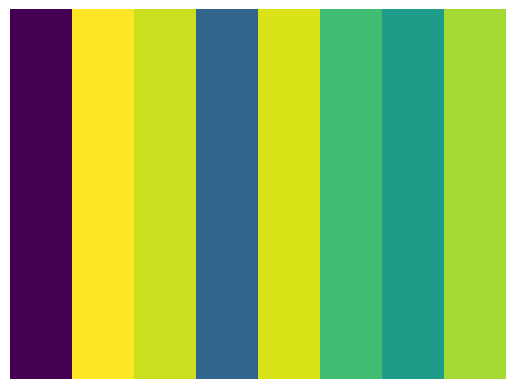}} &\raisebox{-.5\height}{ \includegraphics[height=10mm]{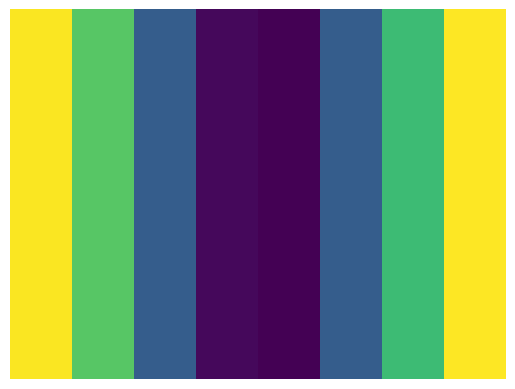}}& & & \\
    \hline
    $\Inv_2$ & \raisebox{-.5\height}{ \includegraphics[height=10mm]{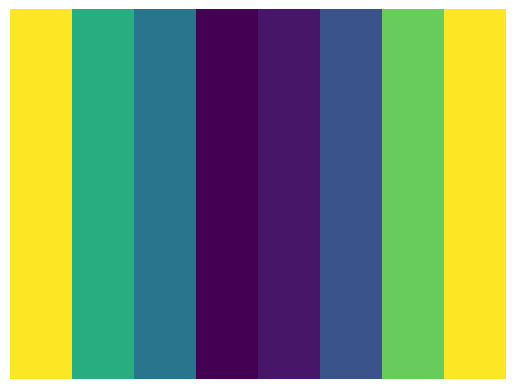}}&\raisebox{-.5\height}{ \includegraphics[height=10mm]{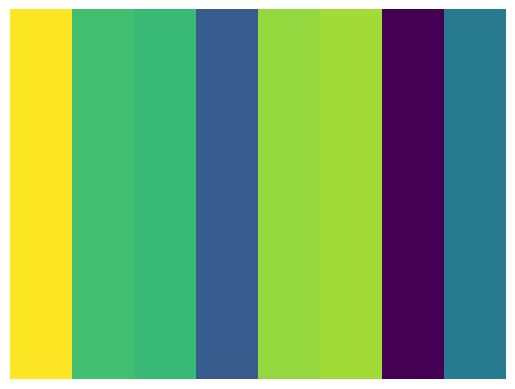}}&\raisebox{-.5\height}{ \includegraphics[height=10mm]{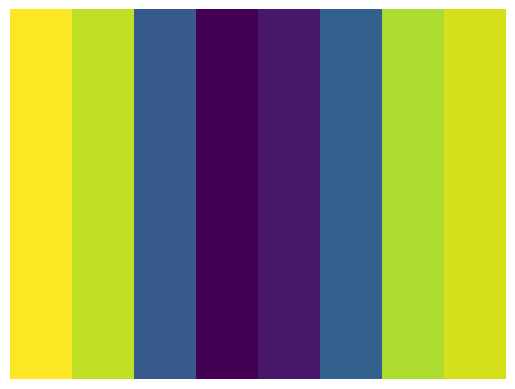}}& \raisebox{-.5\height}{ \includegraphics[height=10mm]{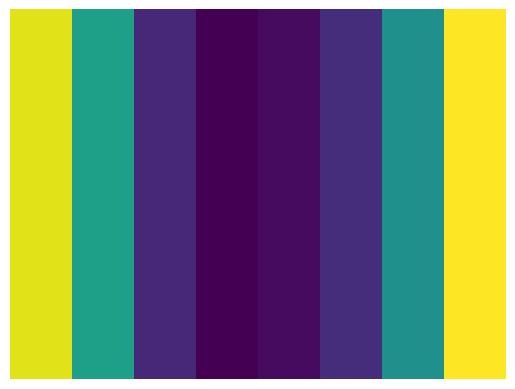}} & & \\
    \hline
    $\Inv_3$ & \raisebox{-.5\height}{ \includegraphics[height=10mm]{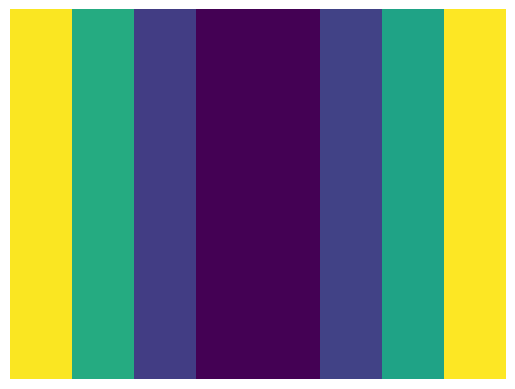}}&\raisebox{-.5\height}{ \includegraphics[height=10mm]{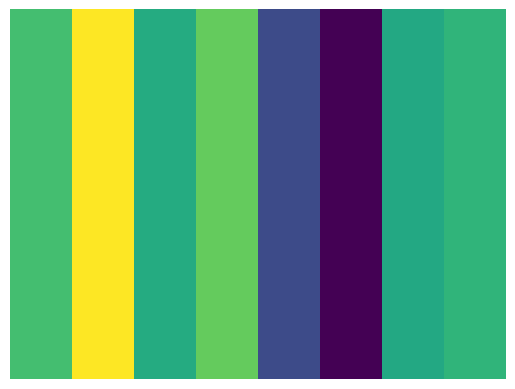}} &\raisebox{-.5\height}{ \includegraphics[height=10mm]{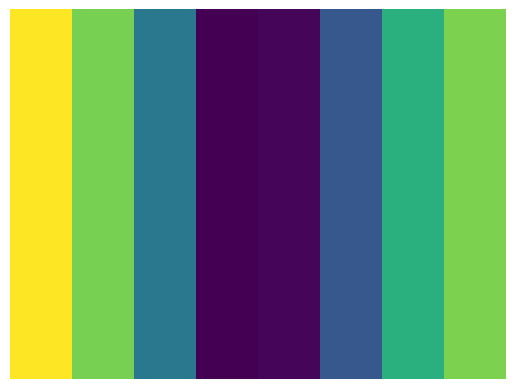}} &\raisebox{-.5\height}{ \includegraphics[height=10mm]{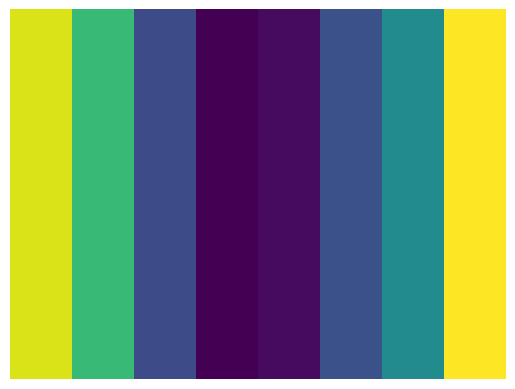}} & \raisebox{-.5\height}{ \includegraphics[height=10mm]{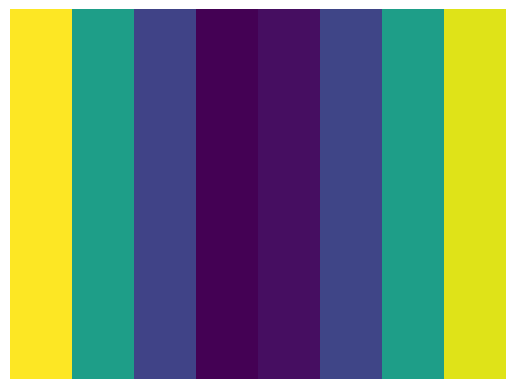}} & \\
    \hline
    $\Inv_4$ & \raisebox{-.5\height}{ \includegraphics[height=10mm]{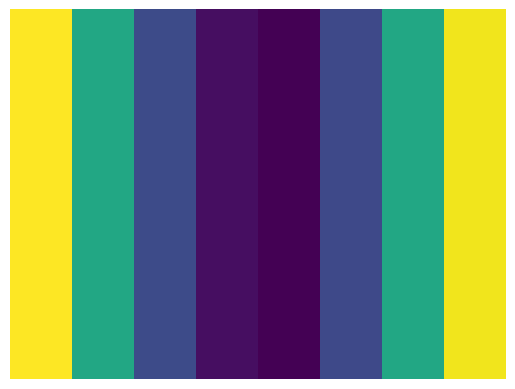}}&\raisebox{-.5\height}{ \includegraphics[height=10mm]{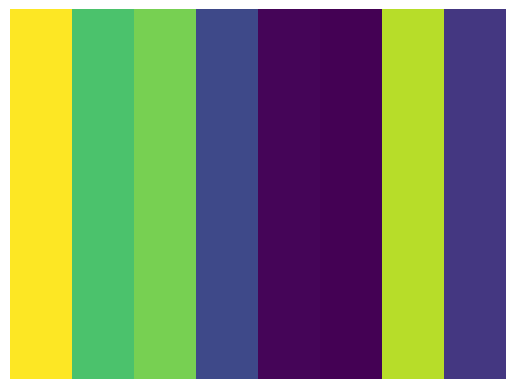}}&  \raisebox{-.5\height}{ \includegraphics[height=10mm]{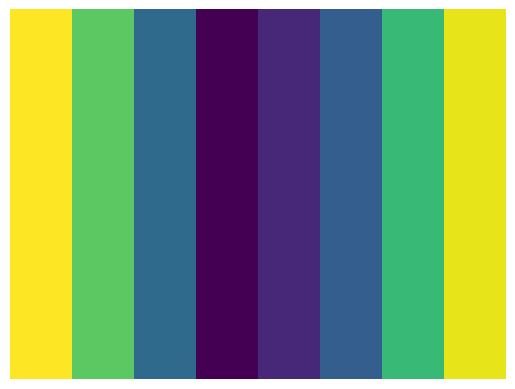}}& \raisebox{-.5\height}{ \includegraphics[height=10mm]{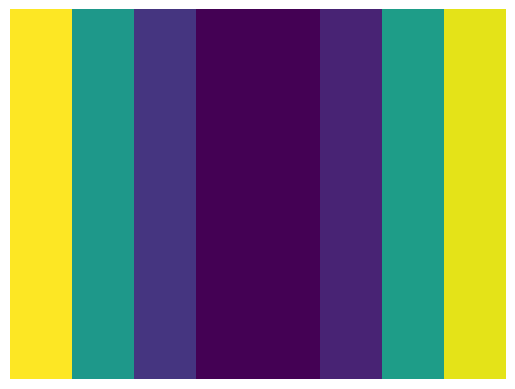}} & \raisebox{-.5\height}{ \includegraphics[height=10mm]{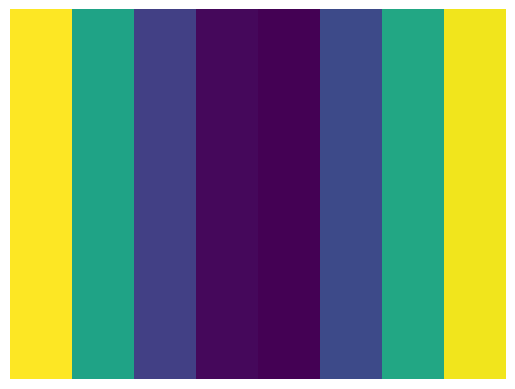}}& \raisebox{-.5\height}{ \includegraphics[height=10mm]{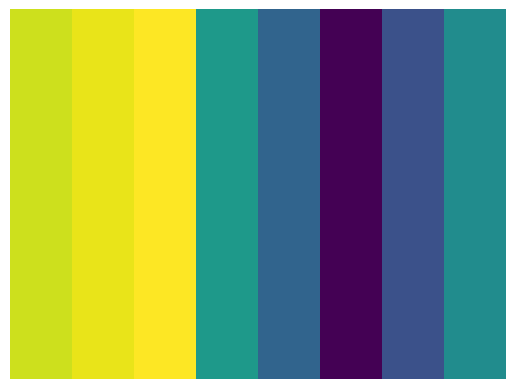}}\\
    \hline
    $\Inv_5$ & \raisebox{-.5\height}{ \includegraphics[height=10mm]{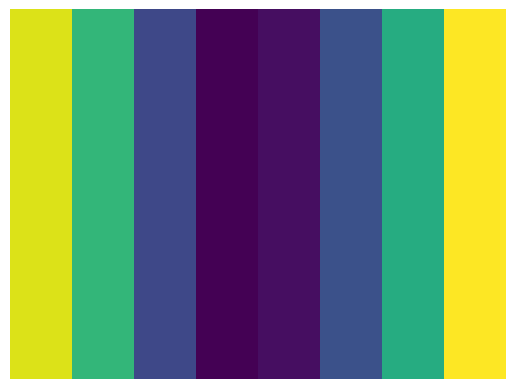}} & \raisebox{-.5\height}{ \includegraphics[height=10mm]{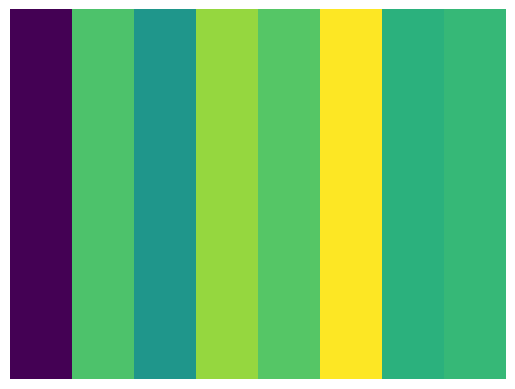}} & \raisebox{-.5\height}{ \includegraphics[height=10mm]{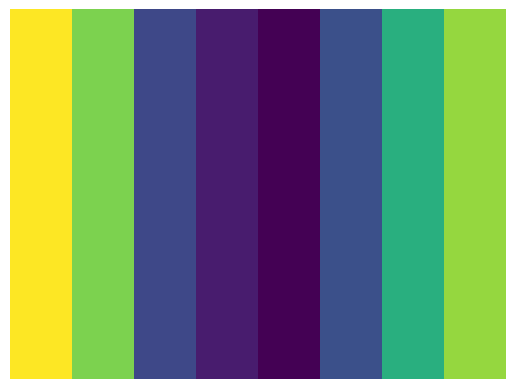}}& \raisebox{-.5\height}{ \includegraphics[height=10mm]{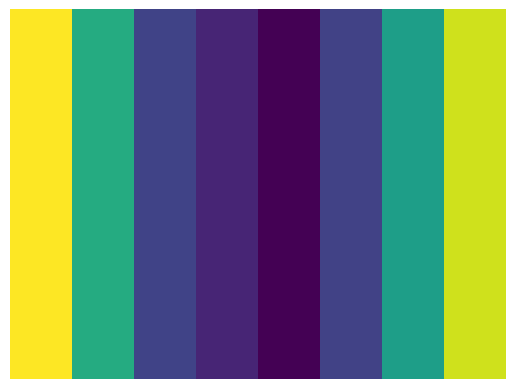}} & \raisebox{-.5\height}{ \includegraphics[height=10mm]{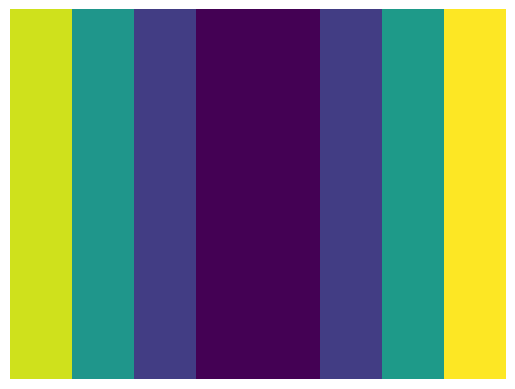}}& \\
    \hline
    $\Inv_6$ & \raisebox{-.5\height}{ \includegraphics[height=10mm]{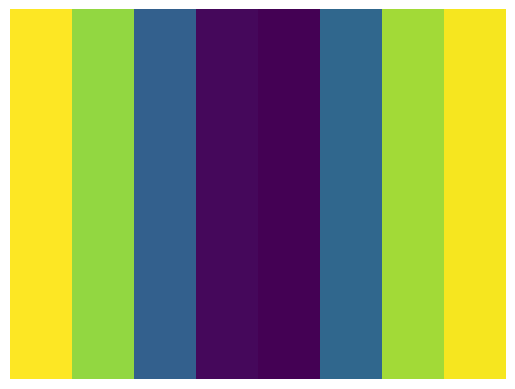}}& \raisebox{-.5\height}{ \includegraphics[height=10mm]{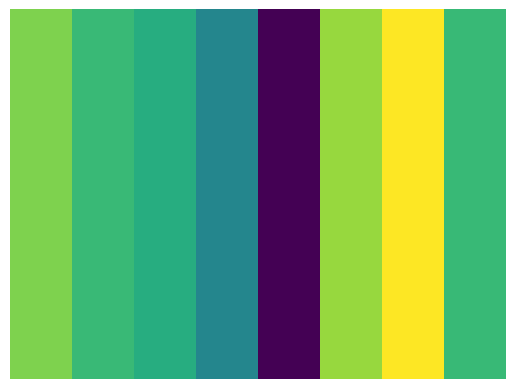}}&  \raisebox{-.5\height}{ \includegraphics[height=10mm]{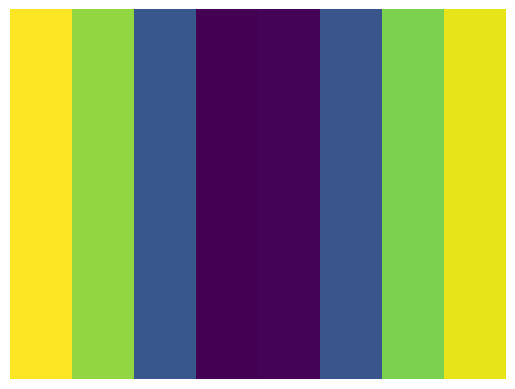}} & \raisebox{-.5\height}{ \includegraphics[height=10mm]{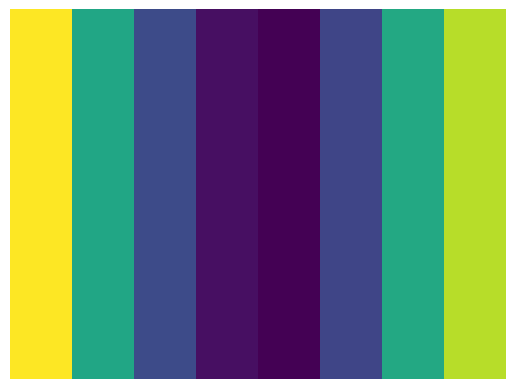}} & &\\
    \hline
    $\Inv_7$ & \raisebox{-.5\height}{ \includegraphics[height=10mm]{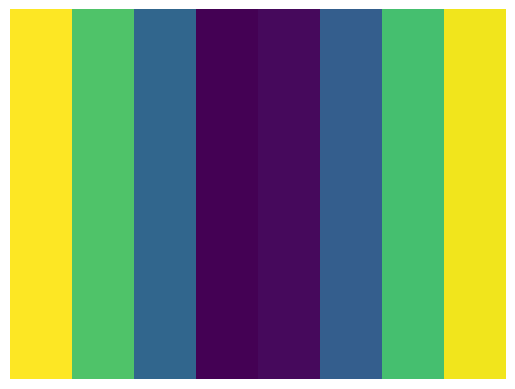}} &\raisebox{-.5\height}{ \includegraphics[height=10mm]{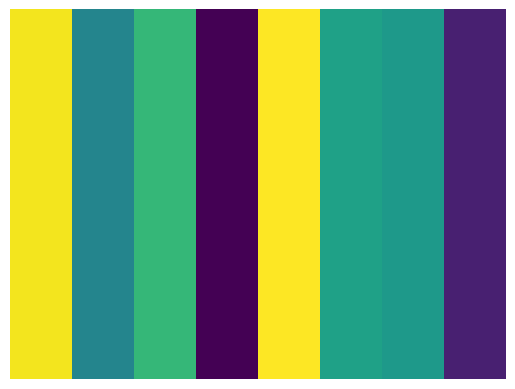}}& \raisebox{-.5\height}{ \includegraphics[height=10mm]{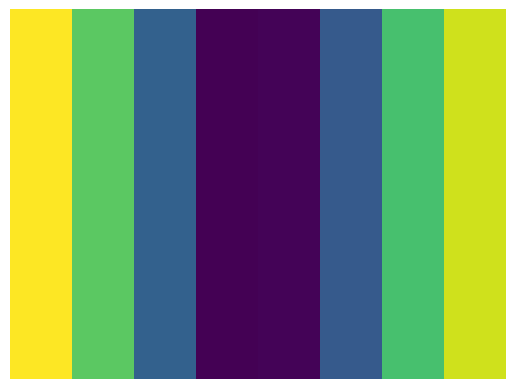}}& & &\\
    \hline
    $\Inv_8$ & \raisebox{-.5\height}{ \includegraphics[height=10mm]{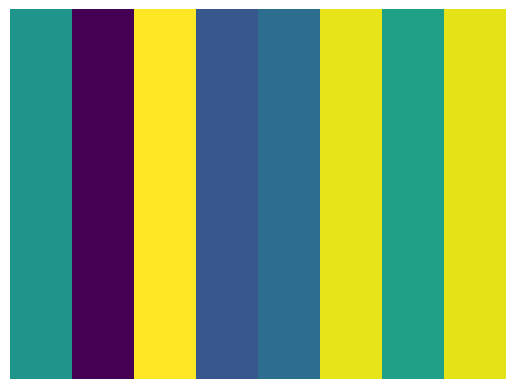}} & \raisebox{-.5\height}{ \includegraphics[height=10mm]{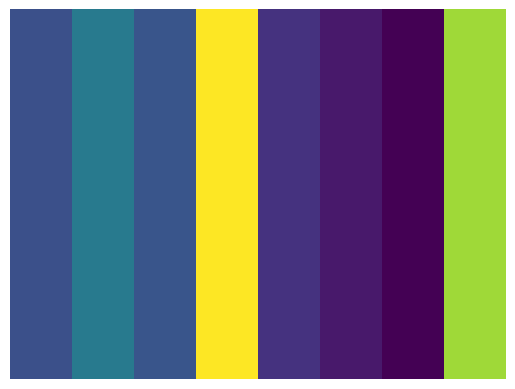}}& & & & \\
    \hline
    \end{tabular}
    \caption{Summary of gradient saliency analysis for each invariant and subinvariant for $E_8$ simple root data. The NN function takes as input the permutation performed on the original Coxeter element and outputs the coefficients of the respective subinvariant. The saliency barcodes hence represent the relative importance of each root in the permutation for computing the subinvariant coefficients. Lighter colours indicate larger gradients and greater importance.}
    \label{tab:GSE8regression}
\end{table}

\section{PCA Results}
\begin{figure}[h!]
    \centering
    \begin{subfigure}{0.32\textwidth}
        \centering
        \includegraphics[width=0.99\textwidth]{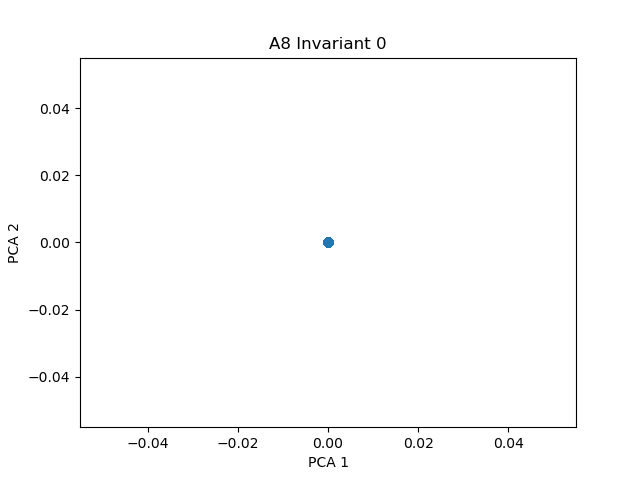}
        \caption{$\Inv_0$}
    \end{subfigure}
    \begin{subfigure}{0.32\textwidth}
        \centering
        \includegraphics[width=0.99\textwidth]{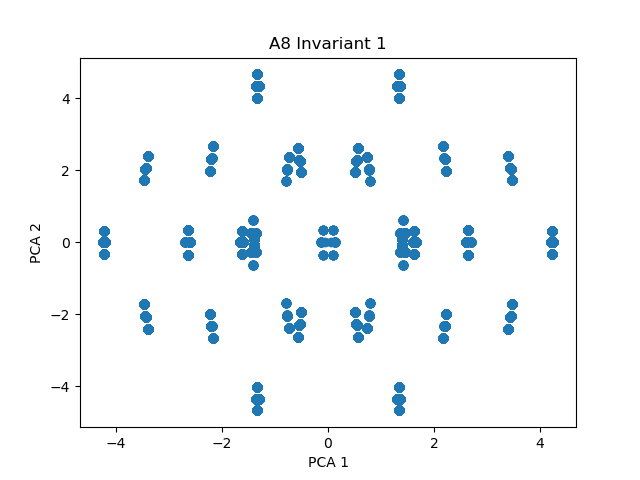}
        \caption{$\Inv_1$}
    \end{subfigure} 
    \begin{subfigure}{0.32\textwidth}
        \centering
        \includegraphics[width=0.99\textwidth]{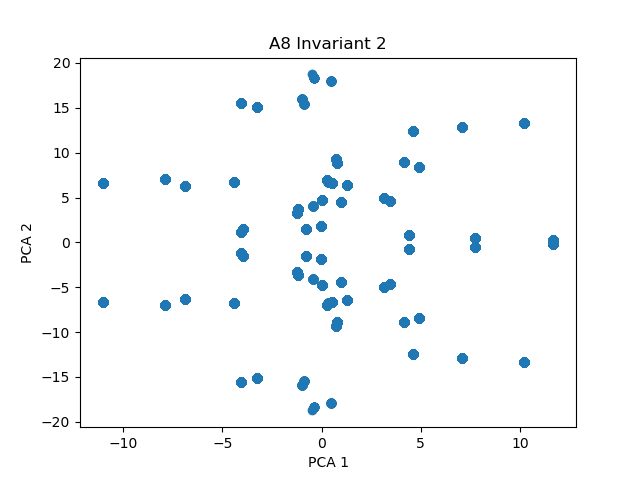}
        \caption{$\Inv_2$}
    \end{subfigure}
    \begin{subfigure}{0.32\textwidth}
        \centering
        \includegraphics[width=0.99\textwidth]{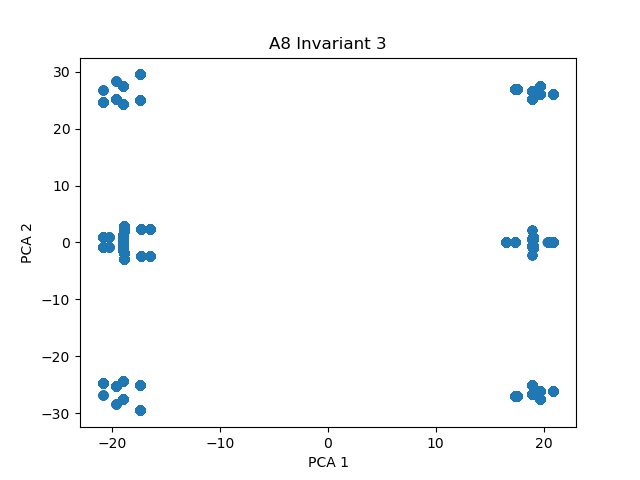}
        \caption{$\Inv_3$}
    \end{subfigure}
    \begin{subfigure}{0.32\textwidth}
        \centering
        \includegraphics[width=0.99\textwidth]{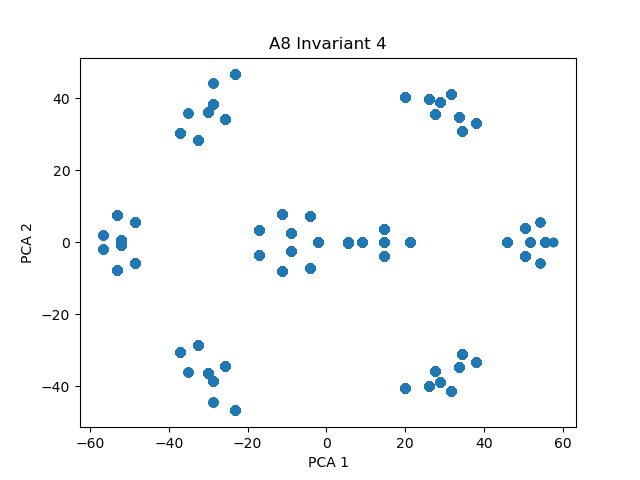}
        \caption{$\Inv_4$}
    \end{subfigure} 
    \begin{subfigure}{0.32\textwidth}
        \centering
        \includegraphics[width=0.99\textwidth]{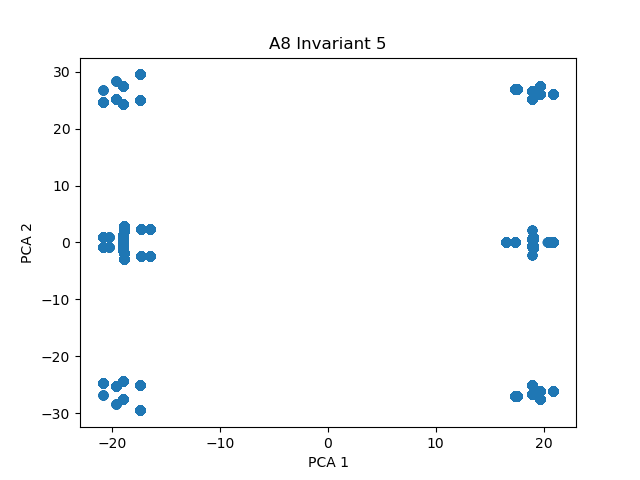}
        \caption{$\Inv_5$}
    \end{subfigure}
    \begin{subfigure}{0.32\textwidth}
        \centering
        \includegraphics[width=0.99\textwidth]{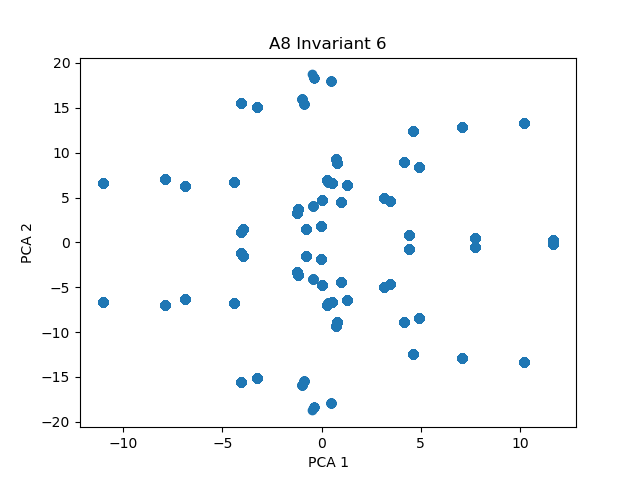}
        \caption{$\Inv_6$}
    \end{subfigure}
    \begin{subfigure}{0.32\textwidth}
        \centering
        \includegraphics[width=0.99\textwidth]{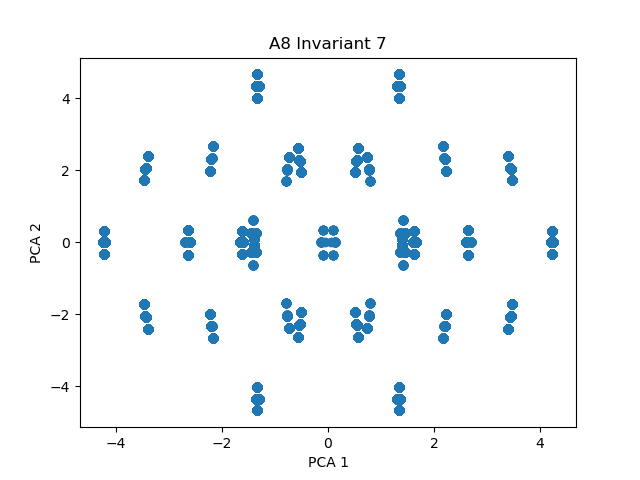}
        \caption{$\Inv_7$}
    \end{subfigure} 
    \begin{subfigure}{0.32\textwidth}
        \centering
        \includegraphics[width=0.99\textwidth]{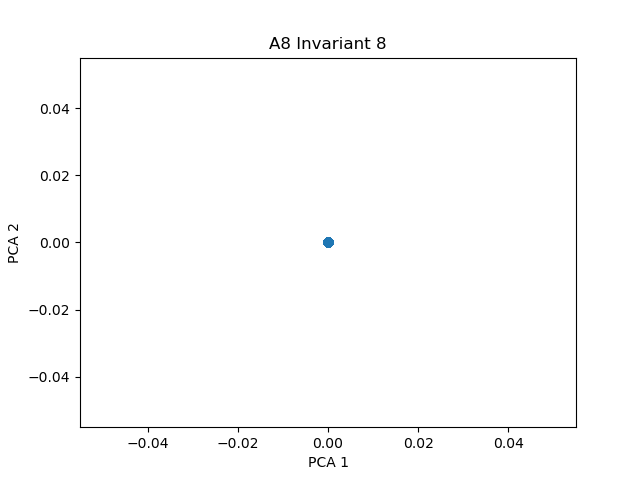}
        \caption{$\Inv_8$}
    \end{subfigure}
    \caption{PCA plots of the 9 orders of invariant (SOCM) for $A_{8}$. The observed mirror symmetry is a good consistency check. }\label{fig:PCA_A8}
\end{figure}

\begin{figure}[h!]
    \centering
    \begin{subfigure}{0.32\textwidth}
        \centering
        \includegraphics[width=0.99\textwidth]{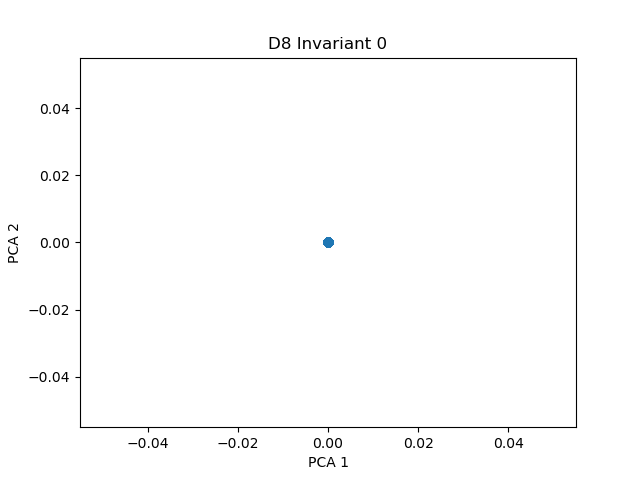}
        \caption{$\Inv_0$}
    \end{subfigure}
    \begin{subfigure}{0.32\textwidth}
        \centering
        \includegraphics[width=0.99\textwidth]{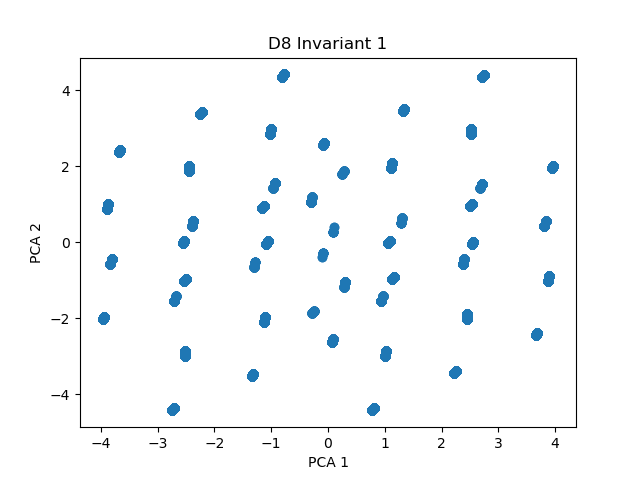}
        \caption{$\Inv_1$}
    \end{subfigure} 
    \begin{subfigure}{0.32\textwidth}
        \centering
        \includegraphics[width=0.99\textwidth]{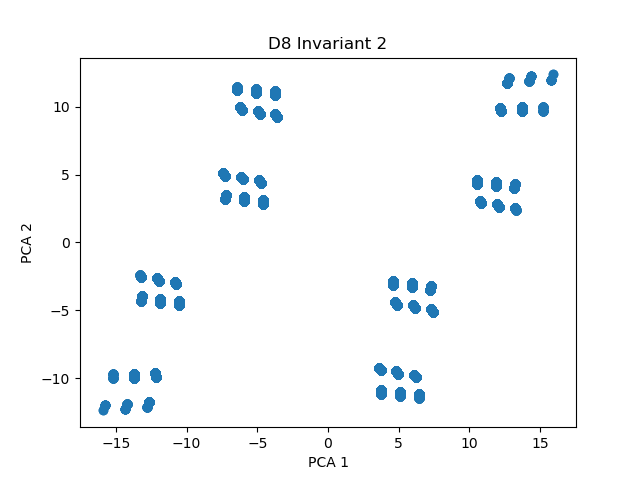}
        \caption{$\Inv_2$}
    \end{subfigure}
    \begin{subfigure}{0.32\textwidth}
        \centering
        \includegraphics[width=0.99\textwidth]{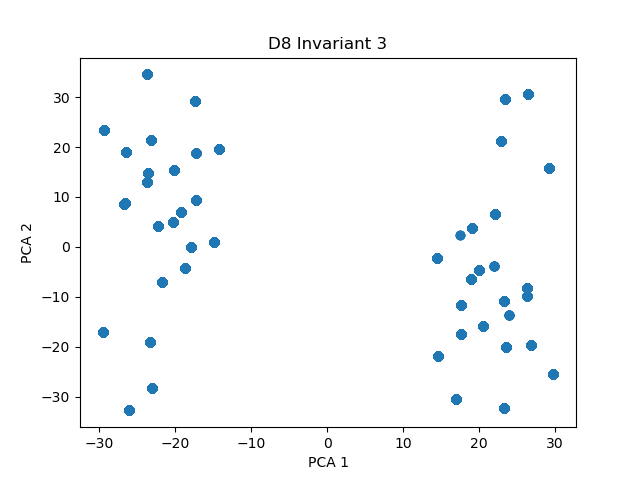}
        \caption{$\Inv_3$}
    \end{subfigure}
    \begin{subfigure}{0.32\textwidth}
        \centering
        \includegraphics[width=0.99\textwidth]{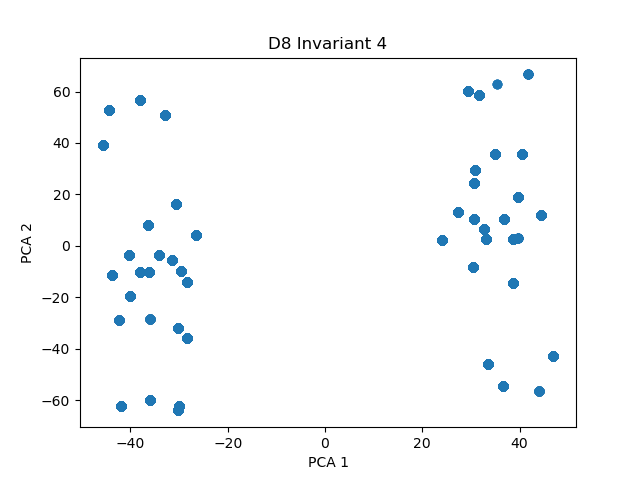}
        \caption{$\Inv_4$}
    \end{subfigure} 
    \begin{subfigure}{0.32\textwidth}
        \centering
        \includegraphics[width=0.99\textwidth]{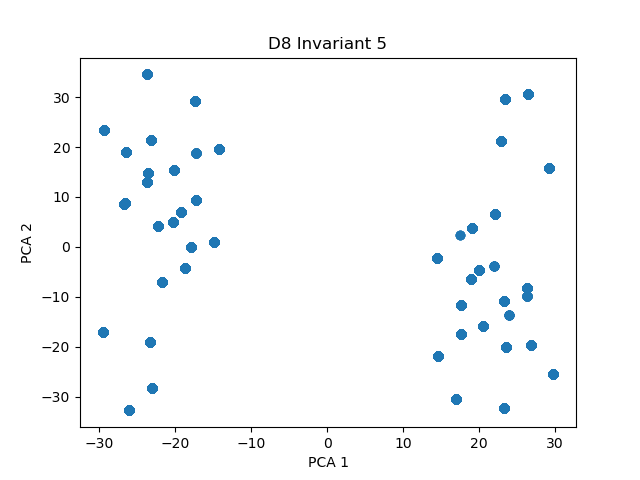}
        \caption{$\Inv_5$}
    \end{subfigure}
    \begin{subfigure}{0.32\textwidth}
        \centering
        \includegraphics[width=0.99\textwidth]{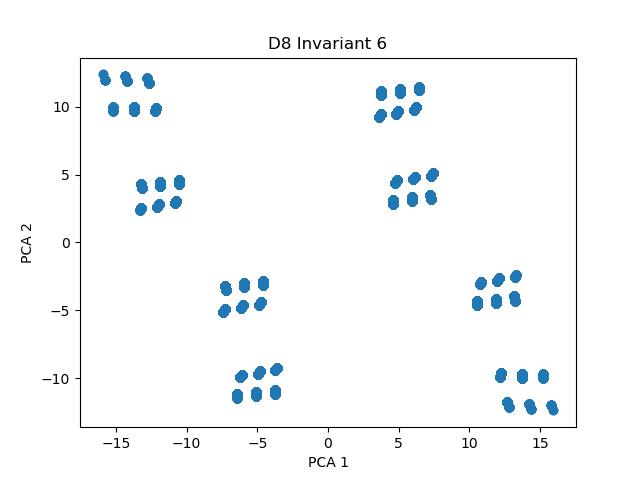}
        \caption{$\Inv_6$}
    \end{subfigure}
    \begin{subfigure}{0.32\textwidth}
        \centering
        \includegraphics[width=0.99\textwidth]{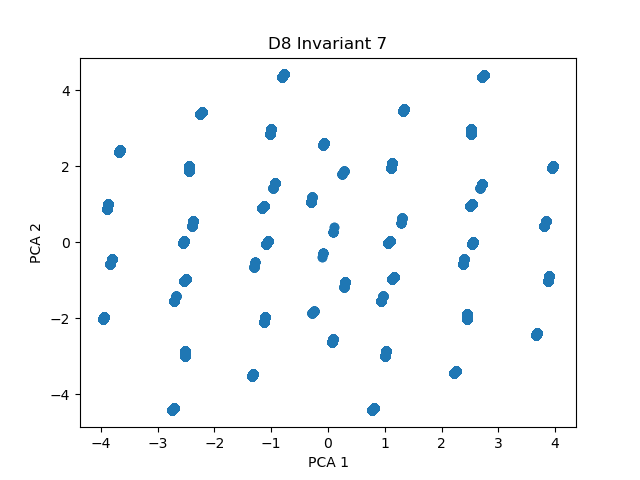}
        \caption{$\Inv_7$}
    \end{subfigure} 
    \begin{subfigure}{0.32\textwidth}
        \centering
        \includegraphics[width=0.99\textwidth]{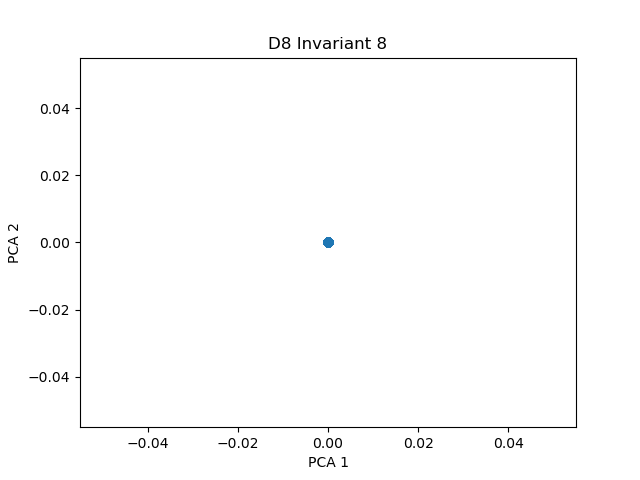}
        \caption{$\Inv_8$}
    \end{subfigure}
    \caption{PCA plots of the 9 orders of invariant for $D_{8}$.}\label{fig:PCA_D8}
\end{figure}

\begin{figure}[h!]
    \centering
    \begin{subfigure}{0.32\textwidth}
        \centering
        \includegraphics[width=0.99\textwidth]{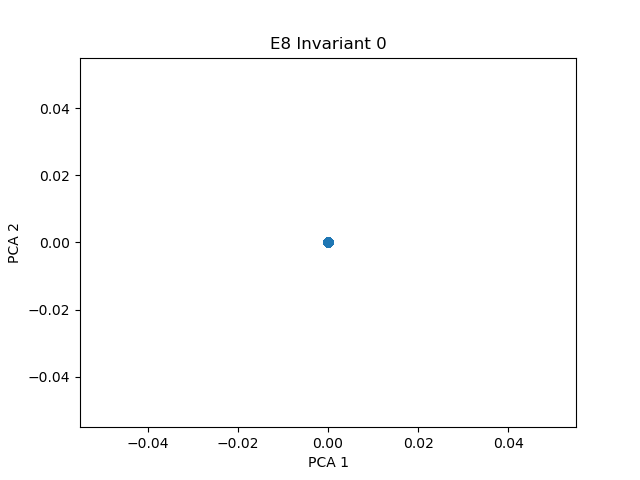}
        \caption{$\Inv_0$}
    \end{subfigure}
    \begin{subfigure}{0.32\textwidth}
        \centering
        \includegraphics[width=0.99\textwidth]{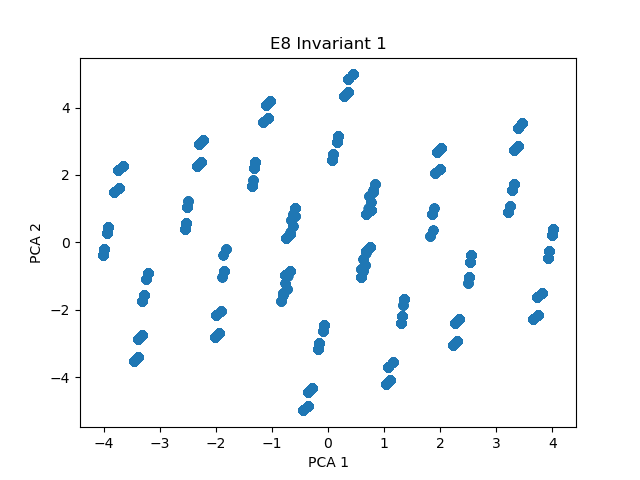}
        \caption{$\Inv_1$}
    \end{subfigure} 
    \begin{subfigure}{0.32\textwidth}
        \centering
        \includegraphics[width=0.99\textwidth]{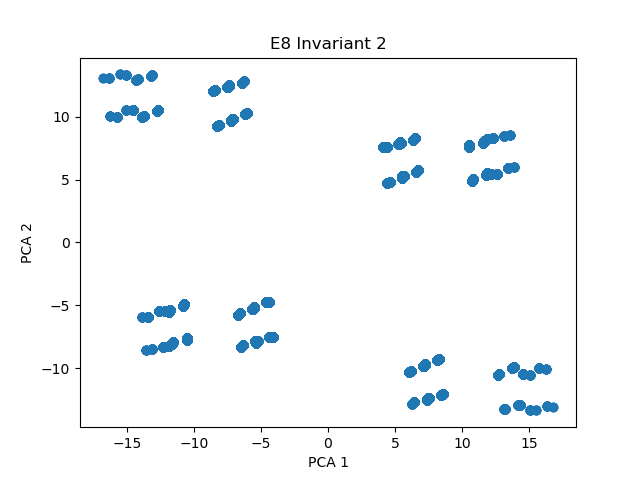}
        \caption{$\Inv_2$}
    \end{subfigure}
    \begin{subfigure}{0.32\textwidth}
        \centering
        \includegraphics[width=0.99\textwidth]{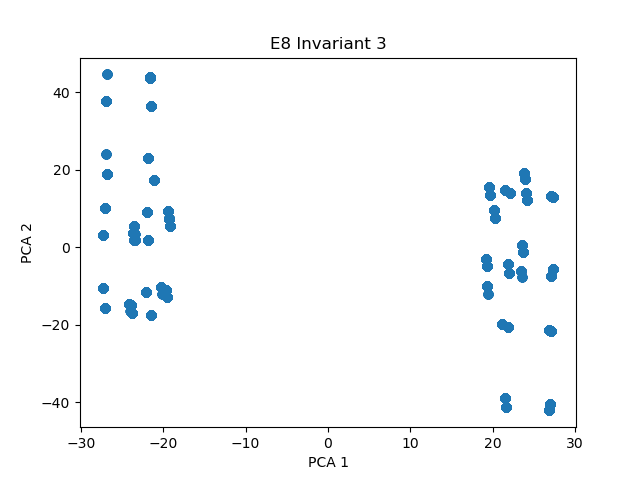}
        \caption{$\Inv_3$}
    \end{subfigure}
    \begin{subfigure}{0.32\textwidth}
        \centering
        \includegraphics[width=0.99\textwidth]{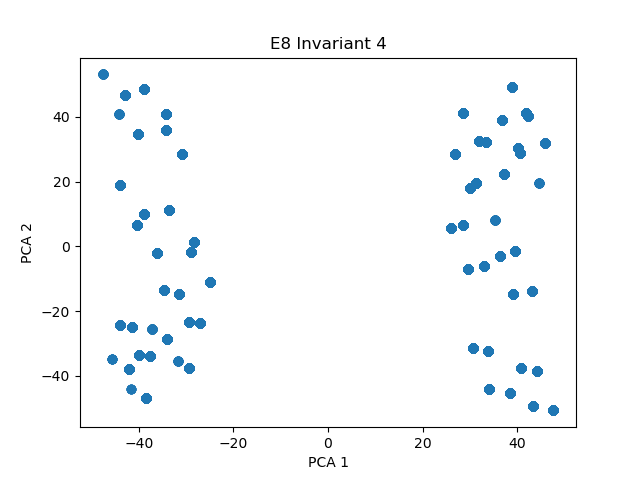}
        \caption{$\Inv_4$}
    \end{subfigure} 
    \begin{subfigure}{0.32\textwidth}
        \centering
        \includegraphics[width=0.99\textwidth]{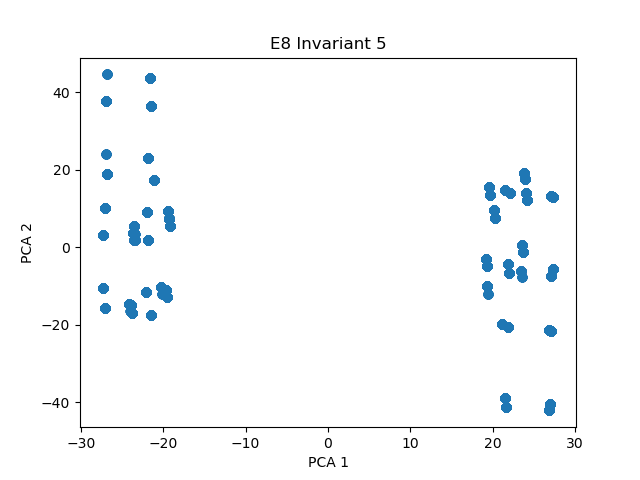}
        \caption{$\Inv_5$}
    \end{subfigure}
    \begin{subfigure}{0.32\textwidth}
        \centering
        \includegraphics[width=0.99\textwidth]{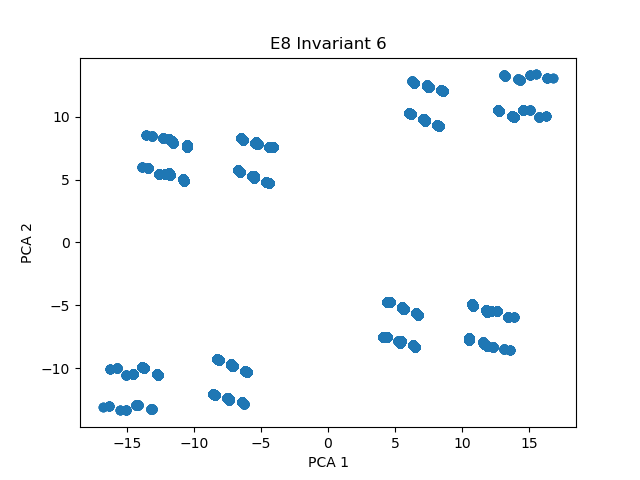}
        \caption{$\Inv_6$}
    \end{subfigure}
    \begin{subfigure}{0.32\textwidth}
        \centering
        \includegraphics[width=0.99\textwidth]{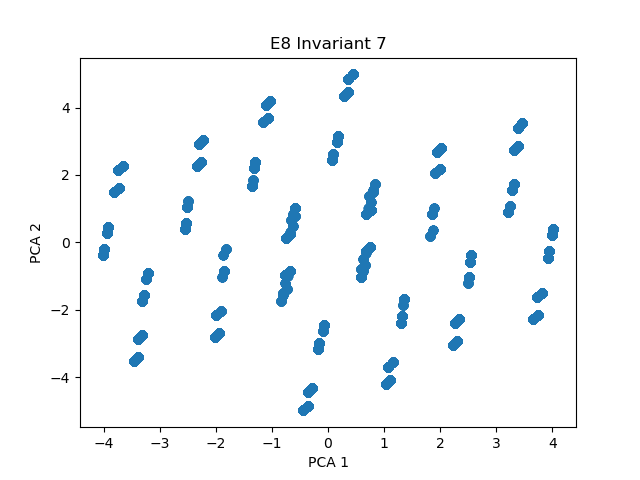}
        \caption{$\Inv_7$}
    \end{subfigure} 
    \begin{subfigure}{0.32\textwidth}
        \centering
        \includegraphics[width=0.99\textwidth]{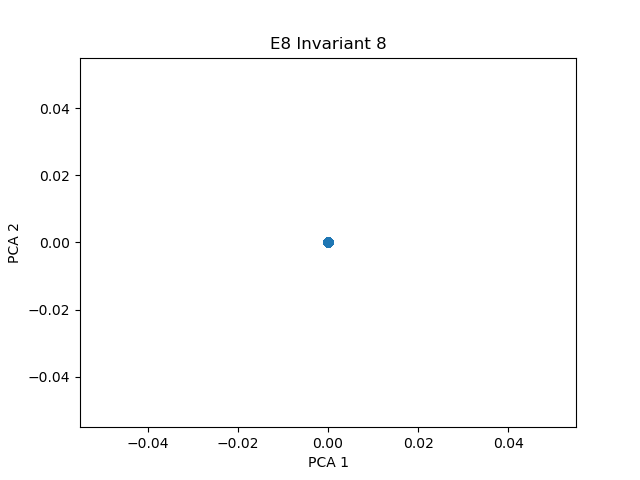}
        \caption{$\Inv_8$}
    \end{subfigure}
    \caption{PCA plots of the 9 orders of invariant for $E_{8}$.}\label{fig:PCA_E8}
\end{figure}

\begin{figure}[h!]
    \centering
    \begin{subfigure}{0.32\textwidth}
        \centering
        \includegraphics[width=0.99\textwidth]{Figures/A8_I0.png}
        \caption{$\Inv_0$}
    \end{subfigure}
    \begin{subfigure}{0.32\textwidth}
        \centering
        \includegraphics[width=0.99\textwidth]{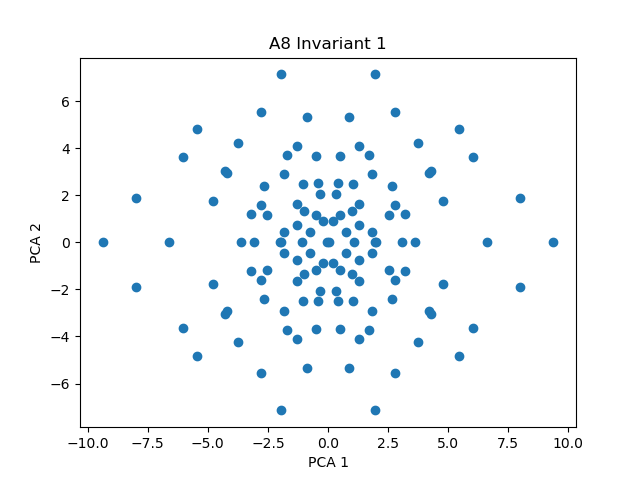}
        \caption{$\Inv_1$}
    \end{subfigure} 
    \begin{subfigure}{0.32\textwidth}
        \centering
        \includegraphics[width=0.99\textwidth]{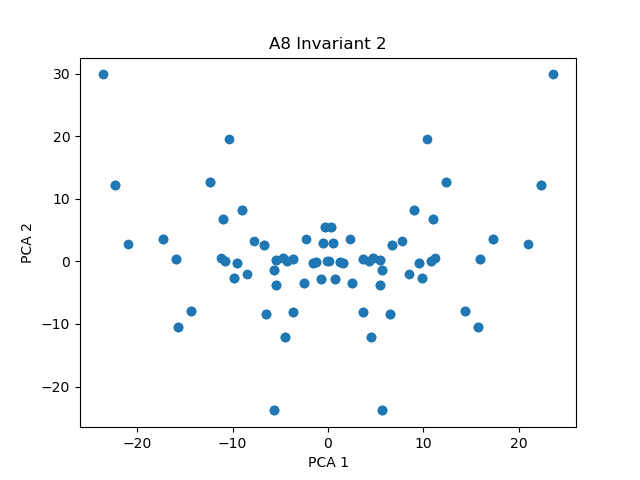}
        \caption{$\Inv_2$}
    \end{subfigure}
    \begin{subfigure}{0.32\textwidth}
        \centering
        \includegraphics[width=0.99\textwidth]{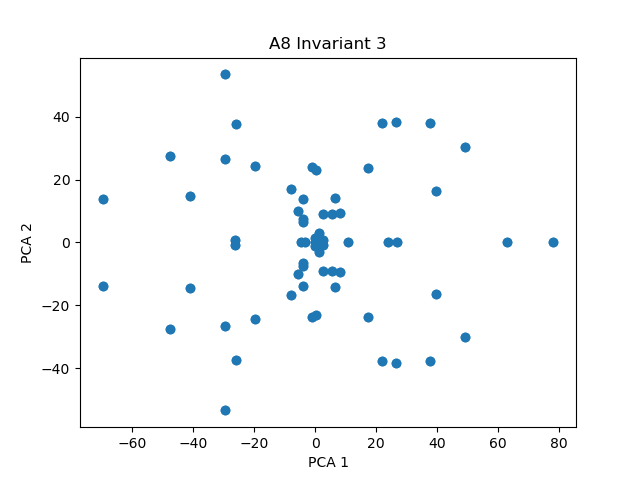}
        \caption{$\Inv_3$}
    \end{subfigure}
    \begin{subfigure}{0.32\textwidth}
        \centering
        \includegraphics[width=0.99\textwidth]{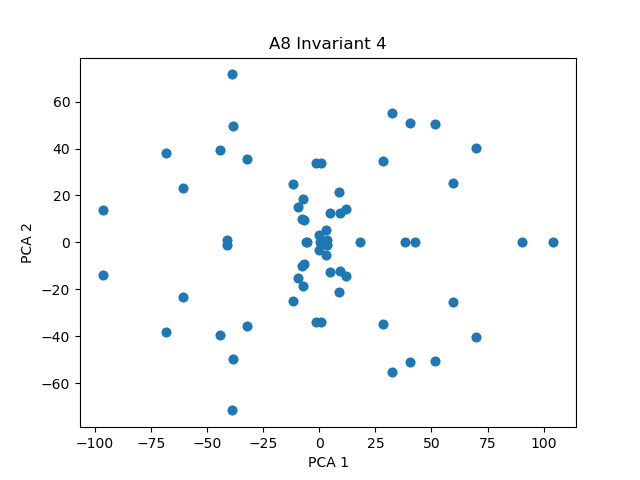}
        \caption{$\Inv_4$}
    \end{subfigure} 
    \begin{subfigure}{0.32\textwidth}
        \centering
        \includegraphics[width=0.99\textwidth]{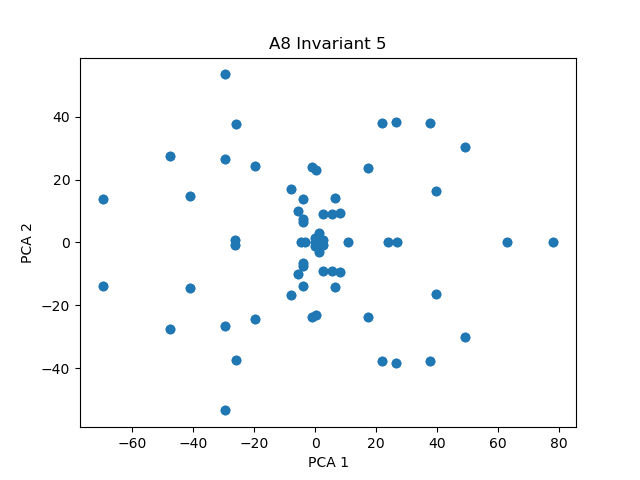}
        \caption{$\Inv_5$}
    \end{subfigure}
    \begin{subfigure}{0.32\textwidth}
        \centering
        \includegraphics[width=0.99\textwidth]{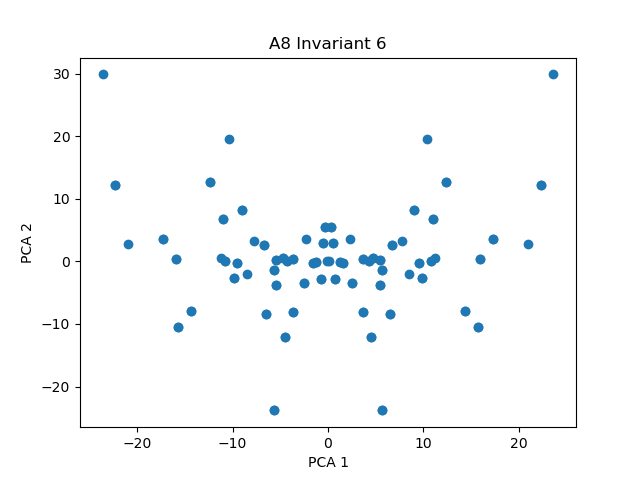}
        \caption{$\Inv_6$}
    \end{subfigure}
    \begin{subfigure}{0.32\textwidth}
        \centering
        \includegraphics[width=0.99\textwidth]{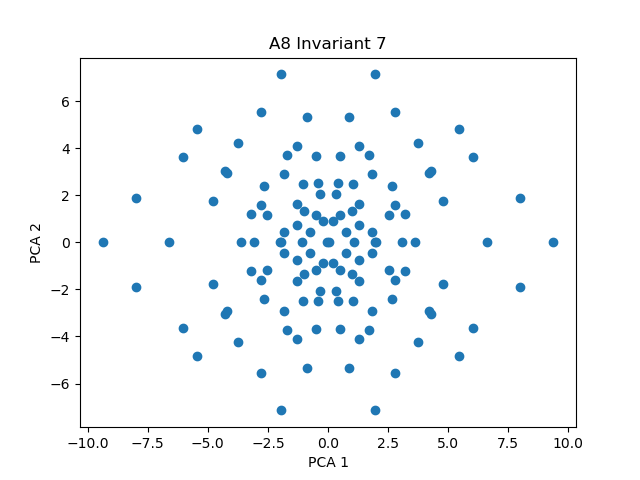}
        \caption{$\Inv_7$}
    \end{subfigure} 
    \begin{subfigure}{0.32\textwidth}
        \centering
        \includegraphics[width=0.99\textwidth]{Figures/A8_I8.png}
        \caption{$\Inv_8$}
    \end{subfigure}
    \caption{PCA plots of reduced datasets (with duplicates deleted) of the 9 orders of invariant for $A_{8}$.}\label{fig:PCA_A8_unique}
\end{figure}

\begin{figure}[h!]
    \centering
    \begin{subfigure}{0.32\textwidth}
        \centering
        \includegraphics[width=0.99\textwidth]{Figures/D8_I0.png}
        \caption{$\Inv_0$}
    \end{subfigure}
    \begin{subfigure}{0.32\textwidth}
        \centering
        \includegraphics[width=0.99\textwidth]{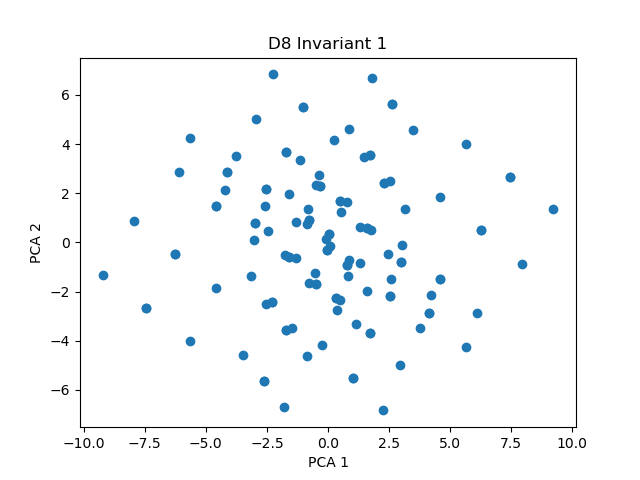}
        \caption{$\Inv_1$}
    \end{subfigure} 
    \begin{subfigure}{0.32\textwidth}
        \centering
        \includegraphics[width=0.99\textwidth]{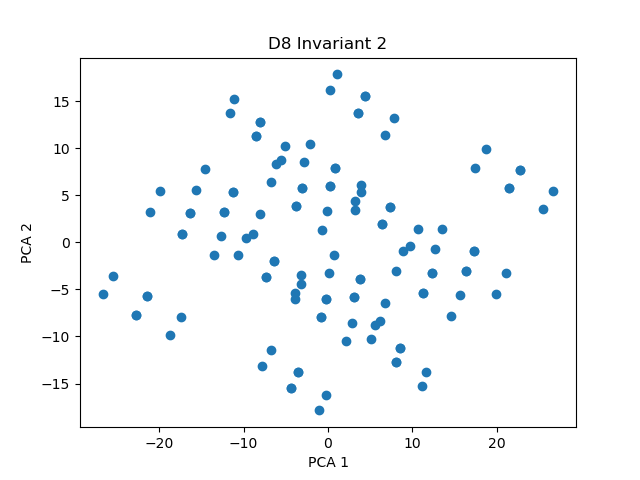}
        \caption{$\Inv_2$}
    \end{subfigure}
    \begin{subfigure}{0.32\textwidth}
        \centering
        \includegraphics[width=0.99\textwidth]{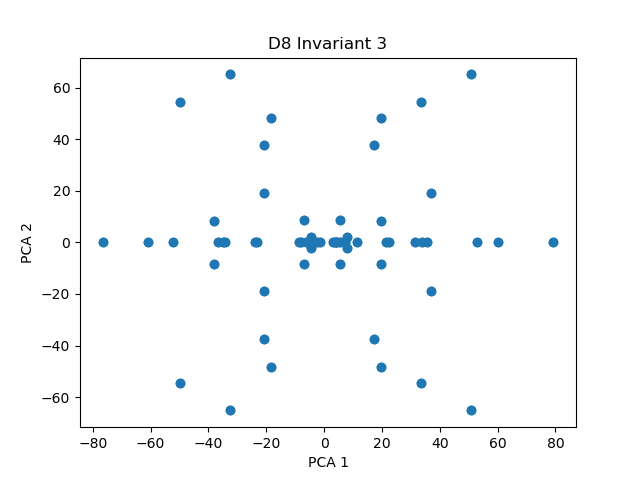}
        \caption{$\Inv_3$}
    \end{subfigure}
    \begin{subfigure}{0.32\textwidth}
        \centering
        \includegraphics[width=0.99\textwidth]{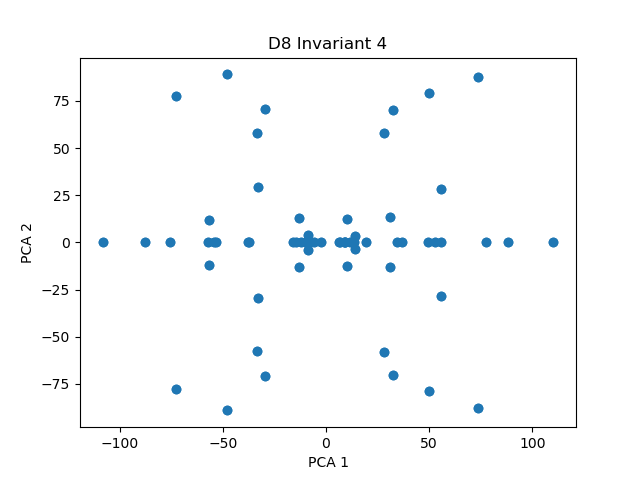}
        \caption{$\Inv_4$}
    \end{subfigure} 
    \begin{subfigure}{0.32\textwidth}
        \centering
        \includegraphics[width=0.99\textwidth]{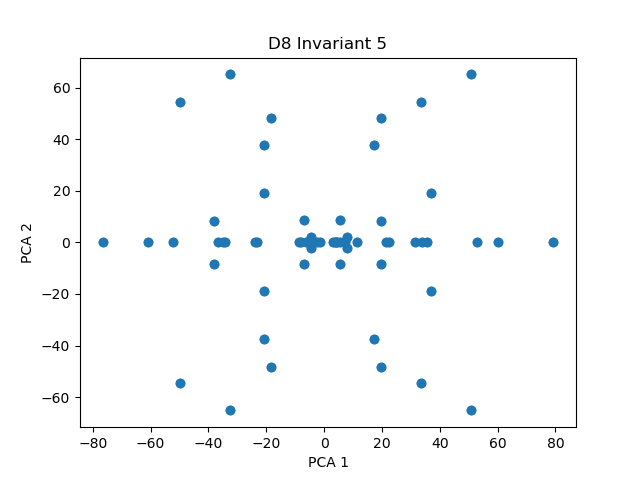}
        \caption{$\Inv_5$}
    \end{subfigure}
    \begin{subfigure}{0.32\textwidth}
        \centering
        \includegraphics[width=0.99\textwidth]{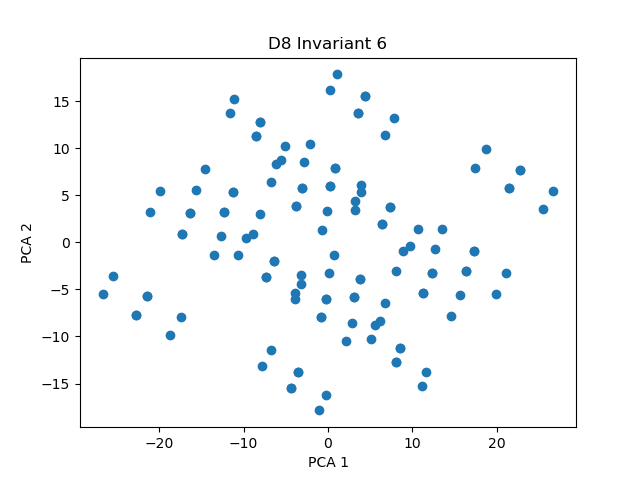}
        \caption{$\Inv_6$}
    \end{subfigure}
    \begin{subfigure}{0.32\textwidth}
        \centering
        \includegraphics[width=0.99\textwidth]{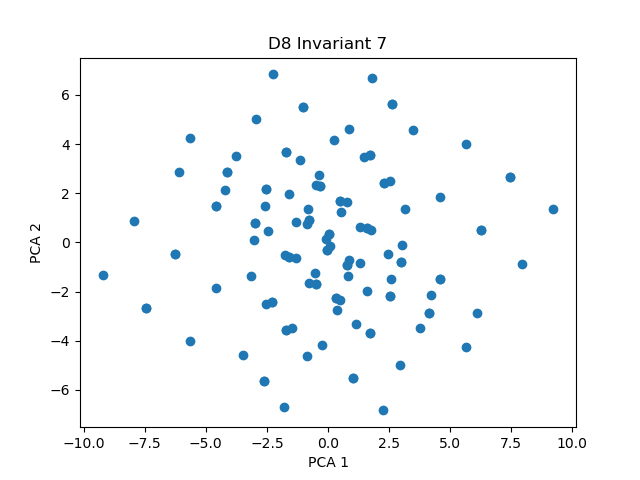}
        \caption{$\Inv_7$}
    \end{subfigure} 
    \begin{subfigure}{0.32\textwidth}
        \centering
        \includegraphics[width=0.99\textwidth]{Figures/D8_I8.png}
        \caption{$\Inv_8$}
    \end{subfigure}
    \caption{PCA plots of reduced datasets (with duplicates deleted) of the 9 orders of invariant for $D_{8}$.}\label{fig:PCA_D8_unique}
\end{figure}

\begin{figure}[h!]
    \centering
    \begin{subfigure}{0.32\textwidth}
        \centering
        \includegraphics[width=0.99\textwidth]{Figures/E8_I0.png}
        \caption{$\Inv_0$}
    \end{subfigure}
    \begin{subfigure}{0.32\textwidth}
        \centering
        \includegraphics[width=0.99\textwidth]{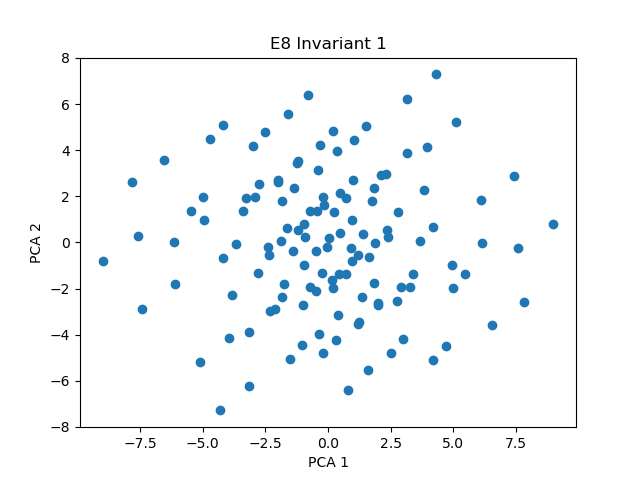}
        \caption{$\Inv_1$}
    \end{subfigure} 
    \begin{subfigure}{0.32\textwidth}
        \centering
        \includegraphics[width=0.99\textwidth]{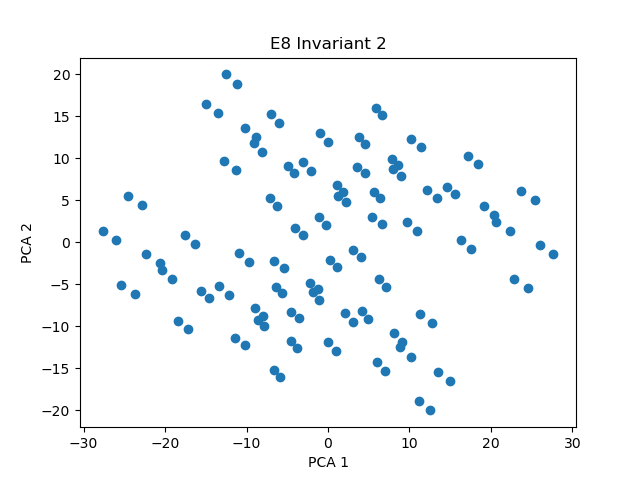}
        \caption{$\Inv_2$}
    \end{subfigure}
    \begin{subfigure}{0.32\textwidth}
        \centering
        \includegraphics[width=0.99\textwidth]{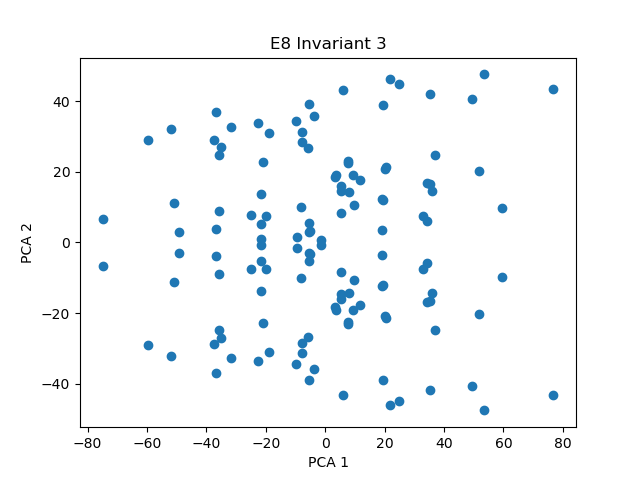}
        \caption{$\Inv_3$}
    \end{subfigure}
    \begin{subfigure}{0.32\textwidth}
        \centering
        \includegraphics[width=0.99\textwidth]{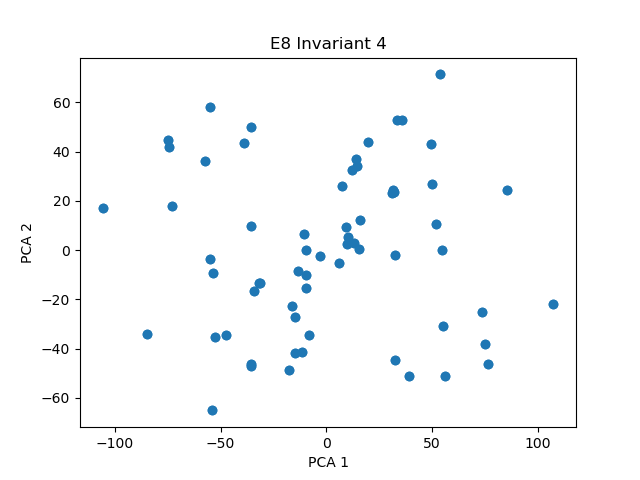}
        \caption{$\Inv_4$}
    \end{subfigure} 
    \begin{subfigure}{0.32\textwidth}
        \centering
        \includegraphics[width=0.99\textwidth]{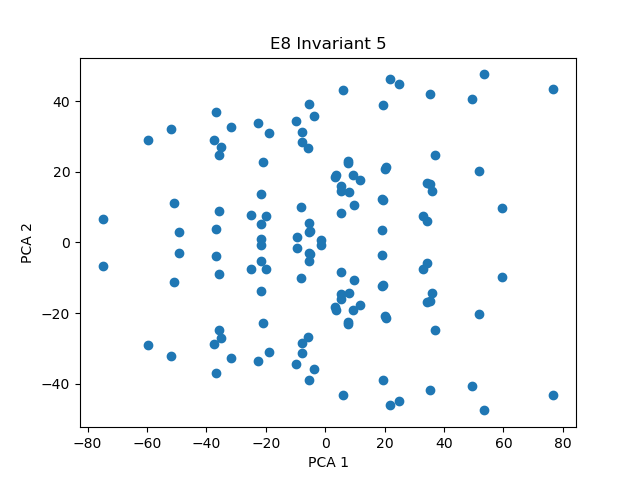}
        \caption{$\Inv_5$}
    \end{subfigure}
    \begin{subfigure}{0.32\textwidth}
        \centering
        \includegraphics[width=0.99\textwidth]{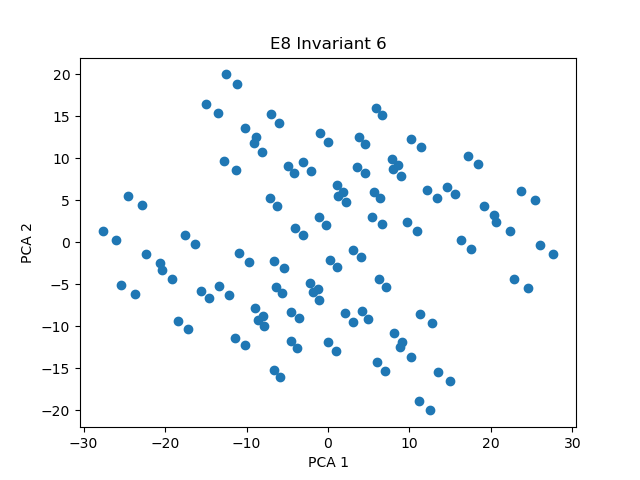}
        \caption{$\Inv_6$}
    \end{subfigure}
    \begin{subfigure}{0.32\textwidth}
        \centering
        \includegraphics[width=0.99\textwidth]{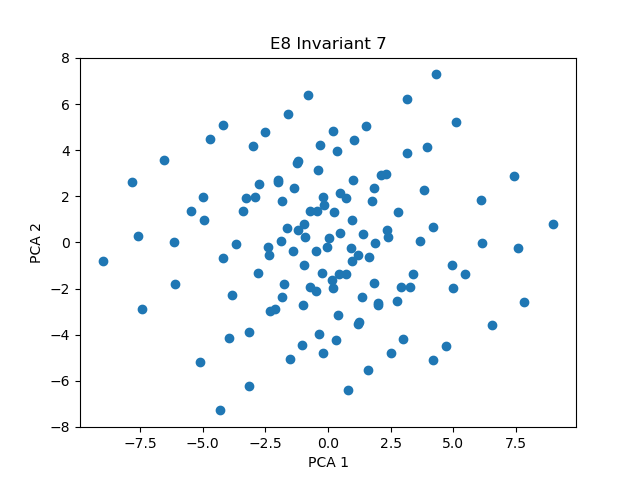}
        \caption{$\Inv_7$}
    \end{subfigure} 
    \begin{subfigure}{0.32\textwidth}
        \centering
        \includegraphics[width=0.99\textwidth]{Figures/E8_I8.png}
        \caption{$\Inv_8$}
    \end{subfigure}
    \caption{PCA plots of reduced datasets (with duplicates deleted) of the 9 orders of invariant for $E_{8}$. }\label{fig:PCA_E8_unique}
\end{figure}

The 2-dimensional PCA projections for each dataset of invariants at each order for each root system $A_8$, $D_8$, $E_8$ are shown in Figures \ref{fig:PCA_A8}, \ref{fig:PCA_D8}, \ref{fig:PCA_E8} respectively, whilst the 2-dimensional PCA projections for the same partitioning of invariants into orders and types -- however now reducing the datasets to unique invariants -- are shown in Figures \ref{fig:PCA_A8_unique}, \ref{fig:PCA_D8_unique}, \ref{fig:PCA_E8_unique} respectively. 

\end{document}